\documentclass[twoside,11pt]{article}

\usepackage{jmlr2e}

\usepackage{lastpage}
\usepackage{amssymb,amsmath,mathrsfs,stmaryrd,bm,amsfonts}
\usepackage{mathtools}
\usepackage{tikz}
\usepackage{xcolor}
\usepackage{pgfplots,graphicx}
\usepackage{subcaption}
\usepackage{multirow}
\usepackage{makecell}
\usepackage{forloop}

\newcommand{\ie}{i.\,e.}
\newcommand{\Dkl}{D_{KL}}
\newcommand{\Cllrmin}{C_{\text{llr}}^{\text{min}}}
\newcommand{\Cllrcal}{C_{\text{llr}}^{\text{cal}}}
\newcommand{\Cllr}{C_{\text{llr}}}

\newcounter{z}

\definecolor{forestgreen}{RGB}{34,139,34}
\definecolor{Maroon}{cmyk}{0, 0.87, 0.68, 0.32}

\DeclareMathOperator{\logit}{logit}
\DeclareMathOperator{\sigmoid}{sigmoid}

\DeclareMathOperator{\EER}{EER}

\DeclareMathOperator{\ilr}{ilr}

\DeclareMathOperator{\vect}{vec}

\DeclareMathOperator{\vech}{vech}

\title{The distribution of calibrated likelihood functions on the probability-likelihood Aitchison simplex}

\author{\name Paul-Gauthier Noé\footnotemark[1] \email paul-gauthier.noe@cnrs.fr\\
  \addr Centre National de la Recherche Scientifique (CNRS)\\
  Laboratoire d'Informatique et des Systèmes (LIS)\\
  Aix-Marseille Université, France
  \AND
  \name Andreas Nautsch \\
  \addr Individual, Germany
  \AND
  \name Driss Matrouf,\\
  \addr Laboratoire d'Informatique d'Avignon (LIA)\\
  Avignon Université, France
  \AND
  \name Pierre-Michel Bousquet,\\
  \addr Laboratoire d'Informatique d'Avignon (LIA)\\
  Avignon Université, France
  \AND
  \name Jean-François Bonastre,\\
  \addr Laboratoire d'Informatique d'Avignon (LIA)\\
  Avignon Université, France}
\editor{}
\begin{document}
\renewcommand{\thefootnote}{\fnsymbol{footnote}}

\maketitle
\footnotetext[1]{Most of this work was done when Paul-Gauthier Noé was a PhD student at LIA, Avignon Université, supported by the VoicePersonae project ANR-18-JSTS-0001.}
\begin{abstract}
  While calibration of probabilistic predictions has been widely studied, this paper rather addresses calibration of likelihood functions. This has been discussed, especially in biometrics, in cases with only two exhaustive and mutually exclusive hypotheses (classes) where likelihood functions can be written as log-likelihood-ratios (LLRs). After defining calibration for LLRs and its connection with the concept of weight-of-evidence, we present the idempotence property and its associated constraint on the distribution of the LLRs. Although these results have been known for decades, they have been limited to the binary case. Here, we extend them to cases with more than two hypotheses by using the Aitchison geometry of the simplex, which allows us to recover, in a vector form, the additive form of the Bayes' rule; extending therefore the LLR and the weight-of-evidence to any number of hypotheses. Especially, we extend the definition of calibration, the idempotence, and the constraint on the distribution of likelihood functions to this multiple hypotheses and multiclass counterpart of the LLR: the isometric-log-ratio transformed likelihood function. This work is mainly conceptual, but we still provide one application to machine learning by presenting a non-linear discriminant analysis where the discriminant components form a calibrated likelihood function over the classes, improving therefore the interpretability and the reliability of the method.
\end{abstract}


\renewcommand*{\thefootnote}{\arabic{footnote}}
\setcounter{footnote}{0}
\section{Introduction}

Calibration has been introduced for probabilistic predictions in the context of weather forecasting, where a forecaster has to make previsions by assigning to each day a probability for rain \citep{brier1950,murphy1968,degroot1983}. The predictions are said to be calibrated if the probabilities match the observed outcomes: over a long sequence of predictions, the relative frequency of days where it actually rained and on which the probability $p$ has been assigned must be $p$ \citep{degroot,dawid1982}. 

In machine learning, we are rather interested in data modeling and classification tasks. Like a weather forecaster, a classifier ``is'' uncertain, and then naturally outputs a probability distribution over the set of classes (we also call \emph{hypotheses} here) rather than making a hard decision. However, for the uncertainty to be well-encoded in the classifier's probabilistic predictions, and for them to be used for cost-sensitive decisions, they have to be calibrated.

Considering a classifier $\bm{q}:\mathcal{X} \to \mathcal{S}^D$ that\footnote{$\mathcal{S}^D$ is the probability simplex. See Section \ref{sec:compo} for more details.}, given an input $x \in \mathcal{X}$, outputs a prediction in the form of a posterior probability distribution over the set of hypotheses (or classes): $\bm{q}(x) = \left[P_\theta(H_1 \mid x),\dots P_\theta(H_D \mid x)\right]^T \in \mathcal{S}^D$. A set $\mathcal{Q}$ of probabilistic predictions is \emph{perfectly} calibrated if \citep{brocker2009reliability}, 
    \begin{equation}
      \label{eq:calib}
      \forall \bm{q}(x) \in \mathcal{Q},~~~\mathbb{P}(H_i \mid \bm{q}(x)) = q_i(x) = P_\theta(H_i \mid x)~~~\forall i.
    \end{equation}
    Meaning that, for all predictions, the conditional distribution over the set of hypotheses (or classes) given the prediction, is equal to the prediction. While calibration has been discussed in machine learning for decades \citep{zadrozny2001learning,zadrozny2002}, it has been the subject of renewed interest especially since \cite{guo2017} discussed the tendency of modern neural networks to produce overconfident predictions.
    
    In this paper, we will look at calibration from a different point of view than the one people in statistics and machine learning are generally used to. In our framework, a prediction will be in the form of a likelihood function over the set of possible hypotheses. This in no way reduces the relevance of our work for the statistical machine learning community since, as we will see, probabilistic predictions and corresponding likelihood functions are isomorphic given a prior and under some scale-invariance equivalence relation. To be more precise, we are rather interested in reporting statistical evidences rather than making predictions. A probabilistic prediction can then be made by combining a prior and a statistical evidence, represented by a calibrated likelihood function, through the Bayes' rule.

    Reporting statistical evidence in the form of a likelihood function has been extensively discussed and promoted in forensic science \citep{aitken2004statistics,meester_slooten_2021, aitken2024role} or in the context of medical diagnostic \citep{thornbury1975likelihood}. In those cases, there are only two competiting hypotheses such that the likelihood function can be written in the form of a log-likelihood-ratio (LLR), or weight-of-evidence (WOE). The Bayes' rule can be written in its log-odds form: the posterior log-odds is the sum of the LLR and the prior log-odds. In this way, the LLR tells how new data is changing the personal belief from the prior to the posterior in an additive manner.

    The calibration of LLRs has been specially developed in the context of speaker verification \citep{brummer2006, brummer2010measuring, ramosPhd}. In particular, the idempotence property of calibrated LLRs and its associated constraint on their distribution are of great importance for understanding the calibration of LLRs. The idempotence tells that ``\textit{the LLR of the LLR is the LLR}'' and that if calibrated LLRs are normally distributed under one hypothesis, it is also normally distributed under the other hypothesis, with an opposite mean, and a shared variance equal to twice the mean. These results have been proofed for calibrated LLRs in the context of speaker verification \citep{leeuwen13_interspeech} but have been known for the WOE since at least the 40s \citep{turingstat}.

    However, the LLR, its calibration, and its associated properties are defined only for the binary case, \ie~when there are only two exhaustive and mutually exclusive hypotheses. The main purpose of this work is to extend these concepts to the multiple hypotheses and multiclass case. Starting from the log-odds form of the Bayes' rule, Section \ref{sec:llr} presents the WOE and recalls the definition of calibration for the LLR, the idempotence property, how it is related to the WOE, and the associated constraint on the distribution of calibrated LLRs. Section \ref{sec:compo} presents the Aitchison geometry of the probability simplex \citep{aitchison1982}. We will see how this allows us to extend the log-odds and additive form of the Bayes' rule generalizing therefore the concept of LLR, in a vector form, to a multiple hypotheses setting. This multiple hypotheses counterpart of the LLR is called the isometric-log-ratio transformed likelihood function (ILRL) and we will see in Section \ref{sec:ILRL} how the idempotence property applies to it.  We will also see how the constraint on the distribution generalizes to ILRs \ie~to likelihood functions over a set of more than two possible hypotheses. Finally, Section \ref{sec:discrim_ILRL} presents one application of these results to machine learning, by proposing a non-linear discriminant analysis where the discriminant space is designed according to the idempotence property to form a space of calibrated likelihood functions.

    Our contributions can be summarized as follows:
    \begin{itemize}
    \item By taking the work by \cite{egozcue2018evidence} over, we extend the concept of LLR to any number of hypotheses thanks to the Aitchison geometry of the simplex. The resulting quantity is the isometric-log-ratio transformed likelihood function (ILRL);
    \item We extend the concept of calibration and the idempotence of the LLR to its multiple hypotheses counterpart: the ILRL;
    \item We prove a constraint on the distribution of calibrated ILRLs: if they are normally distributed under one hypothesis, they are also normally distributed for the other hypotheses with some additional constraints on their parameter. This result generalizes what has been known for the weight-of-evidence and calibrated LLRs for decades \citep{peterson1954,turingstat,good1985weight,leeuwen13_interspeech};
    \item We present, as an application of the above results, a non-linear discriminant analysis we call \emph{Compositional discriminant analysis}, where the discriminant components form a calibrated likelihood function over the set of classes making this approach reliable and easy-to-interpret\footnote{This contribution has been presented as a poster at CoDaWork2024, the 10th International Workshop on Compositional Data Analysis \citep{noe2024compositional}. The binary case has been presented earlier in the context of privacy preservation in speech technologies \citep{Noe22Icassp}.}.
    \end{itemize}

    % where $\mathbb{P}$ usually refers to the concept of \emph{``true''} probability \cite{guo2017,vaicenavicius2019evaluating}. However, in addition of being ill-defined, such probability is not available in general\footnote{Except is some simple cases like in \cite{degroot1983} where the set of allowed prediction is finite and discrete allowing the calculation of the relative frequencies as ``true'' probabilities.}. We prefer the concept of \emph{reference} distribution as recently introduced to the machine learning community in \cite{ferrer2025evaluating}. In any case, these considerations are not so important for understanding our work. When we refer to a so-called ``true'' probability or distribution, we simply mean their empirical counterpart. Indeed, no matter whether it exists or not, this probability is generally not reachable such that we rely instead on observed data.
    % \begin{definition}
    %   \label{def:calibprob}
    %   A set $\mathcal{Q}$ of probabilistic predictions is \emph{perfectly} calibrated with respect to a reference probability distribution $\left[ P_r(H_1 \mid \bm{q}(x)), \dots P_r(H_D \mid \bm{q}(x))\right] \in \mathcal{S}^D$ if, 
    %   \begin{equation}
    %     \forall \bm{q}(x) \in \mathcal{Q},~~~P_r(H_i \mid \bm{q}(x)) = q_i(x) = P_c(H_i \mid x)~~~\forall i.
    %   \end{equation}
    % \end{definition}
    % Note that $P_c$ refers to the probability distribution assigned by the classifier while $P_r$ refers to the reference probability distribution. The latter is discussed 
\section{From the weight-of-evidence to calibrated log-likelihood-ratios}
\label{sec:llr}

 Let's consider a set of exhaustive and mutually exclusive \emph{simple} hypotheses $\mathcal{H} =\{ H_1, \dots H_D \}$\footnote{In a classification context, each hypothesis would correspond to a class such that for a given sample, the hypothesis $H_i$ should be read as ``the sample belongs to the $i$th class''.}. Let's consider an individual who wants to infer which hypothesis is true given the data or \emph{evidence} $x$. Its posterior probabilities are given by the Bayes' rule:
\begin{equation}
  \label{eq:bayes}
  \forall H \in \mathcal{H},~P(H \mid x) \propto P(x\mid H)P(H),
\end{equation}
where $\left[ P(x \mid H_i) \right]_{1 \leq i \leq D} \in \mathbb{R}^{*D}_+$ is the likelihood function over the set of hypotheses, %function\footnote{Note that toward this work, \emph{likelihood function} refers most of the time to a function of the hypothesis or class. Not to be confused with a function of some statistical parameter or a function of the data.} over the set of hypotheses $\mathcal{H}$
and $\left[ P(H_i) \right]_{1 \leq i \leq D} \in \mathcal{S}^D$ is the prior probability distribution representing the prior personal belief of the individual, \ie~its belief based on all the information available to him or her other than the evidence $x$. When there are only two competing hypotheses, \ie~$\mathcal{H} = \left\{ H_1, H_2\right\}$, the Bayes' rule can be written in its \emph{log-odds} or \emph{logit} form:
\begin{equation}
  \label{eq:logit}
 \underbrace{\logit P(H_1 \mid x)}_{\text{posterior log-odds}} = \underbrace{\log \frac{P(x \mid H_1)}{P(x \mid H_2)}}_{\substack{\text{weight-of-evidence}\\\text{(log-likelihood-ratio)}}} + \underbrace{\logit P(H_1)}_{\text{prior log-odds}}
\end{equation}
%because $P(H_2) = 1-P(H_1)$ and $P(H_2 \mid x) = 1-P(H_1 \mid x)$, and
where $\logit(p) = \log \frac{p}{1-p}$ for $0<p<1$.

The posterior is here the sum between a term that depends only on the prior probabilities and a term that depends only on the likelihoods. The latter is the \emph{weight-of-evidence}---or log \emph{Bayes-factor}---and informs about the contribution of the data $x$ in the computation of the posterior. In \cite{good1985weight}, the author wrote that ``\textit{[...] the weight-of-evidence tells us just as much as [$x$] does about the odd of [$H_1$ and $H_2$]}'' stating therefore that:
\begin{equation}
  \label{eq:woe}
  w(x) = \log \frac{P\left(x \mid H_1\right)}{P\left(x \mid H_2\right)} = \log \frac{P\left(w(x) \mid H_1\right)}{P\left(w(x) \mid H_2\right)}.
\end{equation}
This makes the weight-of-evidence $w(x)$ a good candidate for representing the statistical evidence---in favor of $H_1$ and against $H_2$---in the data $x$.

However, the hypotheses are here \emph{simple} statistical hypotheses, such that Equation \ref{eq:woe} is an intrinsic property of the weight-of-evidence as in \cite{meester_slooten_2021} and \cite{good1985weight}. In machine learning, especially with generative classifiers, the likelihoods $P( x \mid H_1)$ and $P( x \mid H_2)$ are computed with respect to statistical models that may not reflect the ``true'' distribution of the data%\footnote{Such a ``true'' distribution is ill-defined. Let's simply see it as the \emph{empirical} distribution. Indeed, no matter whether such a ``true'' distribution exists or not, it is not known in practice such that we rely instead on observed data.}
. This would result in an uncalibrated representation of the statistical evidence. Equation \ref{eq:woe} becomes therefore a desired property for the log-ratio of the likelihoods---computed with respect to the models---to properly represent the statistical evidence, and to be interpreted as a weight-of-evidence.
    
\subsection{Calibration for log-likelihood-ratios}
\label{sec:calibLLR}

In machine learning, especially in the context of generative classification, we do not have access to a weight-of-evidence. We rather compute a log-likelihood-ratio (LLR) as a log-ratio of probability density functions:
\begin{equation}
  \label{eq:llrmodel}
l_{\theta}(x) = \log \frac{f_{\theta_{\mathcal{X}_1}}\left( x \right)}{f_{\theta_{\mathcal{X}_2}}\left( x \right)}
\end{equation}
where $\theta_{\mathcal{X}_i}$ refers to a statistical model for the data under hypothesis $H_i$. The classifier produces here a LLR and the posterior is obtained as a function of the LLR and a prior\footnote{Usually taken as the empirical class proportion in the training set.} through the Bayes' rule:
\begin{equation}
  \label{eq:bayes2}
  \begin{aligned}
    &\logit P_\theta(H_1 \mid x) = \log \frac{f_{\theta_{\mathcal{X}_1}}\left( x \right)}{f_{\theta_{\mathcal{X}_2}}\left( x \right)} + \logit P(H_1)\\
    &~~~~~~\iff q_1(x) = P_\theta(H_1 \mid x)
    = \sigmoid \left( \log \frac{f_{\theta_{\mathcal{X}_1}}\left( x \right)}{f_{\theta_{\mathcal{X}_2}}\left( x \right)} + \logit P(H_1) \right),
  \end{aligned}
\end{equation}
where $\sigmoid(l) = 1/(1+\exp(-l))$, with $l \in \mathbb{R}$, and is the inverse of the $\logit$.

From Equation \ref{eq:bayes2}, we can see that, for a given a prior, there is a bijection between the posterior and the LLR. In the definition of calibration in Equation \ref{eq:calib}, we can therefore interchange the set of probabilistic prediction $\mathcal{Q}$ with the set $\mathcal{L}$ of corresponding LLRs:

$\forall l_{\theta} \in \mathcal{L} = \left\{ l_{\theta}(x) \mid x \in \mathcal{X} \right\},$
\begin{align}
     \label{eq:calibprobllr}
  ~~~&\mathbb{P}(H_i\mid l_{\theta})=q_i(x) =P_{\theta}(H_i \mid x)~~~\forall i \in \{1,2\},\\
    &~~~\iff \log \frac{\mathbb{P}\left(H_1 \mid l_{\theta}\right)}{\mathbb{P}\left(H_2 \mid l_{\theta}\right)} = \log \frac{P_{\theta}\left(H_1 \mid x\right)}{P_{\theta}\left(H_2 \mid x\right)},\\
                                 &~~~\iff \log \frac{f_{\mathcal{L}_1}\left(l \right)}{f_{\mathcal{L}_2}\left(l\right)} + \log \frac{P(H_1)}{P(H_2)} = \log \frac{f_{\theta_{\mathcal{X}_1}}\left(x \right)}{f_{\theta_{\mathcal{X}_2}}\left(x\right)} + \log \frac{P(H_1)}{P(H_2)},\\
                                                                           &~~~\iff \log \frac{f_{\mathcal{L}_1}\left(l_{\theta} \right)}{f_{\mathcal{L}_2}\left(l_{\theta} \right)} = l_{\theta},
                                                                             \label{eq:idemp}
\end{align}
where $f_{\mathcal{L}_i}$ is the probability density function\footnote{Assuming it exists.} of the ``true'' distribution %\footnote{Again, such a ``true'' distribution is ill-defined. Let's simply see it as the \emph{empirical} distribution.}
of the LLR under hypothesis $H_i$. The last line can be read as:
\begin{equation*}
  \textit{``The LLR of the LLR is the LLR''}.
\end{equation*}
This expression was popularized in the context of calibrated LLRs for speaker verification systems \citep{leeuwen13_interspeech} but can be traced back to the theory of signal detectability \citep{birdsall1966theory}.

This is the \emph{idempotence} property of calibrated LLRs and takes us back to Equation \ref{eq:woe} where the weight-of-evidence of the weight-of-evidence is the weight-of-evidence itself. In a way, one intuition behind the calibration of log-likelihood-ratios is to make them interpretable as weights-of-evidence.
  
The equality in Equation \ref{eq:idemp} may not hold because the actual distribution of the LLR may not match the statistical models' assumed distribution. Hence the following definition of calibrated LLRs \citep{leeuwen13_interspeech}:
\begin{definition}
  A set $\mathcal{L}$ of log-likelihood-ratios is perfectly calibrated if they are \emph{idempotent}: % meaning that the LLR of the LLR is the LLR:
  \begin{equation}
    \forall l_{\theta} \in \mathcal{L},~~~\log \frac{f_{\mathcal{L}_1}\left(l_{\theta} \right)}{f_{\mathcal{L}_2}\left(l_{\theta} \right)} = l_{\theta}.% \log \frac{f_{\theta_r}\left(l(x) \mid H_1\right)}{f_{\theta_r}\left(l(x) \mid H_2\right)}
    \label{eq:defcalibllr}
 \end{equation}
    \label{def:defcalibllr}
  \end{definition}

  The intuition is the same as the standard definition of calibration in Equation \ref{eq:calib}, where we want the conditional distribution over the hypotheses given the prediction to be equal to the prediction. Here, we want the log-likelihood-ratio of the prediction---where the prediction is here in a form of a LLR---to be equal to the prediction, \ie~the LLR.

  In the following, we will see how the idempotence property leads to a constraint on the distribution of the LLRs.
  
\subsection{The distribution of calibrated LLRs}
\label{sec:distribLLR}  
Equation \ref{eq:defcalibllr} can be rewritten as $f_{\mathcal{L}_1}(l) = e^l f_{\mathcal{L}_2}(l)$. This shows that if the distribution of the log-likelihood-ratio is known for one hypothesis, the distribution under the other hypothesis is completely determined. Therefore, the idempotence property leads to a constraint on the distribution of calibrated LLRs. The Gaussian case is illustrated by the following proposition.
\begin{theorem}
  \label{prop:distrib_LLR}
  If $l \mid H_1 \sim \mathcal{N}(\mu, \sigma^2)$, then $l \mid H_2 \sim \mathcal{N}(-\mu, \sigma^2)$ and $\sigma^2 = 2 \mu$.
\end{theorem}

In other words, if calibrated LLRs are normally distributed for one hypothesis, they are necessarily normally distributed for the other hypothesis, with an opposite mean, and the variances are the same and are equal to twice the mean.

Because of Equation \ref{eq:woe}, this result holds for the weight-of-evidence and is known by statisticians since at least I.J. Good and A.M. Turing's work \citep{turingstat,good1985weight}, and has been used in detection theory for radars like in \citep{peterson1954}. It has then been reproofed and discussed later in the context of calibrated LLRs with applications on speaker verification \citep{leeuwen13_interspeech} and in the context of forensic identification \citep{meester_slooten_2021}. We give a detailed proof in Appendix \ref{app:distributionLLR}.

\paragraph{Example:} In order to illustrate Definition \ref{def:defcalibllr} and Theorem \ref{prop:distrib_LLR}, let's consider the linear discriminant analysis as a simple example. Let's consider two classes $H_1$ and $H_2$ for which the data is assumed to be normally distributed with the same covariance matrix:
\begin{equation}
  \begin{aligned}
    \bm{x} \mid H_1 &\sim \mathcal{N}\left( \bm{\mu}_1 , \bm{\Sigma}\right),\\
    \bm{x} \mid H_2 &\sim \mathcal{N}\left( \bm{\mu}_2 , \bm{\Sigma}\right).
  \end{aligned}
  \label{eq:gaussassum}
\end{equation}
The LLR is given by:
\begin{equation}
  l(\bm{x}) = \log \frac{f_{\theta_{\mathcal{X}_1}}\left( \bm{x} \right)}{f_{\theta_{\mathcal{X}_2}}\left( \bm{x} \right)} = \bm{x}^T\bm{\Sigma}^{-1}(\bm{\mu}_1-\bm{\mu}_2) + \frac{1}{2} \left( \bm{\mu}_2^T \bm{\Sigma}^{-1}\bm{\mu}_2 - \bm{\mu}_1^T \bm{\Sigma}^{-1}\bm{\mu}_1 \right).
  \label{eq:lda}
\end{equation}
Let's consider that the Gaussian assumptions are correct and that the data is indeed distributed according to Equation \ref{eq:gaussassum} (this is like having \emph{simple} hypotheses where each hypothesis specify the distribution of the data). Then, for each class, $l(\bm{x})$ is normally distributed (because this is a projection of a normally distributed random variable). It can be shown that for the $i$th class, the mean and variance of the LLR are:
\begin{equation}
  \begin{aligned}
    \mathbb{E}_{\bm{x}\sim\mathcal{N}\left( \bm{\mu}_i , \bm{\Sigma} \right)} \left[ l(\bm{x}) \right] &= \frac{(-1)^{i-1}}{2}\left( \bm{\mu}_1 - \bm{\mu}_2 \right)^T \bm{\Sigma}^{-1} \left( \bm{\mu}_1 - \bm{\mu}_2 \right),\\
                                                                                                       &= (-1)^{i-1}\mu,~~~\text{where}~ \mu = \frac{1}{2}\left( \bm{\mu}_1 - \bm{\mu}_2 \right)^T \bm{\Sigma}^{-1} \left( \bm{\mu}_1 - \bm{\mu}_2 \right)\\
    \mathbb{V}_{\bm{x}\sim\mathcal{N}\left( \bm{\mu}_i , \bm{\Sigma} \right)} \left[ l(\bm{x}) \right] &= \left( \bm{\mu}_1 - \bm{\mu}_2 \right)^T \bm{\Sigma}^{-1} \left( \bm{\mu}_1 - \bm{\mu}_2 \right) = 2\mu,
  \end{aligned}
\end{equation}
respecting therefore Theorem \ref{prop:distrib_LLR}. With the normally distributed LLRs as described just above%, where $l \mid H_1 \sim \mathcal{N}(\mu, 2\mu)$ and $l \mid H_2 \sim \mathcal{N}(-\mu, 2\mu)$ with $\mu = \frac{1}{2}\left( \bm{\mu}_1 - \bm{\mu}_2 \right)^T \bm{\Sigma}^{-1} \left( \bm{\mu}_1 - \bm{\mu}_2 \right)$
, it can also be shown that $\log \frac{f_{\mathcal{L}_1}\left(l \right)}{f_{\mathcal{L}_2}\left(l \right)} = l$ recovering the idempotence property. However, note that the idempotence and Theorem \ref{prop:distrib_LLR} are here valid because we have considered the data as distributed according to the Gaussian assumptions. Otherwise, this would not have been the case, resulting in uncalibrated LLRs.

\begin{figure}
  \centering
  \begin{subfigure}[b]{0.3281\linewidth}
    \centering
    \pgfplotsset{
compat=1.11,
legend image code/.code={
\draw[mark repeat=2,mark phase=2]
plot coordinates {
(0cm,0cm)
(0.15cm,0cm)        %% default is (0.3cm,0cm)
(0.3cm,0cm)         %% default is (0.6cm,0cm)
};%
}
}
\begin{tikzpicture}
\begin{axis}[
        xmin=-20, xmax=20,
		ymin=0, ymax=0.11,
		grid=major, % activate major grid lines
                height=4cm,
                width=5.5cm,
		xlabel=$l$,
                ytick={0,0.1},
		samples=500
	]
	\addplot[thick,blue,domain=-20:20]
        {exp(-((x-10.8238)^2/(4*10.8238)))/(2*10.8238*sqrt(2*pi))};
        \label{plot_one}
        \addplot[thick,red,domain=-20:20]
	{exp(-((x+10.8238)^2/(4*10.8238)))/(2*10.8238*sqrt(2*pi))};
	\label{plot_two}
    \addlegendimage{/pgfplots/refstyle=plot_one}\addlegendentry{\scriptsize$f_{\mathcal{L}_1}(l)$}
    \addlegendimage{/pgfplots/refstyle=plot_two}\addlegendentry{\scriptsize$f_{\mathcal{L}_2}(l)$}
\end{axis}
\end{tikzpicture}
%%% Local Variables:
%%% mode: latex
%%% TeX-master: "../../badoc"
%%% End:
    \caption{$\EER=0.01$, $D_{KL}\approx10.8$.}
  \end{subfigure}
  \begin{subfigure}[b]{0.3281\linewidth}
    \centering
    \pgfplotsset{
compat=1.11,
legend image code/.code={
\draw[mark repeat=2,mark phase=2]
plot coordinates {
(0cm,0cm)
(0.15cm,0cm)        %% default is (0.3cm,0cm)
(0.3cm,0cm)         %% default is (0.6cm,0cm)
};%
}
}
\begin{tikzpicture}
\begin{axis}[
        xmin=-20, xmax=20,
		ymin=0, ymax=0.11,
		grid=major, % activate major grid lines
                height=4cm,
                width=5.5cm,
		xlabel=$l$,
                ytick={0,0.1},
		samples=500
	]
	\addplot[thick,blue,domain=-20:20]
        {exp(-((x-3.28475)^2/(4*3.28475)))/(2*3.28475*sqrt(2*pi))};
        \label{plot_one}
        \addplot[thick,red,domain=-20:20]
	{exp(-((x+3.28475)^2/(4*3.28475)))/(2*3.28475*sqrt(2*pi))};
	\label{plot_two}
    \addlegendimage{/pgfplots/refstyle=plot_one}\addlegendentry{\scriptsize$f_{\mathcal{L}_1}(l)$}
    \addlegendimage{/pgfplots/refstyle=plot_two}\addlegendentry{\scriptsize$f_{\mathcal{L}_2}(l)$}
\end{axis}
\end{tikzpicture}
%%% Local Variables:
%%% mode: latex
%%% TeX-master: "../../main"
%%% End:
    \caption{$\EER=0.1$, $D_{KL}\approx3.3$.}
  \end{subfigure}
  \begin{subfigure}[b]{0.3281\linewidth}
    \centering
    \pgfplotsset{
compat=1.11,
legend image code/.code={
\draw[mark repeat=2,mark phase=2]
plot coordinates {
(0cm,0cm)
(0.15cm,0cm)        %% default is (0.3cm,0cm)
(0.3cm,0cm)         %% default is (0.6cm,0cm)
};%
}
}
\begin{tikzpicture}
\begin{axis}[
        xmin=-20, xmax=20,
		ymin=0, ymax=1.1,
		grid=major, % activate major grid lines
                height=4cm,
                width=5.5cm,
		xlabel=$l$,
                ytick={1},
                yticklabel={1.0}
	]
	\draw[->,thick](0,0) -- (0,1);
        \addlegendimage{color=black}\addlegendentry{\scriptsize$f_{\mathcal{L}_{1,2}}(l)$}
        %\addlegendimage{color=white}\addlegendentry{\scriptsize$f_{H_2}(l)$}
\end{axis}
\end{tikzpicture}
%%% Local Variables:
%%% mode: latex
%%% TeX-master: "../main"
%%% End:
    \caption{$\EER=0.5$, $D_{KL}=0$.}
  \end{subfigure}
  \caption{Examples of Gaussian densities of calibrated LLRs under each hypothesis.}
   \label{fig:exampledensities}
 \end{figure}
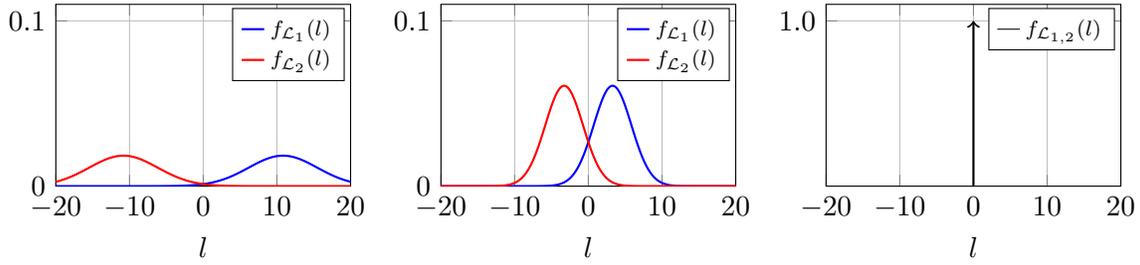

\paragraph{The parameter and the separability.}
The only parameter of the distributions in Theorem \ref{prop:distrib_LLR} is a scalar: the mean $\mu$ (or equivalently the variance $\sigma^2= 2 \mu$). This parameter can be expressed in terms of the separability between the two densities which can also be seen as the separabilities between the two classes in a classification context. In \cite{leeuwen13_interspeech}, the authors expressed the parameter in terms of the Equal-Error-Rate (EER). Here, we express the parameter in terms of the Kullback-Leibler divergence ($D_{KL}$):
\begin{equation}
  \label{eq:dklllr}
  \begin{aligned}
    D_{KL}\left(f_{\mathcal{L}_1} \Vert f_{\mathcal{L}_2}\right) &= \int_{-\infty}^{+\infty}f_{\mathcal{L}_1}\left( l\right) \log \frac{f_{\mathcal{L}_1}\left( l \right)}{f_{\mathcal{L}_2}\left( l \right)} dl,\\
                                                                 &= \int_{-\infty}^{+\infty}f_{\mathcal{L}_1}\left( l  \right)l dl~\text{because of the idempotence: $\log \frac{f_{\mathcal{L}_1}\left( l \right)}{f_{\mathcal{L}_2}\left( l \right)} = l$},\\
    &= \mathbb{E}_{l \sim \mathcal{N}(\mu,2\mu)} \left[ l \right] = \mu.
  \end{aligned}
\end{equation}
The mean is therefore equal to the Kullback-Leibler divergence. Note that since the two densities are Gaussian with the same variance, the $\Dkl$ is symmetric.
Figure \ref{fig:exampledensities} shows examples of Gaussian densities of the LLR under each hypothesis. When the mean increases, the variance and the separability increase. When the separability is 0, \ie~when $\EER=0.5$ and $D_{KL}=0$, both densities are a Dirac delta function at 0.

Theorem \ref{prop:distrib_LLR} gives a reference distribution for the LLRs to be calibrated, which has been applied especially for the calibration and the evaluation of of speaker verification systems. However, all the results presented so far are for the two hypotheses case: we have presented the concept of calibration for LLRs. The additivity of the Bayes' rule in its log-odds form, and the concepts of LLR and WOE, have been considered as not extensible to cases where more than two hypotheses are possible, see for instance Section 4.3 in \cite{jaynes}. We agree with this but only when log-ratios are treated one by one, independently from one another. In the next section, we will see how treating probability distributions and likelihood functions as compositional data provides an elegant manner for treating all log-ratios, at once, in a vector form; and how the additivity of the Bayes' rule is recovered generalizing therefore the concept of LLR.

\section{The Aitchison geometry of the probability-likelihood simplex}
\label{sec:compo}

In this section, we will see how treating discrete probability distributions and discrete likelihood functions as compositional data allows us to recover the additive form of the Bayes' rule, extending therefore the concept of LLR, in a vector form, to any number of hypotheses.

Compositional data carries relative information. A composition is a vector where each element \emph{describes a part of some whole} \citep{pawlowskymodeling} like vectors of proportions, concentrations, and probabilities. Compositional data analysis aims in treating such data by taking into account the compositional nature and structure of the data\footnote{For an overview of compositional data analysis, the reader can refer to \cite{pawlowskymodeling}.}. A $D$-part composition is a vector of $D$ non-zero positive real numbers that sum to a constant $k$. Each element of the vector is a part of the \emph{whole} $k$. The sample space of compositional data is known as the simplex:
\begin{equation}
  \label{eq:simplex}
    \mathcal{S}^D = \left\{ \bm{x} = [x_1, x_2,\dots x_{D}]^T \in \mathbb{R}^{*D}_{+} \Big| \sum_{i=1}^{D} x_i = k \right\}.
\end{equation}
This is how the simplex is usually defined. However, having the sum of each parts equal to a constant $k$ is not what really matter. Only the relative information between parts is important. We therefore introduce the following equivalence relation:
\begin{equation}
  \bm{x}, \bm{y}\in \mathbb{R}^{*D}_{+},~~~\bm{y} \sim \bm{x} \iff \exists c > 0,\text{ such that } \bm{y} = c \bm{x}.
\end{equation}
The simplex is then defined as the set of equivalent classes, \ie~as the quotient space:
\begin{equation}
  \label{eq:quotient}
  \mathcal{S}^D = \mathbb{R}^{*D}_{+} / \sim.
\end{equation}
This formulation allows us to see both the probability distributions and the likelihood functions as living in the same space: \emph{the probability simplex} as the set of equivalent classes (where $k=1$). Indeed, while the sum of the probabilities is equal to one, the likelihoods do not sum to a constant. However, since multiplying all the likelihoods by the same constant carries the same information\footnote{See the \emph{likelihood princible} \citep{berger1988likelihood}.}%the Bayes' rule: multiplying every likelihoods by the same constant results in the same posterior. In this way, scaled likelihood functions carry the same information. Hence their scale-invariance.}
, likelihood functions can be seen as compositional data too. Hence, from now on, when we discuss a likelihood function, as a vector $\bm{w} \in \mathbb{R}^{*D}_{+}$ of likelihoods, we refer to its equivalent that lives on the probability simplex.

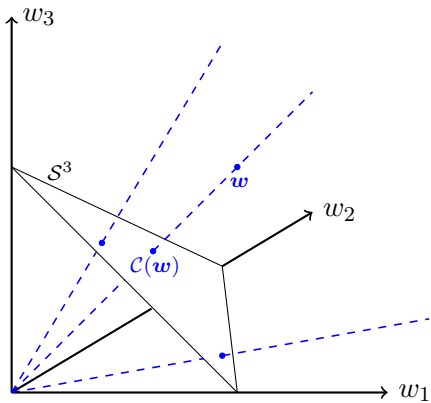
\begin{figure}[h!]
  \centering
  \begin{tikzpicture}[scale=1]

\draw [->,thick] (0,0) -- (0,5) node [right] {$w_3$};

\draw [->,thick] (0,0) -- (5,0) node [right] {$w_1$};

\draw [thick] (0,0) -- (1.86,1.116);

\draw [->,thick] (2.8,1.68) -- (4,2.4) node [right] {$w_2$};

\draw [] (0,3) -- (3,0);

\draw [] (0,3) -- (2.8,1.68);

\draw [] (3,0) -- (2.8,1.68);

\draw [] (0.65,2.95) node {\scriptsize $\mathcal{S}^3$};

\draw [blue,dashed,semithick] (0,0) -- (1.5,1.5);
%\draw [dotted,blue] (1.5,1.5) -- (1.88,1.88);
\draw [blue,dashed,semithick] (1.88,1.88) -- (4,4);
% \draw [blue] (1.88,1.88) node {\tiny\textbullet};
\node[minimum size=2pt,blue,fill,draw,circle,inner sep=0pt] () at (1.88,1.88){};
\draw [blue] (1.91,1.68) node {\scriptsize $\mathcal{C}(\bm{w})$};

% \draw [blue] (3,3) node {\tiny\textbullet};
\node[minimum size=2pt,blue,fill,draw,circle,inner sep=0pt] () at (3,3){};
\draw [blue] (3.03,2.8) node {\scriptsize $\bm{w}$};

\draw [blue,dashed,semithick] (0,0) -- (1.123,1.865);
%\draw [dotted,blue] (1.123,1.865) -- (1.2,1.99);
\draw [blue,dashed,semithick] (1.2,1.99) -- (2.83,4.7);
%\draw [blue] (1.2,1.99) node {\tiny\textbullet};
\node[minimum size=2pt,blue,fill,draw,circle,inner sep=0pt] () at (1.2,1.99){};

\draw [blue,dashed,semithick] (2.8,0.49) -- (5.55,0.98);
%\draw [dotted,blue] (2.55,0.45) -- (2.8,0.49);
\draw [blue,dashed,semithick] (0,0) -- (2.55,0.45);
\node[minimum size=2pt,blue,fill,draw,circle,inner sep=0pt] () at (2.8,0.49){};

\end{tikzpicture}

%%% Local Variables:
%%% mode: latex
%%% TeX-master: "../badoc"
%%% End:
  \caption{The probability-likelihood simplex and likelihood lines as equivalent classes. All likelihood functions that live on the same blue dashed ray are equivalent, and can be represented by the likelihood function that lives on the probability simplex.}
  \label{fig:equiv}
\end{figure}

We refer to this simplex as the \emph{probability-likelihood} simplex. Figure \ref{fig:equiv} illustrates likelihood equivalent classes. Likelihood lines (in dashed blue) go through the simplex $\mathcal{S}^3$. Within a line, all likelihood functions are equivalent and we take the likelihood function $\mathcal{C}(\bm{w})$ that lives on the simplex as the representative of this equivalent class.

This equivalence is materialized by the closure operator $\mathcal{C}$. Since only the relative information matter, scaling factors are irrelevant and a composition $\bm{x}$ is equivalent to its normalized version that lives on the simpex. The closure is defined for $k=1$ as:
\begin{equation}
  \label{eq:1}
  \mathcal{C}\left(\bm{x} \right) = \left[ \frac{x_1}{\lVert \bm{x} \rVert_1}, \frac{x_2}{\lVert \bm{x} \rVert_1} ,\dots \frac{x_D}{\lVert \bm{x} \rVert_1} \right]^T,
\end{equation}
where $\bm{x} \in \mathbb{R}_+^{*D}$ and $\lVert \bm{x} \rVert_1 = \sum_{i=1}^D \lvert x_i \rvert$. Therefore, any vector of positive real numbers can be mapped to its equivalent on the simplex using the closure.

To handle the scale-invariance nature of compositional data, John Aitchison introduced the use of log-ratios of components \citep{aitchison1982}. He defined several operations on the simplex leading to the \emph{Aitchison geometry of the simplex}.

\subsection{The Aitchison geometry of the simplex}
\label{sec:aitchison_geometry_simplex}

John Aitchison defined an internal operation called \emph{perturbation}, an external one called \emph{powering}, and an inner product:
\begin{itemize}
\item perturbation\footnote{Where ``$=_{\sim}\cdot$'' is ``$= \mathcal{C}\left(\cdot \right)$'' or also ``$\propto$''. Any scaling factor is indeed irrelevant under the equivalent relation $\sim$.}:
  \begin{equation}
    \label{eq:perturb}
        \bm{x}\oplus \bm{y} =_\sim [x_1y_1,\dots x_{D}y_{D}]^T,
    \end{equation}
    \item powering:
    \begin{equation}
        \alpha \odot \bm{x} =_\sim [x_{1}^{\alpha},\dots x_{D}^{\alpha}]^T,
    \end{equation}
    \item inner product:
    \begin{equation}
        \langle \bm{x},\bm{y} \rangle_{\mathcal{A}} = \frac{1}{2D}\sum_{i=1}^{D} \sum_{j=1}^{D} \log \frac{x_i}{x_j}\log \frac{y_i}{y_j}
    \end{equation}
\end{itemize}
where $\bm{x},\bm{y}\in \mathbb{R}^{*D}_{+}$ and $\alpha \in \mathbb{R}$. The perturbation and powering give to the simplex a $(D-1)$-dimensional vector space structure and the inner product makes it Euclidean. The corresponding norm and distance are:
\begin{equation}
    \lVert\bm{x}\rVert_{\mathcal{A}} =  \sqrt{\frac{1}{2N}\sum_{i=1}^{D} \sum_{j=1}^{D} \left( \log \frac{x_i}{x_j}\right)^2 },
\end{equation}
\begin{equation}
\begin{aligned}
    d_{\mathcal{A}}(\bm{x},\bm{y}) &= \lVert \bm{x}\ominus \bm{y} \rVert_{\mathcal{A}} = \lVert \bm{x}\oplus ((-1)\odot\bm{y}) \rVert_{\mathcal{A}}\\
    &= \sqrt{\frac{1}{2D} \sum_{i=1}^{D} \sum_{j=1}^{D} \left( \log \frac{x_i}{x_j} - \log \frac{y_i}{y_j} \right)^2 },
\end{aligned}
\end{equation}
respectively called the \emph{Aitchison norm} and the \emph{Aitchison distance}. This Euclidean vector space structure of the simplex is called the \emph{Aitchison geometry of the simplex}.

One can already notice the extensive use of log-ratios of parts. Hence the analogy with the log-odds of Section \ref{sec:llr}.

\subsection{The isometric-log-ratio transformation}

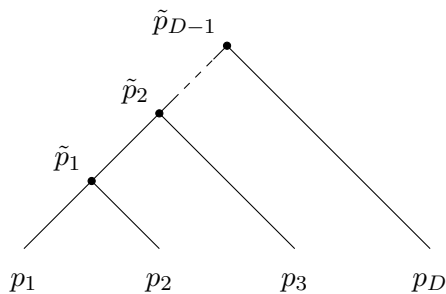
\begin{figure}
    \centering
    \begin{tikzpicture}[scale=1.8]
  \draw (0,0) -- (1.1,1.1);
  \draw[dashed] (1.1,1.1) -- (1.4,1.4);
  \draw (1.4,1.4) -- (1.5,1.5);
  \draw (0.5,0.5) -- (1,0);
  \draw (1,1) -- (2,0);
  \draw (1.5,1.5) -- (3,0);

  \filldraw[black] (0.5,0.5) circle (0.75pt) node[anchor=south east]{$\tilde{p}_{1}$};
  \filldraw[black] (1,1) circle (0.75pt) node[anchor=south east]{$\tilde{p}_{2}$};
  \filldraw[black] (1.5,1.5) circle (0.75pt) node[anchor=south east]{$\tilde{p}_{D-1}$};

  \draw[black] (0,-0.25) node{$p_1$};
  \draw[black] (1,-0.25) node{$p_2$};
  \draw[black] (2,-0.25) node{$p_3$};
  \draw[black] (3,-0.25) node{$p_{D}$};
\end{tikzpicture}

%%% Local Variables:
%%% mode: latex
%%% TeX-master: "../badoc"
%%% End:
    \caption{Bifurcating tree corresponding to the orthonormal basis of Equation \ref{eq:aitchisonbasis} obtained with the Gram-Schmidt procedure \citep{egozcue2003isometric}.}
    \label{fig:bifurctree}
  \end{figure}

Thanks to the Euclidean vector space structure of the simplex, the probability distributions and likelihood functions can be expressed in a Cartesian coordinate system using the Aitchison inner product and an orthonormal basis of the simplex. Let the set $\left\{\bm{e}^{(i)} \in \mathcal{S}^D, i\in\left\{ 1,\dots D-1 \right\}\right\}$ be such an \emph{Aitchison} orthonormal basis.
The elements of one basis obtained using the Gram-Schmidt procedure as in \cite{egozcue2003isometric} are defined for all $i \in \left\{1,\dots D-1 \right\}$ as follows:
  \begin{equation}
    \label{eq:aitchisonbasis}
     \bm{e}^{(i)} = \mathcal{C}\left(\left[ \underbrace{\exp \left( \sqrt{\frac{1}{i(i+1)}}\right),\dots \exp \left( \sqrt{\frac{1}{i(i+1)}} \right)}_{\text{The first $i$ elements}}, \exp \left( -\sqrt{\frac{i}{i+1}} \right),1, \dots 1 \right] \right).
  \end{equation}
  The Isometric-Log-Ratio (ILR) transformation \citep{egozcue2003isometric} allows to express a composition $\bm{p} \in \mathcal{S}^D$ in a Cartesian coordinate system by projecting it onto the basis as follows\footnote{We use the definite article \emph{the} to refer to the ILR transformation. This may suggest that there is only one ILR transformation, while there are as many ILR transformations as they are Aitchison orthonormal bases on the simplex \ie~an uncountable number. Along this article and without loss of generality, the expression \emph{``the ILR transformation''} will refer to the one with the orthonormal basis defined in Equation~\ref{eq:aitchisonbasis}. The use of this specific basis in no way excludes the general aspect of the following results since Aitchison orthonormal bases are related through unitary transformations \citep{egozcue2003isometric}.}:
\begin{equation}
    \tilde{\bm{p}}=\ilr(\bm{p}) = \left[\langle \bm{p}, \bm{e}^{(1)} \rangle_{\mathcal{A}} ,\dots \langle \bm{p}, \bm{e}^{(D-1)} \rangle_{\mathcal{A}}  \right]^T.
  \end{equation}
This defines an isometric isomorphism\footnote{An isometric isomorphism is an invertible mapping that preserves the distances.} between $\mathcal{S}^{D}$ and $\mathbb{R}^{D-1}$. Different bases could be used but the one presented above has a simple and intuitive recursive structure. The ILR transformation of the probability (or likelihood) vector results in a recursive grouping of the probabilities (or likelihoods) as illustrated by the bifurcation tree in Figure \ref{fig:bifurctree}.
  Considering a vector $\bm{p} = \left[ p_1,\dots p_D \right]^T \in \mathcal{S}^D$ and its ILR transformation $\tilde{\bm{p}} = \ilr \left( \bm{p} \right) = \left[ \tilde{p}_1,\dots \tilde{p}_{D-1} \right]^T \in \mathbb{R}^{D-1}$, each node of the tree corresponds to a component $\tilde{p}_i$ of $\tilde{\bm{p}}$. The first component compares the probabilities (or likelihoods) for the two first hypotheses. Each next component then recursively compares the probability (likelihood) for the next hypothesis with the probabilities (likelihoods) for the previous ones. The $i$th element $\tilde{p}_i$ of the ILR transformation of a composition $\bm{p}$ can be obtained with the following formula:
  
  \begin{equation}
    \label{eq:generalilrcomponent}
    \tilde{p}_i = \langle \bm{p}, \bm{e}^{(i)} \rangle_{\mathcal{A}} = \frac{1}{\sqrt{i\,(i+1)}} \log \left( \frac{\prod\limits_{j = 1}^{i}p_j}{(p_{i+1})^{i}} \right).
  \end{equation}
  An ILR component can therefore be interpreted as a weight (like a weight-of-evidence) comparing a probability (likelihood) with a group of other probabilities (likelihoods). When the probability (likelihood) for the $(i+1)$th hypothesis increases and the probabilities (likelihoods) for the hypotheses $H_{1\leq j\leq i}$ decrease, the score $\tilde{p}_i$ decreases. Therefore, a low $\tilde{p}_i$ goes in favor of the $(i+1)$th hypothesis against the hypotheses $H_{1\leq j\leq i}$ independently of the hypotheses $H_{i+2 \leq j \leq D}$.

  We saw that a composition carries relative rather than absolute information. The treatment of compositional data is therefore based on ratios and in particular on log-ratios. It is worth noting the natural analogy with log-odds and log-likelihood-ratios as presented in Section \ref{sec:llr} with the Bayes' rule. In the next section, we take a deeper dive into this analogy.

\subsection{The Bayes' rule as a vector translation}

The computation of the posterior probabilities through the Bayes' rule is the product of the prior probabilities with the likelihoods, normalized by $P(x)$ given by the law of total probability. This is exactly the perturbation (Equation \ref{eq:perturb}) of the prior probability vector by the likelihood vector (where the closure ensures the normalization). Let\footnote{Note that the Aitchison geometry is based on log-ratios such that a composition can not contain zeros. In the definition of the simplex in Equations \ref{eq:simplex} and \ref{eq:quotient}, the zeros are indeed excluded. Dealing with zeros has been problematic in compositional data analysis \citep{martin2003dealing, pawlowskymodeling}. However, banning probabilities equal to zero is not an issue for us. In the context of Bayesian updating, probabilities equal to zero are indeed not desirable. Since a posterior probability is proportional to the product of the prior probability and the likelihood, if the prior probability is zero, the posterior is necessarily equal to zero no matter which evidence is observed. If you have a prior probability equal to zero, it means that this is already certain for you that the corresponding hypothesis is false. No matter what evidence you observe or how someone is trying to convince you, your opinion about this hypothesis can not change. The rule excluding certainty in the prior belief, \ie~banning prior probabilities equal to 0 or 1, has been proposed by Dennis Lindley and is called \emph{the Cromwell's rule} \citep{lindley}. If you initially consider a set of possible hypotheses $\mathcal{H} = \{H_1,H_2,H_3\}$ and you finally proved \emph{logically} that $H_2$ is wrong, you must not assign a probability 0 to $H_2$. You must instead redefine your decision problem and your range of possibility as $\mathcal{H} = \{H_1,H_3\}$, \ie~everything that is for you \emph{``neither certainly true nor certainly false''} \citep{deFinetti}. However, the Cromwell's rule holds for probabilities only. There might be situations where the likelihood for a hypothesis of observing an evidence is zero. If such likelihood value is permitted in Bayesian updating, likelihood vectors treated as compositions can not contain zeros. In this case, the zeros have to be replaced. Zeros replacement strategies for compositional data are discussed by \cite{martin2003dealing}.}:
\begin{itemize}
\item $\bm{\pi} = \left[ P(H_1), P(H_2), \dots P(H_{D}) \right]^T \in \mathcal{S}^D$ be the vector of prior probabilities assigned to each hypothesis, \ie~the prior probability distribution;
\item $\bm{w} = \left[ P(x \mid H_1),P(x \mid H_2), \dots P(x \mid H_{D}) \right]^T \in \mathbb{R}_+^{*D}$ be the vector of likelihoods, \ie~the likelihood function;
\item $\bm{P} = \left[ P(H_1 \mid x), P(H_2 \mid x) \dots P(H_{D} \mid x) \right]^T \in \mathcal{S}^D$ be the posterior probability distribution.
\end{itemize}
The Bayes' rule is:
\begin{equation}
  \label{eq:bayesperturb}
  \begin{aligned}
    \forall i,~~ &P(H_i \mid x) = \frac{P(x \mid H_i) P(H_i)}{P(x)} = \frac{w_i \pi_i}{\displaystyle\sum_{j=1}^{D}w_j \pi_j},\\
    &\iff \bm{P} = \left[ \frac{w_1\pi_1}{\displaystyle\sum_{j=1}^{D}w_j \pi_j}, \frac{w_2\pi_2}{\displaystyle\sum_{j=1}^{D}w_j \pi_j}, \dots \frac{w_D\pi_D}{\displaystyle\sum_{j=1}^{D}w_j \pi_j} \right]^T,\\
    &\iff \bm{P} = \mathcal{C}\left( \left[ w_1 \pi_1, w_2 \pi_2, \dots w_D \pi_D \right] \right) = \bm{w} \oplus \bm{\pi}.
  \end{aligned}
\end{equation}
The Bayes' rule is the perturbation of the prior distribution by the likelihood function.
  
In the Isometric-Log-Ratio (ILR) space, \ie~the space $\mathbb{R}^{D-1}$ isometrically isomorphic to the simplex through the ILR transformation, a perturbation is a vector translation. Therefore, in the coordinate representation given by the ILR transformation, the Bayes' rule can be written as a vector translation of the prior by the likelihood function \citep{egozcue2018evidence}:
\begin{equation}
    \label{eq:ilrbayes}
    \begin{aligned}
    \bm{P} &= \bm{w} \oplus \bm{\pi},\\
    \ilr(\bm{P}) &= \ilr(\bm{w}) + \ilr(\bm{\pi}),\\
    \tilde{\bm{P}} &= \tilde{\bm{w}} + \tilde{\bm{\pi}}.
    \end{aligned}
  \end{equation}
  
  Just like the logit transformation in Equation \ref{eq:logit}, the ILR transformation allows us to write the Bayes' rule as a sum between a term that depends only on the prior probabilities and a term that depends only on the likelihoods. The ILR transformation is therefore the multidimensional, multiple hypotheses, or multiclass, extension of the logit transformation.

  In this way, the appealing additivity of the Bayes' rule is recovered. To be more precise, the likelihood function $\tilde{\bm{w}}$ translates a prior probability distribution $\tilde{\bm{\pi}}$ into a posterior distribution $\tilde{\bm{P}}$. Moreover, the ILR transformation in a two hypotheses case results in a one-element vector: the log-ratio of the probabilities (or likelihoods)\footnote{To be more precise, with the basis of Equation \ref{eq:aitchisonbasis}, there is a scaling factor $\frac{1}{\sqrt{2}}$.}; which is consistent with Equation \ref{eq:logit}.
\begin{figure}
    \centering
    \begin{tikzpicture}[scale=3.3]
        \draw[->,line width=0.4mm] (-1.3, 0) -- (1.3, 0) node[right] {\large $\tilde{p}_1$};
        \draw[->,line width=0.4mm] (0, -0.9) -- (0, 0.9) node[above] {\large $\tilde{p}_2$};
        \node[inner sep=0pt,label=left:$0.5$] () at (0,0.5){+};
        \node[inner sep=0pt,label=above:$0.5$] () at (0.5,0){+};

        \draw[dashed, blue,line width=0.6mm] (0, 0) -- (0, 0.9);
        \draw[dashed, blue,line width=0.6mm] (0, 0) -- (1.1258330249197706, -0.64999999999999975);
        \draw[dashed, blue,line width=0.6mm] (0, 0) -- (-1.1258330249197706, -0.6499999999999997);
        
        \node[orange,fill,draw,circle,inner sep=0pt, outer sep=0pt,minimum size=5pt,label=right:\large $\bm{\tilde{\pi}}$] (prior) at (0.25, 0.75) {};
        \node[purple,fill,draw,circle,inner sep=0pt, outer sep=0pt,minimum size=5pt,label=below:\large $\bm{\tilde{w}}$] (evidence) at (-0.5, -0.55){};
        \node[forestgreen,fill,draw,circle,inner sep=0pt, outer sep=0pt,minimum size=5pt,label=left:\large $\bm{\tilde{P}}$] (posterior) at (-0.25, 0.2){};
        
        \draw[->,line width=0.3mm] (0, 0) -- (evidence);
        \draw[->,line width=0.3mm] (prior) -- (posterior) ;

        \draw[->,Maroon,line width=0.8mm] (0,-0.5) -- (0,-0.8);
        \node[Maroon] at (-0.1, -0.8) {\large $H_{3}$};
        \draw[->,Maroon,line width=0.8mm] (0.433013, 0.25) -- (0.69282, 0.4);
        \node[Maroon] at (0.79282, 0.4) {\large $H_{1}$};
        \draw[->,Maroon,line width=0.8mm] (-0.433013, 0.25) -- (-0.69282, 0.4);
        \node[Maroon] at (-0.79282, 0.4) {\large $H_{2}$};

\end{tikzpicture}

%%% Local Variables:
%%% mode: latex
%%% TeX-master: "../badoc"
%%% End:
    \caption{Bayesian updating in the three-hypotheses ILR space. The posterior distribution $\tilde{\bm{P}}$ is the translation of the prior distribution $\tilde{\bm{\pi}}$ by the likelihood function $\tilde{\bm{w}}$. The red arrows indicate the directions that go in favor of one hypothesis against the two others. The dashed blue rays mark out the maximum probability (or likelihood) decision regions.}
    \label{fig:ilrbayes}
  \end{figure}
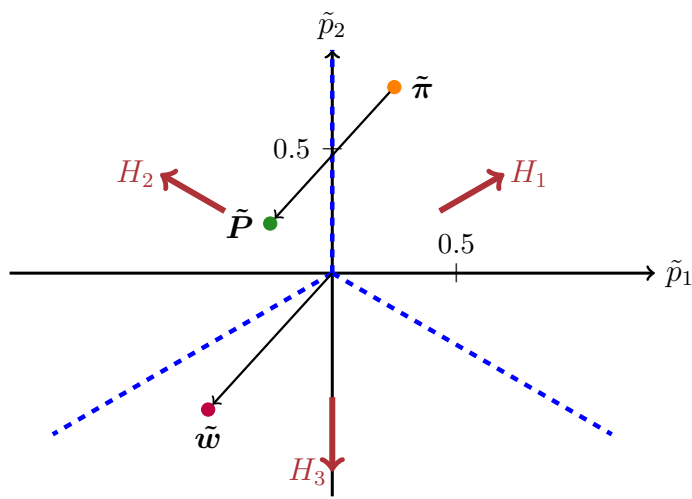
  
  Figure \ref{fig:ilrbayes} shows an example of a Bayesian updating in a three-hypotheses ILR space. The first component $\tilde{p}_1$ compares the probability (likelihood) for hypothesis $H_1$ with the probability (likelihood) for hypothesis $H_2$, and the second component $\tilde{p}_2$ compares the third probability (likelihood) against the two others as illustrated by the bifurcation tree in Figure \ref{fig:bifurctree}. Each red arrow shows the direction that goes in favor of one hypothesis against the two others. These three directions are naturally separated by an angle of $120^{\circ}$ \ie~one-third of $360^{\circ}$. The dashed blue rays mark out the maximum probability (likelihood) decision regions. Here, the simplex is 2-dimensional because we consider three possible hypotheses but keep in mind that for $D$ hypotheses, the simplex is $(D-1)$-dimensional. When there are only two possible hypotheses, the simplex is one-dimensional such that if one goes against one hypothesis, it necessarily goes in favor of the other, and we recover the situation of Section \ref{sec:llr}. With more hypotheses, the number of directions is now uncountable.

\subsection{Statistical evidence representation for multiple hypotheses: from the LLR to the Isometric-Log-Ratio Likelihood function}
\label{sec:multiplehypoevidence}

Recovering the additive form of the Bayes' rule---being the \emph{``basic property''} of the weight-of-evidence \citep{good1985weight}---the concept of LLR and weight-of-evidence can be now extended to cases with more than two hypotheses. The ILR transformation of the likelihood function---that we will now call ILRL for Isometric-Log-Ratio transformed Likelihood function---can be seen as a multidimensional extension of the LLR making it a good candidate for representing the statistical evidence when there are more than two possible hypotheses.

The direction of the ILRL informs which hypotheses the data may or may not support. The norm of the ILRL informs how strong. Like the absolute value of the LLR, the absolute value of each ILRL component gives the \emph{strength-of-the-evidence} in the support of one hypothesis against some others as shown by the bifurcation tree (Figure \ref{fig:bifurctree}). However, one basis does not provide all possible comparisons of hypotheses, this would have been redundant. If one wants to do a specific comparison, let's say for instance $p_3$ against $p_1$ only, he or she will have to use another basis resulting in a different bifurcation tree \citep{egozcue2005groups}, %which basically corresponds to a rotation and/or a permutation in $\mathbb{R}^{D-1}$,
or alternatively, to project the ILRL vector on the corresponding direction. The Aitchison norm of the likelihood function (\ie~the Euclidean norm of its ILR transformation) can be regarded as a \emph{global} \emph{strength-of-evidence} and is given by:
\begin{equation}
  \label{eq:normlikelihood}
  \lVert \bm{w}(x) \rVert_{\mathcal{A}} = \lVert \tilde{\bm{w}}(x) \rVert_2 = \sqrt{\frac{1}{D} \sum_{i=1}^D \sum_{j=i+1}^D \left (\log \frac{P(x \mid H_i)}{P(x \mid H_j)} \right)^2}.
\end{equation}
This is proportional to the square root of the sum of the square of all possible LLR. This informs how much the evidence $x$ is changing the belief, \ie~how far the posterior distribution is from the prior distribution, regardless of any direction: this is the Aitchison distance between the posterior distribution and the prior distribution.

\section{Calibrated likelihood functions on the simplex}
\label{sec:ILRL}

In Section \ref{sec:calibLLR}, we presented the idempotence property of calibrated LLRs and the constraint on their distribution. In Section \ref{sec:compo}, by introducing and completing elements from \cite{egozcue2003isometric}, and \cite{egozcue2018evidence}, we saw how the Aitchison geometry of the probability simplex allows us to extend the concept of LLR and weight-of-evidence, in a vector form, to any number of hypotheses.
In the current section, we extend the definition of calibration and the idempotence property to the multiple hypotheses extension of the LLR: the Isometric-Log-Ratio Likelihood function (ILRL). We also show how the constraint on the distribution of calibrated LLRs generalizes to the ILRLs. This gives a reference distribution for the likelihood function to be calibrated and to properly represent the statistical evidence in a multiple hypotheses and multiclass context.

Let's consider the prior $\bm{\pi}$, the likelihood function $\bm{w}_{\theta}$, and the posterior $\bm{P}_\theta$ as compositions:
\begin{equation}
  \begin{aligned}
    \bm{\pi} &= \left[ P(H_1),\dots P(H_D) \right]^T \in \mathcal{S}^D,\\
    \bm{w}_{\theta}(\cdot) &= \left[ f_{\theta_{\mathcal{X}_1}}(\cdot),\dots f_{\theta_{\mathcal{X}_D}}(\cdot) \right]^T \in \mathbb{R}^{*D}_+,\\
    \bm{P}_{\theta}(\cdot) &= \left[ P_{\theta}(H_1 \mid \cdot),\dots P_{\theta}(H_D \mid \cdot) \right]^T \in \mathcal{S}^D,\\
  \end{aligned}
\end{equation}
and their isometric-log-ratio transform:
\begin{equation}
  \begin{aligned}
    \tilde{\bm{\pi}} &= \ilr\left(\bm{\pi}\right) \in \mathbb{R}^{D-1},\\
    \bm{l}_\theta(\cdot) &= \tilde{\bm{w}}_{\theta}(\cdot) = \ilr\left(\bm{w}_\theta(\cdot)\right) \in \mathbb{R}^{D-1},\\
    \tilde{\bm{P}}_{\theta}(\cdot) &= \ilr\left(\bm{P}_{\theta}(\cdot)\right) \in \mathbb{R}^{D-1},
  \end{aligned}
\end{equation}
where $f_{\theta_{\mathcal{X}_i}}$ is the probability density function of the statistical model for the data under hypothesis $H_i$.

Let $\mathcal{L}$ be a set of (isometric-log-ratio transformed) likelihood functions (ILRLs). Starting from the definition of calibration like in Equation \ref{eq:calibprobllr} and applying the isometric-log-ratio transformation we get:

$\forall \bm{l}_\theta \in \mathcal{L} = \left\{ \bm{l}_\theta(x) \mid x \in \mathcal{X} \right\},$
\begin{equation}
  \begin{aligned}
    \mathbb{P}(H_i\mid \bm{l}_\theta)=q_i(x) =P_{\theta}(H_i \mid x)~\forall i \in \{1,\dots D\} &\iff \left[ \mathbb{P}(H_1\mid \bm{l}_\theta),\dots \mathbb{P}(H_D\mid \bm{l}_\theta) \right]\\
    &~~~~~~~~~= \left[ P_{\theta}(H_1\mid x),\dots P_{\theta}(H_D\mid x) \right],\\
    &\iff \bm{l}_{\mathcal{L}}(\bm{l}_\theta) + \tilde{\bm{\pi}} = \bm{l}_{\theta}(x) + \tilde{\bm{\pi}}\\
    &\iff \bm{l}_{\mathcal{L}}(\bm{l}_\theta) = \bm{l}_{\theta}.                                     \end{aligned}
\end{equation}
where $\bm{l}_{\mathcal{L}}(\bm{l}_\theta) = \ilr \left( \left[ f_{\mathcal{L}_1}(\bm{l}_\theta), \dots f_{\mathcal{L}_D}(\bm{l}_\theta) \right]^T  \right)$, and where $f_{\mathcal{L}_i}$  refers to the probability density function of the distribution, over $\mathbb{R}^{D-1}$, of $\bm{l}_{\theta}$ under hypothesis $H_i$. This is the same reasoning as in Equations \ref{eq:calibprobllr}-\ref{eq:idemp} but in a multiple hypothesis and multidimensional setting. Here LLRs are replaced by ILRLs, and the logit transformation is the isometric-log-ratio transformation.

Hence the extension of the idempotence property for calibrated likelihood functions that can be read as:
\begin{center}
  \emph{``The ILRL of the ILRL is the ILRL''},
\end{center}
or simply:
\begin{center}
  \emph{``the likelihood function of the likelihood function is the likelihood function''}
\end{center}
since the ILRL and the likelihood function are isomorphic through the ILR transformation. Hence the following definition of calibrated likelihood functions:
\begin{definition}
  A set $\mathcal{L}$ of (isometric-log-ratio) likelihood functions is perfectly calibrated if they are \emph{idempotent}:
  \begin{equation}
    \forall \bm{l}_{\theta} \in \mathcal{L},~~~\bm{l}_{\mathcal{L}}(\bm{l}_{\theta}) = \bm{l}_{\theta}.
  \end{equation}
    \label{def:defcalibilrl}
\end{definition}

Exactly like in Theorem \ref{prop:distrib_LLR} with the LLR, we will see how the idempotence property leads to a constraint on the distributions of the likelihood functions.

\subsection{The distribution of calibrated ILRLs}
\label{sec:distribILRL}

Let $\bm{A} \in \mathcal{M}_{D-1,D-1}(\mathbb{R})$ be a real square matrix and $\bm{B} \in \mathcal{M}_{D-1,(D-1)^2}(\mathbb{R})$ be a real block matrix. These matrices are fixed and defined by the used Aitchison basis. See Appendix \ref{ap:distribILRL} for more details.

The idempotence property of the ILRLs leads to the following constraint on their distributions:
\begin{theorem}
  \label{prop:distribILRL}
  If $\bm{l} \mid H_1 \sim \mathcal{N}(\bm{\mu}_1,\bm{\Sigma})$, then $\forall i \in \left\{2,\dots D \right\}$, $~\bm{l} \mid H_i \sim \mathcal{N}(\bm{\mu}_i,\bm{\Sigma})$,
\textit{where}
\begin{equation*}
    \bm{\mu}_i = \bm{\mu}_1 - \bm{\Sigma} \bm{a}_{i-1} - \sum_{j=1}^{i-2} \frac{1}{j+1} \bm{\Sigma} \bm{a}_j,
\end{equation*}
and $\bm{\mu}_1 = \bm{A}^{-1}\bm{B} \vect(\bm{\Sigma})$, where the $(D-1)^2$-dimensional vector $\vect(\bm{\Sigma})$ is the vectorization of the covariance matrix~$\bm{\Sigma}$, and $\forall i \in \left\{1,\dots D-1 \right\}$,  $\bm{a}_i = \sqrt{\frac{i+1}{i}} \bm{e}_i$ where $\bm{e}_i$ is the $i$th vector of the standard canonical basis of $\mathbb{R}^{D-1}$.
\end{theorem}

In other words, if under one hypothesis, the likelihood function is normally distributed on the Aitchison simplex\footnote{The multivariate normal distribution that appears with the ILR coordinate representation is also known as the \emph{additive logistic-normal distribution}, the \emph{logistic-normal distribution} \citep{aitchison1980}, or simply the \emph{normal distribution on the simplex} \citep{pawlowskymodeling}.}, it is also normally distributed for all the other hypotheses, with the same covariance matrix and the means are entirely determined by the covariance matrix. The proof of this result is given in Appendix \ref{ap:distribILRL}. This is a proof by induction where each density is recursively determined thanks to the recursive form of the bifurcation tree.

Since $\bm{\mu}_1 = \bm{A}^{-1}\bm{B} \vect(\bm{\Sigma})$, the only parameter of the distributions is $\bm{\Sigma}$ which is a $(D-1)\times(D-1)$ symmetric positive definite matrix%\footnote{In general, covariance matrices are symmetric positive semi-definite but here, we restrict to symmetric positive definite because we need $\bm{\Sigma}^{-1}$.}
. Therefore, it corresponds to $\frac{D(D-1)}{2} = \binom{D}{2}$ scalar parameters which is equal to the number of pairs of hypotheses. In the next paragraph, we will see how these parameters can be expressed in terms of the Kullback-Leibler divergences between each density. This relation between the mean vector and the covariance matrix, and how it is related to the divergences, extends what was presented in the two hypotheses case in Section \ref{sec:distribLLR} where $\mu = \frac{\sigma^2}{2}$, and is equal to the Kullback-Leibler divergence between the densities of the LLRs.

\paragraph{The covariance matrix of the ILRL distribution and the divergences.}
The covariance matrix $\bm{\Sigma}$, \ie~the parameter of the Gaussian ILRL distributions, can be expressed in terms of the Kullback-Leibler divergences ($D_{KL}$) between each density. In a classification context, these divergences can be seen as the between class separabilities.

Let $\bm{\Delta} = \{ d_{i,j} \}_{1 \leq i,j \leq D} \in \mathcal{M}_{D \times D} (\mathbb{R}_+)$, where $ d_{i,j} = D_{KL} \left( f_{\mathcal{L}_i} \Vert f_{\mathcal{L}_j} \right)$, be the matrix of Kullback-Leibler divergences between each density. Since the densities are multivariate Gaussian with the same covariance matrix, the divergences are symmetric and the $\bm{\Delta}$ is therefore a symmetric matrix. Since $d_{i,j} = 0$ for $i=j$, the $D$ diagonal elements are $0$. Therefore, only $\frac{D(D-1)}{2} = \binom{D}{2}$ degrees of freedom remain for the matrix of divergences which is the same as for the covariance matrix $\bm{\Sigma}$. The divergences can be expressed from the covariance matrix as follow:
\begin{equation}
  \label{eq:fromcov2div}
  \vech_{\neg\smallsetminus}(\bm{\Delta}) = \bm{M} \vech(\bm{\Sigma}),
\end{equation}
where $\vech$ is the half-vectorization of a matrix, $\vech_{\neg\smallsetminus}$ is the half-vectorization without the diagonal elements, and $\bm{M} \in \mathcal{M}_{\frac{D(D-1)}{2} \times \frac{D(D-1)}{2}}(\mathbb{R})$ is a real square matrix. See Appendix \ref{ap:covdiv} for more details. The divergences can therefore be computed from the parameter $\bm{\Sigma}$\footnote{Unfortunately, we did not prove that $\bm{M}$ is invertible. Assuming this is the case, the covariances can be expressed in terms of the divergences as: $\vech(\bm{\Sigma}) = \bm{M}^{-1}\vech_{\neg\smallsetminus}(\bm{\Delta})$.}.

\begin{figure}
  \centering
  \begin{subfigure}[b]{1.0\linewidth}
      \centering
  \includegraphics[scale=0.5]{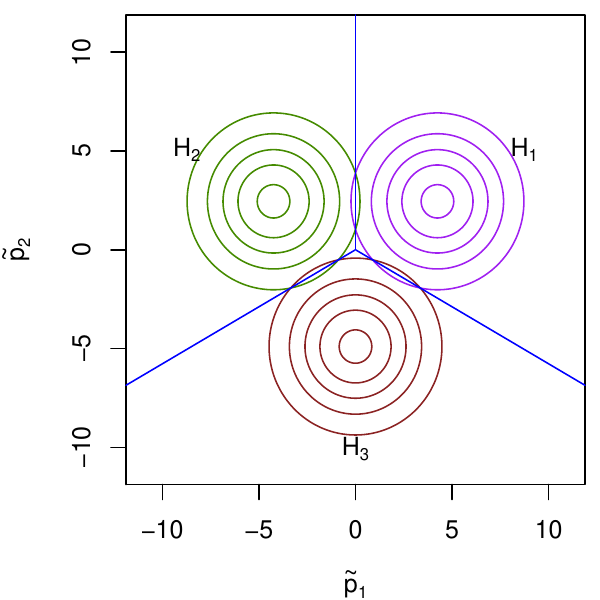}\includegraphics[scale=0.57]{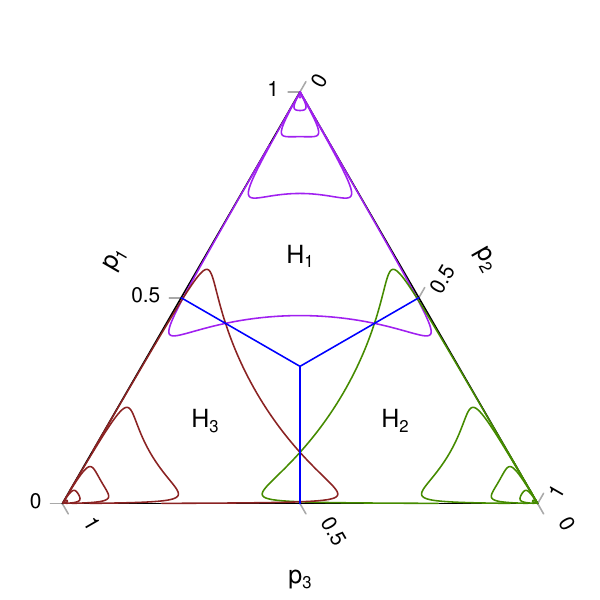}
  \caption{$\EER_{1,2} = \EER_{2,3} = \EER_{1,3} \approx 4.16\%$, $d_{1,2} = d_{2,3} = d_{1,3} = 6  $.}
  \label{fig:isotropgauss}
  \end{subfigure}\\
  \vspace{-0.8cm}
  \begin{subfigure}[b]{1.0\linewidth}
      \centering
  \includegraphics[scale=0.5]{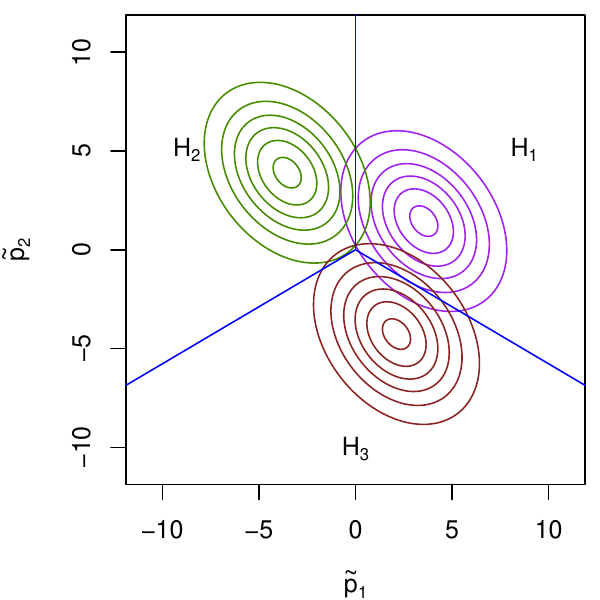}\includegraphics[scale=0.57]{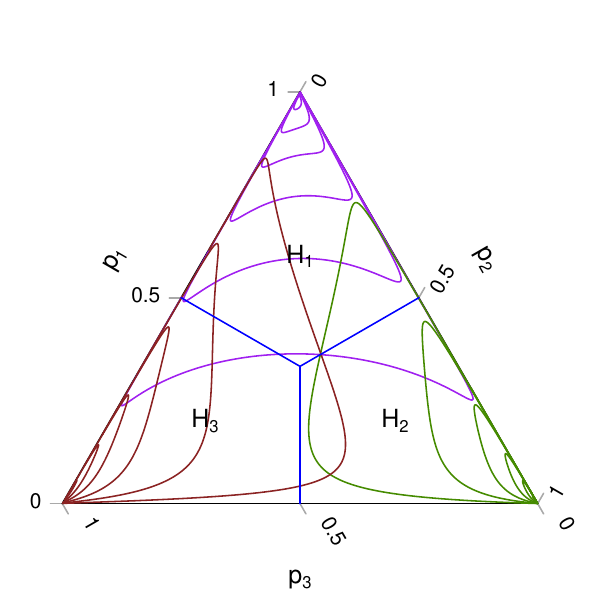}
  \caption{$\EER_{1,2} \approx 5.69\%$, $d_{1,2} = 5$; $\EER_{2,3}\approx 3.07\%$, $d_{2,3} =7 $; $\EER_{1,3} \approx 7.86\% $, $d_{1,3} = 4$.} 
\end{subfigure}\\
  \vspace{-0.8cm}
  \begin{subfigure}[b]{1.0\linewidth}
      \centering
    \includegraphics[scale=0.5]{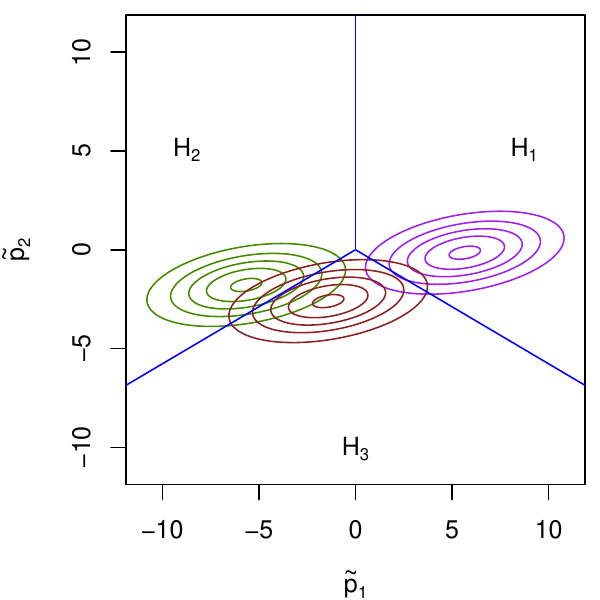}\includegraphics[scale=0.57]{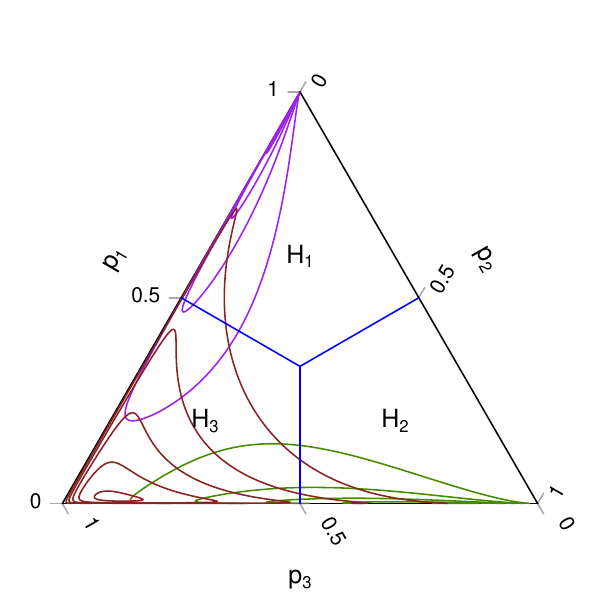}
    \caption{$\EER_{1,2} \approx 2.28\% $, $d_{1,2} = 8 $; $\EER_{2,3} \approx 15.87\% $, $d_{2,3} =2 $; $\EER_{1,3} \approx 7.86\% $, $d_{1,3}=4 $.}
  \end{subfigure}
  \caption{Few contours of Gaussian densities of the likelihood function in a three-hypotheses case. They are parameterized by a shared covariance matrix that can be expressed in terms of the three separabilities between each density. The densities on the ILR space (left) are with respect to the Lebesgue measure while the densities on the simplex (right) are with respect to the Aitchison measure \citep{mateu2011principle}.}
  \label{fig:conddensILRL}
\end{figure}

Figure \ref{fig:conddensILRL} shows a few examples of densities of the ILRL under each of the three hypotheses  $H_1$, $H_2$, and $H_3$. The blue rays mark out the maximum probability (likelihood) decision regions. The figures on the right side show the densities on a ternary plot and the figures on the left side show the densities in $\mathbb{R}^2$ with the ILR coordinate representation. The parameters of the densities are linked and constrained according to Theorem \ref{prop:distribILRL}. Their parameters can be expressed in terms of the three divergences as explained above. When the separabilities between each density are all the same, the covariance matrix is isotropic and the means are equidistant. When they are different, the densities stretch accordingly. When two classes get closer, the densities crush on the corresponding decision boundary and when the separability between all classes goes to 0, the densities collapse at $\bm{0}$ and tend to be a Dirac delta function.

In this Section, we extended the idempotence property and the definition of calibrated LLRs to their multiple hypotheses and multiclass counterpart: the isometric-log-ratio tranformed likelihood function (ILRL). Because likelihood functions and ILRLs are isomorph, those results can naturally be pulled back to probability-likelihood simplex.

As for the LLR, the idempotence leads to a constraint on the distribution of calibrated likelihood functions. We showed that if, under one hypothesis (or class), the likelihood function is normally distributed on the simplex, It is also normally distributed for the other hypotheses with the same covariance matrix and the means are entirely defined by this matrix. This give a reference distribution for the likelihood functions to be calibrated and to properly represent the statistical evidence.

These properties are in fact the generalization, to the multiple hypotheses case ($D\geq 2$), of results that are well-known for the binary case ($D=2$), and that can be widely found in the literature from different fields: for the weight-of-evidence since the 40s \citep{turingstat,good1985weight}, in the context of signal detectability since the 50s \citep{peterson1954,birdsall1966theory}, and in the context of LLRs calibration in forensic identification since the 2000s \citep{leeuwen13_interspeech,meester_slooten_2021}.
  
In the next Section, we provide one application of these results by presenting how the discriminant space of a discriminant analysis can be designed to form the space of calibrated ILR transformed likelihood functions.

\section{Application: Compositional discriminant analysis}
\label{sec:discrim_ILRL}

In \cite{Noe22Icassp}, the authors presented how the idempotence property and its constraint on the distributions of the LLR can be used to design a non-linear discriminant analysis where the discriminant component forms a calibrated LLR\footnote{This has been presented in the context of privacy in speech technology. In the discriminant space, the LLR can be set to zero for hiding the evidence related to a binary attribute, in accordance with the concept of \emph{perfect secrecy} \citep{shannon,nautsch2020}. See \cite{noe2023representing} for more details.}. Being based on results that were known only for the LLR, the approach was naturally limited to the two-classes case. However, with the results presented in Section \ref{sec:ILRL}, the discriminant analysis can now be defined for any number of classes. We call this approach: \emph{Compositional discriminant analysis} (CDA), not to be confused with discriminant analysis that aims in modeling compositional data as discussed for instance in \cite{filzmoser2011discriminant} or in Section 8.4 of \cite{pawlowskymodeling}. The compositional nature of the CDA is on the treatment of the produced vector of likelihoods. In a nutshell, the idea is to design the discriminant space in accordance with Theorem \ref{prop:distribILRL} such that the discriminant components form a calibrated isometric-log-ratio transformed likelihood function (ILRL). This approach is presented in details in this section\footnote{This has been presented as a poster at CoDaWork2024, the 10th International Workshop on Compositional Data Analysis \citep{noe2024compositional}.}.

Let's go back to the linear discriminant analysis (LDA). For a two-classes case, as discussed in the example of Section \ref{sec:distribLLR}, a LLR is computed (Equation \ref{eq:lda}). This is the same as projecting the feature vector onto the discriminant direction that minimizes the within-class variability and maximizes the between-class variability. However, since the observations are not necessarily normally distributed, violating therefore the Gaussian assumption, the computed LLRs may not be well-calibrated. Other discriminant analysis approaches relax the assumptions on the distribution of the data. The quadratic discriminant analysis (QDA) also assumes the features to be normally distributed for each class, but without the shared covariance assumption. The LLR computation becomes a quadratic form of the data, and the resulting discriminant function is non-linear. However, QDA is still based on Gaussian assumptions and the discriminant function is not necessarily invertible contrary to the LDA's mapping \citep{dimredhastie,bishop}\footnote{Having an invertible mapping is not required for applications that focus only on classification. However, some applications may require an invertible mapping, for doing data transformation by manipulating the discriminant space like in \cite{Noe22Icassp} or like the interpolation example of Section \ref{sec:mnist}; or for generation.}. Some approaches make no explicit assumption on the distribution of the data. In \cite{deepldamatt}, the authors proposed what they called ``DeepLDA'' where the general LDA eigenvalue problem is solved on the top of an artificial neural network. However, the approach is fully discriminative and looses the generative and statistical modeling nature of the LDA. In this sense, and since our work is set in the realm of generative classifiers, we do not consider this approach as a non-linear version of the LDA. Generalized \citep{andreStuh} and kernel-based \citep{mika1999kernelda} discriminant analysis are good candidates for generalizing the LDA in a non-linear manner. However, like the QDA, they do not have a trivial inverse mapping from the discriminant space back to the feature space \citep{NIPS2003_ac1ad983}. Works like \cite{izmailov2020semi} and \cite{ardizzone2020nips} propose generative classifiers by modeling the class-conditional distributions in the features space using normalizing flow (NF)\footnote{Normalizing flow (NF) is a family of invertible neural networks that learn a diffeomorphism between the feature space and a \emph{base} space. Some literature on NF uses the term \emph{latent} rather than \emph{base}. We were also doing so in~\cite{Noe22Icassp}, however, agreeing with the argument proposed in~\cite{papamakarios}, we have adopted the term \emph{base}.}. In the base space, the class-conditional distributions are chosen to be multivariate normal. However, the choice of the normal distributions' parameters in the base space is arbitrary, and contrary to the CDA presented below, the resulting components in the base space can not be interpreted as a calibrated LLR or ILRL.

\subsection{Proposed compositional discriminant analysis}
\label{sec:discrimILRL}

Let's consider the set $\mathcal{C} = \{ C_1, C_2,\dots C_D \}$ of $D$ classes and an observed vector $\bm{x}$ which belongs to one of these classes. The proposed discriminant analysis, named compositional discriminant analysis (CDA), is a generative classifier that models the distribution of the data under each class by learning an invertible and differentiable mapping between the feature space and a base space in which the class-conditional distributions are known. The class-conditional distributions in the base space are designed according to the idempotence property constraint of Theorem \ref{prop:distribILRL}. In this way, the mapping transforms the observed vectors into a same-dimensional base space where the first $D-1$ dimensions form the isometric-log-ratio transformed likelihood function (ILRL), and the other dimensions form the residual meaning ``everything in the data that is independent of the class variable''.

Let's introduce some notations. Let:
\begin{itemize}
\item $\mathcal{X} \subset \mathbb{R}^d$ be the $d$-dimensional feature space,
\item $\bm{l}(\bm{x}) \in \mathcal{L} \subset \mathbb{R}^{D-1}$ be the ILRL of an observation $\bm{x} \in \mathcal{X}$,
\item $\bm{r}(\bm{x}) = \left[r_1(\bm{x}),r_2(\bm{x}),\dots r_{d-D+1}(\bm{x}) \right]^T \in \mathcal{R} \subset \mathbb{R}^{d-D+1}$ be the residual of $\bm{x}$.
\end{itemize}
We want to find a diffeomorphism that maps the data from the feature space to the base space $\mathcal{Z} = \mathcal{L}\oplus\mathcal{R}$ in which the first $D-1$ dimensions form a calibrated ILRL, representing therefore the statistical evidence about the classes, while the other dimensions form the residual.

\subsubsection{Class-conditional distributions in the base space}
\label{sec:classconddensitiescompo}

In order to properly represent the statistical evidence, we want the first $D-1$ dimensions of the base space to form a calibrated ILRL. The class-conditional distributions in the base space are therefore chosen according to the idempotence property constraint of Theorem \ref{prop:distribILRL}:
\begin{equation}
  \label{eq:classcondbase}
   \forall i \in \left\{1,\dots D \right\},~\bm{z} \mid C_i \sim \mathcal{N}\left(\bm{m}_i,\bm{C}\right),
 \end{equation}
 where:

 \begin{itemize}
 \item  $\bm{\Sigma}$ is a $(D-1)\times(D-1)$ symmetric positive definite matrix, and is the only parameter of the distributions in the base space,
 \item the means $\bm{m}_i \in \mathcal{Z} \subset \mathbb{R}^d$ are the concatenation of $\bm{\mu}_i \in \mathcal{L}$ and the $(d-D+1)$-dimensional zero vector. $\bm{\mu}_i$ is defined according to Theorem \ref{prop:distribILRL} and its expression is given by:
\begin{equation}
  \begin{aligned}
    \label{eq:genmean2}
    \forall i \in \left\{1,\dots D \right\},~\bm{\mu}_i = \bm{A}^{-1}\bm{B} \vect(\bm{\Sigma}) - \bm{\Sigma} \bm{a}_{i-1} - \sum_{j=1}^{i-2} \frac{1}{j+1} \bm{\Sigma} \bm{a}_j,
    \end{aligned}
  \end{equation}
  where $\bm{A} \in \mathcal{M}_{D-1,D-1}(\mathbb{R})$ and $\bm{B} \in \mathcal{M}_{D-1, (D-1)^2}(\mathbb{R})$ are fixed matrices, $\bm{a}_i = \sqrt{\frac{i+1}{i}} \bm{e}_i$ where $\bm{e}_i$ is the $i$th vector of the standard canonical basis of $\mathbb{R}^{D-1}$ and $\bm{a}_0 = \bm{0}$ is the $(D-1)$-dimensional zero vector (see Appendix \ref{ap:distribILRL} for more details),
   \item  the covariance matrix $\bm{C}$ is the following block matrix:
\begin{equation}
    \label{eq:classcond}
    \bm{C} = 
    \begin{bmatrix}
    \bm{\Sigma} & \bm{0}_{D-1,d-D+1}\\
    \bm{0}_{d-D+1,D-1} & \bm{I}_{d-D+1}
    \end{bmatrix},
\end{equation}
where $\bm{I}_K$ is the $K\times K$ identity matrix and $\bm{0}_{K,L}$ is the $K \times L$ zero matrix.
\end{itemize}
In this way, the $D-1$ first dimensions are distributed according to Theorem \ref{prop:distribILRL} and the remaining dimensions are normally distributed with a zero mean and an identity covariance regardless of the class variable.

\begin{lemma}
  With $\bm{z} \mid C_i \sim \mathcal{N}\left(\bm{m}_i,\bm{C}\right)~\forall i \in \left\{1,\dots D \right\}$, as described above, the first $D-1$ dimensions of $\bm{z}$ form its isometric-log-ratio transformed likelihood function:
  \begin{equation}
    \left[ z_1,\dots z_{D-1} \right] = \ilr \left( \left[ f_{\mathcal{Z}_1}(\bm{z}),\dots f_{\mathcal{Z}_D}(\bm{z}) \right] \right).
  \end{equation}
  \label{prop:ilrlzdims}
\end{lemma}
See Appendix \ref{ap:ilrlzdims} for a proof.

\subsubsection{Diffeomorphism between the feature space and the base space}
\label{sec:diffeo}

Let $g^{-1} : \mathcal{X} \mapsto \mathcal{Z}$ be a diffeomorphism that maps the data into the base space such that\footnote{$g : \mathcal{Z} \mapsto \mathcal{X}$, where $g$ stands for \emph{generator}.}:
\begin{equation}
  \label{eq:2}
  \forall i \in \left\{1,\dots D \right\},~g^{-1}(\bm{x}) \mid C_i \sim \mathcal{N}\left(\bm{m}_i,\bm{C}\right)
\end{equation}

\begin{theorem}
  With $g^{-1}(\bm{x}) \mid C_i \sim \mathcal{N}\left(\bm{m}_i,\bm{C}\right)~\forall i \in \left\{1,\dots D \right\}$, the first $D-1$ dimensions of $\bm{z} = g^{-1}(\bm{x})$ form the isometric-log-ratio transformed likelihood function of $\bm{x}$:
  \begin{equation}
    \left[ z_1,\dots z_{D-1} \right] = \ilr \left( \left[ f_{{\mathcal{X}_1}}(\bm{x}),\dots f_{{\mathcal{X}_D}}(\bm{x}) \right] \right).
  \end{equation}
  \label{th:ilrx}
\end{theorem}
This means that the $D-1$ first dimension in the base space form the isometric-log-ratio transformed likelihood function of the data. Given Lemma \ref{prop:ilrlzdims}, the proof is straightforward since the likelihood function of $\bm{x}$ and the likelihood function of $\bm{z}$ are the equivalent. Indeed, they are proportional, where the Jacobian determinant of the mapping is the scaling factor.

With the distributions in the base space defined by Equation \ref{eq:classcondbase} and thanks to Theorem \ref{th:ilrx}, the first $D-1$ dimensions of $\bm{z}$ represent the statistical evidence in $\bm{x}$ about the classes in the form of a ILRL, while the other dimensions form the residual normally distributed with a zero mean vector and an identity covariance matrix regardless of the class.

The diffeomorphism $g$ can be learned through Normalizing Flow (NF) \citep{papamakarios}\footnote{In our experiments we used the RealNVP architecture \citep{dinh2017}.}. Let $\mathcal{D} = \left\{ (\bm{x}^{(1)}, c^{(1)}) ,\dots (\bm{x}^{(N)}, c^{(N)}) \right\}$ be a training set of observed feature vectors with their corresponding class. The parameters $\theta_g$ of $g$, and $\bm{\Sigma}$, are learned by maximizing the log-likelihood of the data:
\begin{equation}
  \log f \left( \mathcal{D} ; \theta_g, \bm{\Sigma} \right) = \sum_{i=1}^{D} \left( \sum_{(\bm{x},c) \in \mathcal{D} \mid c = C_i} \log \left( f_{\mathcal{Z}_i}(\bm{z}; \bm{\Sigma}) \left| \det \left( \frac{\partial \bm{x}}{\partial \bm{z}} \right) \right|^{-1} \right) \right),
\end{equation}
where the densities for $\bm{x}$ are computed through the densities for $\bm{z}$ and the change of variable formula. In our experiments, $\theta_g$ and $\bm{\Sigma}$ are learned through negative log-likelihood minimization with automatic differentiation and gradient descent. Regarding the initialization of $\bm{\Sigma}$, see Appendix \ref{ap:initSigma}.
\paragraph{Remark:} We have not made any explicit assumption on the distribution of the data. However, this does not mean there is no assumption at all. The use of the CDA implicitly assumes the existence of a diffeomorphism that would transform the distribution of the data into the target Gaussians and that the NF is flexible enough to reach this diffeomorphism.

\subsubsection{Regarding the interpretability of the compositional discriminant analysis}
\label{sec:about_interp}

The proposed discriminant analysis maps the data into a space where the first $D-1$ dimensions are discriminant and form the ILRL of the observation. With the standard LDA, the $D-1$ dimensions given by the non-zero-eigenvalue eigenvectors of the matrix $\bm{\Sigma}_W^{-1}\bm{\Sigma}_B$, where $\bm{\Sigma}_W$ is the shared within-class covariance matrix and $\bm{\Sigma}_B$ is the between-class covariance matrix, are also the discriminant ones. They are usually sorted in the descending order of the eigenvalues which inform how much each direction is discriminant.

In the CDA, the discriminant dimensions are not sorted according to their discriminant power. However, thanks to the compositional nature of the base space, each dimension is instead opposing a class with a group of classes in an intuitive recursive manner as illustrated by the used bifurcation tree (we use here the one in Figure \ref{fig:bifurctree}). The discriminations between the classes are given by the parameter $\bm{\Sigma}$ as discussed in Section \ref{sec:distribILRL}.

Moreover, since the densities of the ILRL in the base space are designed to respect the idempotence constraint of Theorem \ref{prop:distribILRL}, the approach tends to produce a set of ILRLs that is well-calibrated. The resulting classifier can therefore be used for uncertainty-aware predictions avoiding under or overconfident decisions. In addition, the first $D-1$ dimensions of the base space benefit from the Euclidean vector space structure of the Aitchison geometry, allowing distance measure, posterior probability distribution computation by simply shifting the likelihood by the prior, and straightforward and meaningful interpolation as we will give an example in Section \ref{sec:mnist}.

\subsection{Experiments}

As a proof of concept, we report results of toy experiments. We first consider synthetic two-classes and two-dimensional datasets. Indeed, in this case, the base space is two-dimensional and can be fully visualized. We will then discuss a gaussian three-classes example. In a three-classes case, the discriminant subspace is two-dimensional and can therefore be fully visualized. Finally, we report discriminant and interpolation results on a ten-classes case with the hand-written digits dataset MNIST.

\subsubsection{Concerning the log-likelihood-ratio cost and the so-called expected calibration error} Before going any further, we need to introduce the log-likelihood-ratio cost $\Cllr$. Even though this metric is popular for the evaluation of forensic identification systems, it is still little-known in the machine learning community. It has been introduced for the evaluation of systems that produce LLRs in the context of speaker verification \citep{brummer2006}. The $\Cllr$ measures the goodness of a set of LLRs in terms of both discrimination and calibration. Among the several possible intepretations of the $\Cllr$, one that will be familiar to the people working on calibration, is as an expected proper scoring rule (PSR) \citep{tilmann2007,brocker2009reliability,silva2023classifier}. To be more precise, the $\Cllr$ is the empirical expectation of the cross-entropy (with a log score\footnote{For the choice of the log score for assessing LLRs, the reader can refers to \citep{brummer2006}.})---also known as the binary cross-entropy loss in machine learning---where each probability is computed through the log-likelihood-ratio and a non-informative prior log-odds of 0. Like every expected PSR, it can be decomposed into two terms: a calibration term and a discrimination term. This decomposition is insured by the pool adjacent violator (PAV) algorithm which is known to minimize the expected PSR under a monotonicity constraint \citep{pav_brummer}. The $\Cllr$ can therefore be written as $\Cllr = \Cllrcal + \Cllrmin$, where $\Cllrmin$ is obtained with the LLRs that have been calibrated through the PAV algorithm and informs on the discrimination quality of the LLRs. $\Cllrcal = \Cllr - \Cllrmin$ informs on the calibration quality of the LLRs. However, the reader has to keep in mind that calibration only and discrimination only do not inform on the goodness of the LLRs. As an example, a set of LLRs can be perfectly calibrated but non-discriminant at all and therefore non-informative. The goodness of the LLRs should be assessed through the $\Cllr$ which incorporate both aspects and informs on the quality of the information provided by the LLRs \citep{brummer2006}, like expected PSR should be used for evaluating the goodness of posterior probabilities \citep{ferrer2025evaluating} regardless of their discrimination and calibratiton quality separately.

In machine learning, a popular metric for evaluating the calibration of probabilistic predictions is the Expected Calibration Error (ECE) \citep{naeini2015obtaining,guo2017}. However, it is based on a suboptimal and unstable binning strategy, whereas the optimal one is given by the PAV algorithm \citep{pav_brummer,timostable}. One possible reason for the popularity of the ECE is that it can be traced back to the decomposition of the expected Brier score in \cite{degroot1983}. However, in \cite{degroot1983}, the forecaster is allowed to choose a probability within a finite set of values: $\left\{ 0, 0.1, 0.2,\dots 1 \right\}$. In this case, the decomposition of the expected Brier score is natural and corresponds to the ECE's binning. In machine learning, predictions are not limited to this finite set of allowed values. ECE's binning strategy becomes therefore arbitrary and suboptimal. Consequently, we do not rely on ECE and instead report LLRs' quality in terms of the well-established $\Cllr$ decomposition\footnote{Basically, the $\Cllrcal$ can be seen as an ECE, but with an optimal binning, with the log score instead of the Brier score, and with a non-informative prior.}. Now, let's come back to our experiments.

\subsubsection{Two-classes and two-dimensional examples}
We provide a few two-classes and two-dimensional experiments to illustrate the CDA. We compare our approach---in terms of both discrimination and calibration---with the LDA and the QDA on three artificial datasets. The first dataset, called \emph{Gaussians}, consists of two multivariate Gaussians with different means and covariance matrices. The second dataset, called \emph{Moons}, consists of two interleaving noisy half-circles. The third one, called \emph{Circles}, consists of a large noisy circle containing a smaller one\footnote{These datasets are generated with scikit-learn \citep{scikit-learn}.}. For each set, 12000 samples are generated, 10000 are used for training and 2000 are used for testing the discriminant analysis. The results are assessed in terms of $\Cllr$, $\Cllrmin$, and scatter-plot visualizations.

\begin{figure}
  \centering
  \begin{subfigure}[b]{0.31\linewidth}
    \centering
    \includegraphics[width=\linewidth]{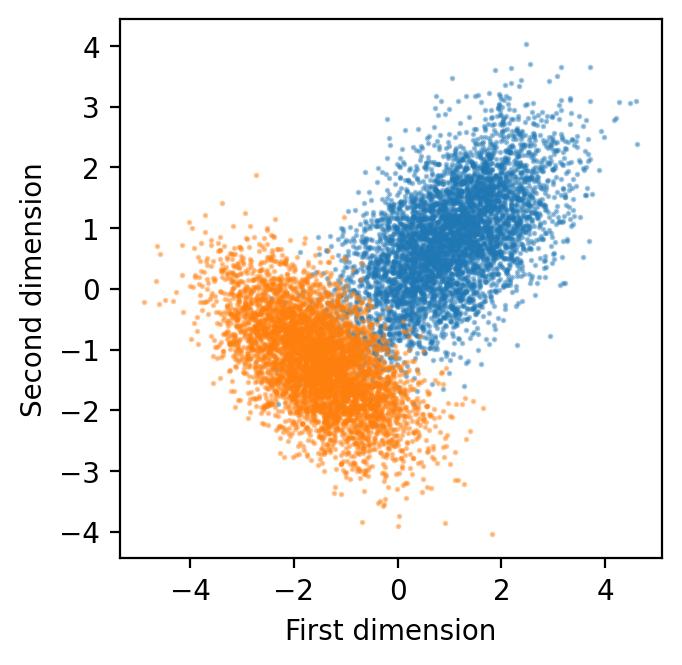}
    \caption{Gaussians dataset.}
    \label{fig:traingauss}
  \end{subfigure}
  \begin{subfigure}[b]{0.31\linewidth}
    \centering
    \includegraphics[width=\linewidth]{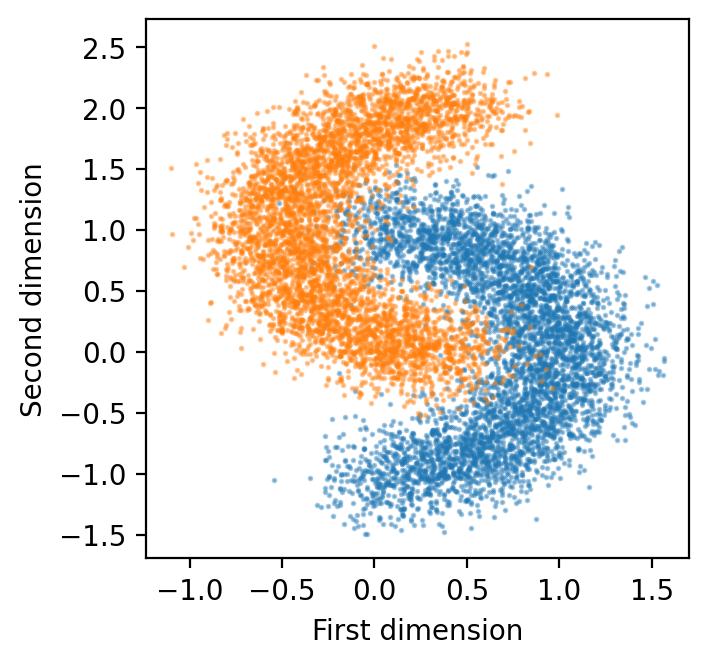}
    \caption{Moons dataset.}
    \label{fig:trainmoons}
  \end{subfigure}
  \begin{subfigure}[b]{0.31\linewidth}
    \centering
    \includegraphics[width=\linewidth]{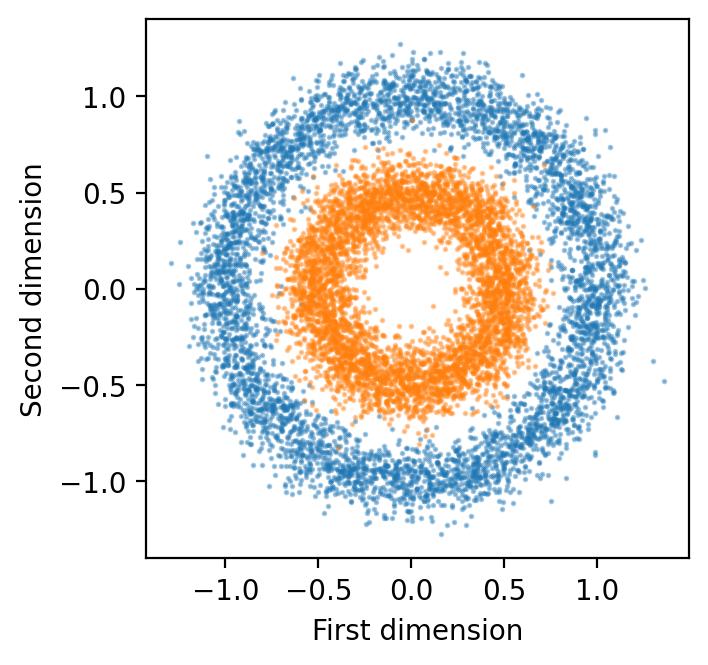}
    \caption{Circles dataset.}
    \label{fig:traincircles}
  \end{subfigure}
  \caption{Training sets for the two-classes and two-dimensional examples. The color indicates to which class a sample belongs: blue for $C_1$ and orange for $C_2$.}
\end{figure}
\begin{table}
  \centering
  \caption{$\Cllr$ measures of the discriminant analysis on the two-classes examples.}
  \begin{tabular}{|c|cc|cc|cc|}
    \cline{1-7}
    \multirow{2}{*}{datasets} & \multicolumn{2}{c|}{\makecell{LDA}} & \multicolumn{2}{c|}{\makecell{QDA}} & \multicolumn{2}{c|}{\makecell{CDA}}\\
    \cline{2-7}
    & $\Cllrmin$ {\scriptsize[bit]} & $\Cllr$ {\scriptsize[bit]} & $\Cllrmin$ & $\Cllr$ & $\Cllrmin$ & $\Cllr$ \\
    \hline
    \hline
    Gaussians & 0.125 & 0.198 & 0.115 & 0.126 & 0.117 & 0.155\\
    \hline
    Moons & 0.387 & 0.432 & 0.387 & 0.432 & 0.105 & 0.118\\
    \hline
    Circles & 0.839 & 1.000 & 0.023 & 0.491 & 0.040 & 0.054\\
    \hline
  \end{tabular}
  \label{tab:cllr}
\end{table}

\begin{figure}[h]
  \centering
  \begin{subfigure}[b]{0.244\linewidth}
    \centering
    \includegraphics[width=\linewidth]{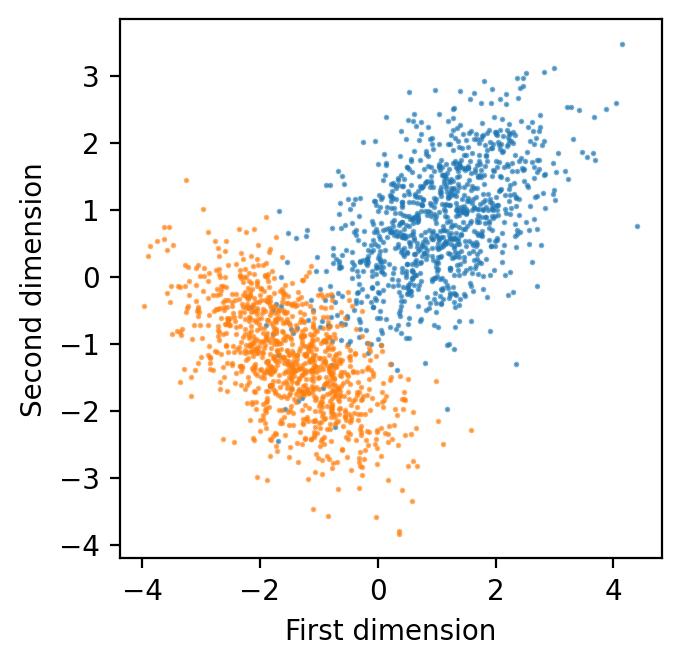}
    \caption{Ground truth.}
  \end{subfigure}
  \begin{subfigure}[b]{0.244\linewidth}
    \centering
    \includegraphics[width=\linewidth]{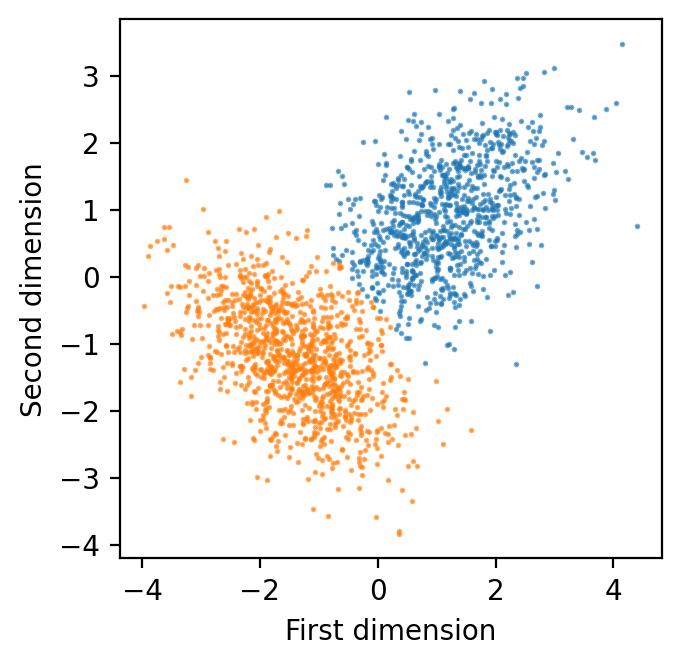}
    \caption{With LDA.}
  \end{subfigure}
  \begin{subfigure}[b]{0.244\linewidth}
    \centering
    \includegraphics[width=\linewidth]{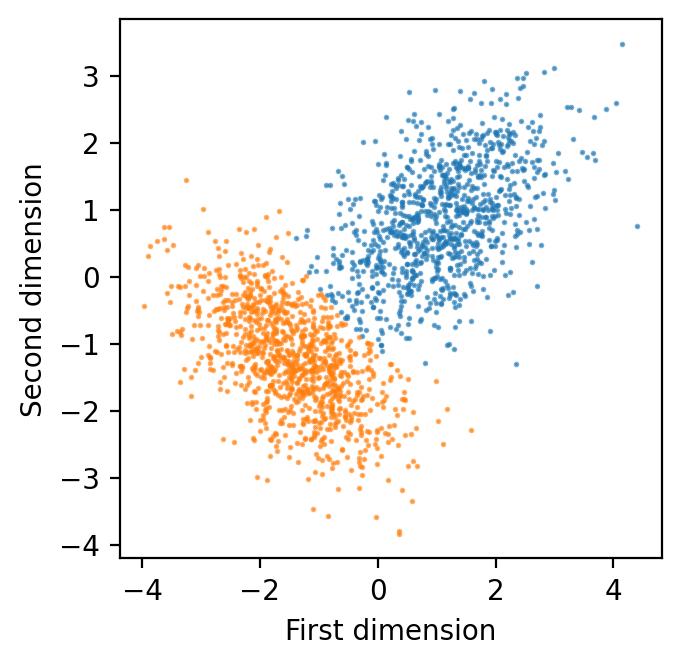}
    \caption{With QDA.}
  \end{subfigure}
  \begin{subfigure}[b]{0.244\linewidth}
    \centering
    \includegraphics[width=\linewidth]{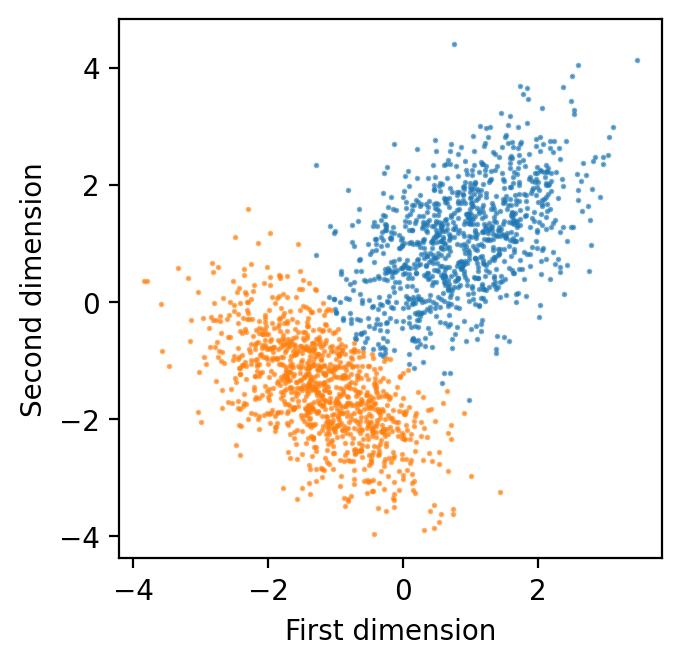}
    \caption{With CDA.}
  \end{subfigure}
  \caption{Maximum likelihood classification on the Gaussians dataset. In (a), the colors indicate the true label. For the other figures, the colors indicate the predicted class.}
  \label{fig:testgaussians}
\end{figure}

Figure \ref{fig:traingauss} shows the training set for the Gaussian example. Figure \ref{fig:testgaussians} shows the results of the maximum likelihood classification using LDA, QDA, and the proposed CDA. $\Cllr$ measures are reported in Table \ref{tab:cllr}. Both LDA and QDA are based on the Gaussian assumption. However, the LDA assumes that both classes share the same covariance which is not the case. The LDA has therefore a discrimination and a calibration that are not as good as the QDA. The $\Cllrmin$ is 0.125 bit for the LDA while it is 0.115 bit for the QDA and the calibration cost $\Cllrcal = \Cllr - \Cllrmin$ is 0.073 bit for the LDA and 0.011 for the QDA. Here, the QDA is also better than the CDA which has a $\Cllrmin$ of 0.117 bit and a $\Cllrcal$ of 0.038 bit. The goodness of the QDA is here not surprising since the assumed data distribution and the actual distribution match.
\begin{figure}[h]
  \centering
  \begin{subfigure}[b]{0.244\linewidth}
    \centering
    \includegraphics[width=\linewidth]{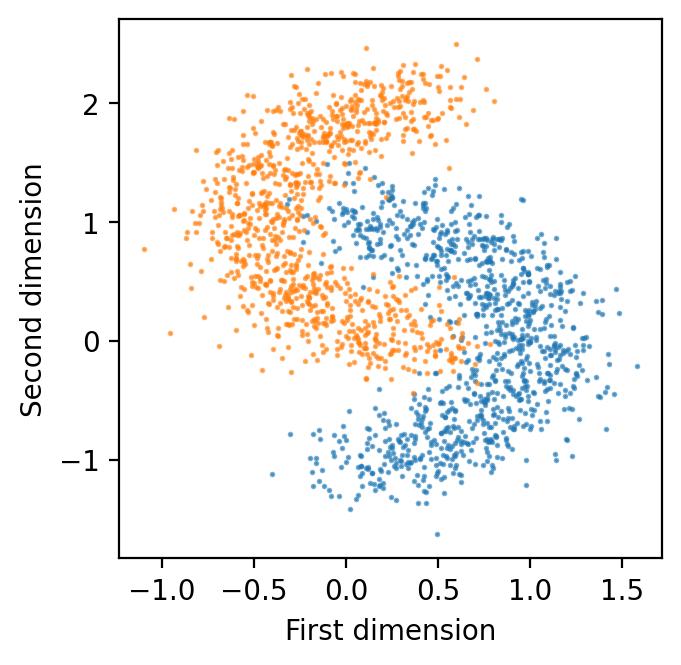}
    \caption{Ground truth.}
  \end{subfigure}
  \begin{subfigure}[b]{0.244\linewidth}
    \centering
    \includegraphics[width=\linewidth]{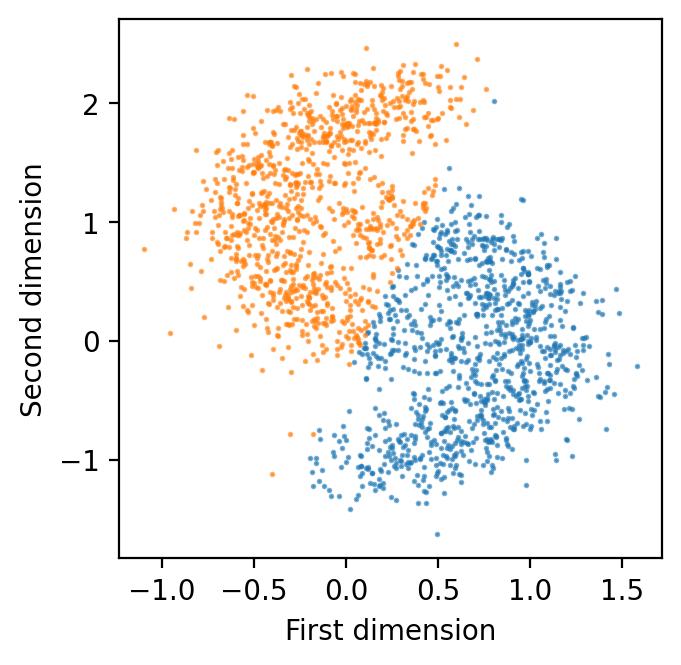}
    \caption{With LDA.}
  \end{subfigure}
  \begin{subfigure}[b]{0.244\linewidth}
    \centering
    \includegraphics[width=\linewidth]{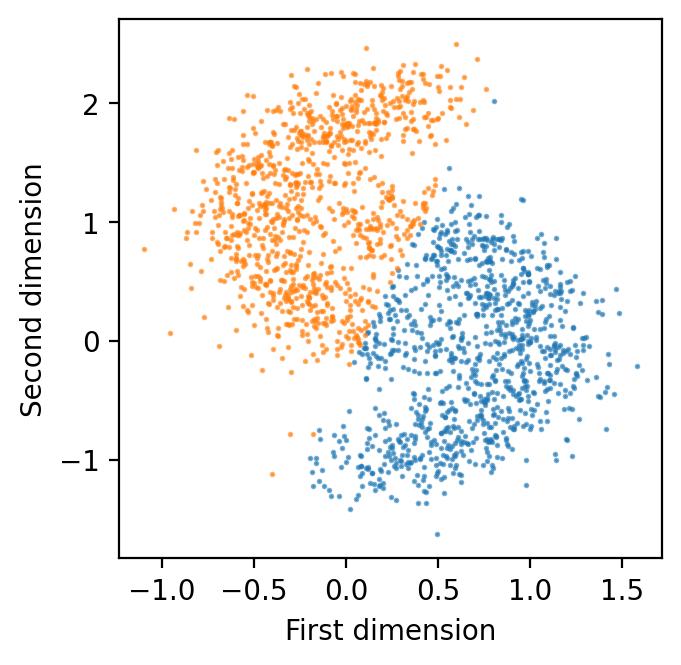}
    \caption{With QDA.}
  \end{subfigure}
  \begin{subfigure}[b]{0.244\linewidth}
    \centering
    \includegraphics[width=\linewidth]{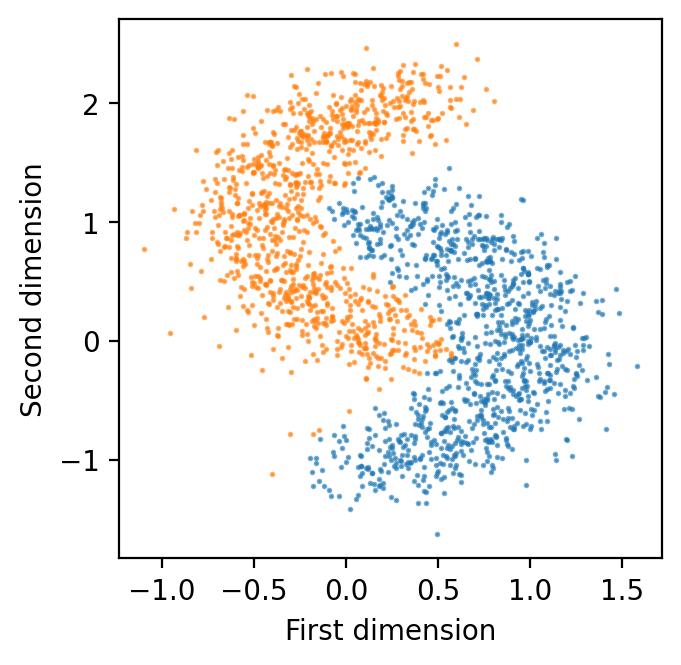}
    \caption{With CDA.}
  \end{subfigure}
  \caption{Maximum likelihood classification on the Moons dataset. In (a), the colors indicate the true label. For the other figures, the colors indicate the predicted class.} 
  \label{fig:testmoons}
\end{figure}
\begin{figure}[h]
  \centering
  \begin{subfigure}[b]{0.244\linewidth}
    \includegraphics[width=\linewidth]{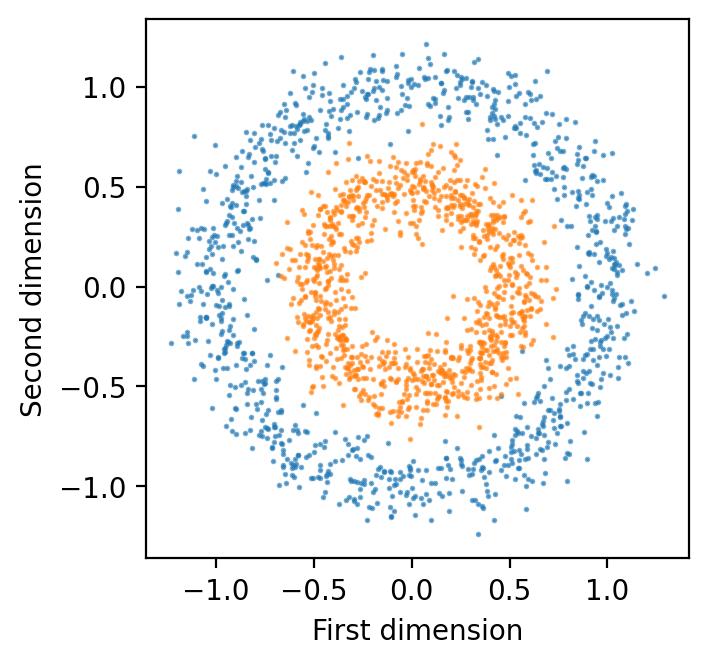}
    \caption{Ground truth.}
  \end{subfigure}
  \begin{subfigure}[b]{0.244\linewidth}
    \includegraphics[width=\linewidth]{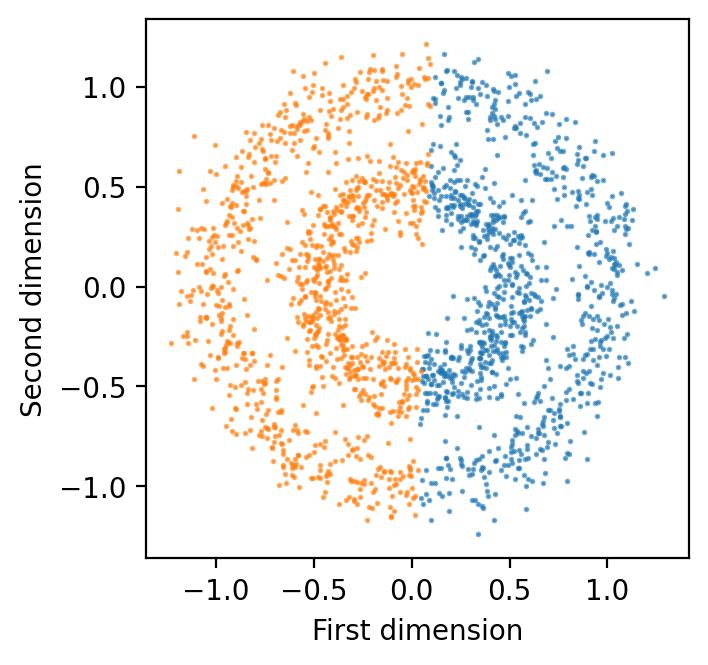}
    \caption{With LDA.}
  \end{subfigure}
  \begin{subfigure}[b]{0.244\linewidth}
    \includegraphics[width=\linewidth]{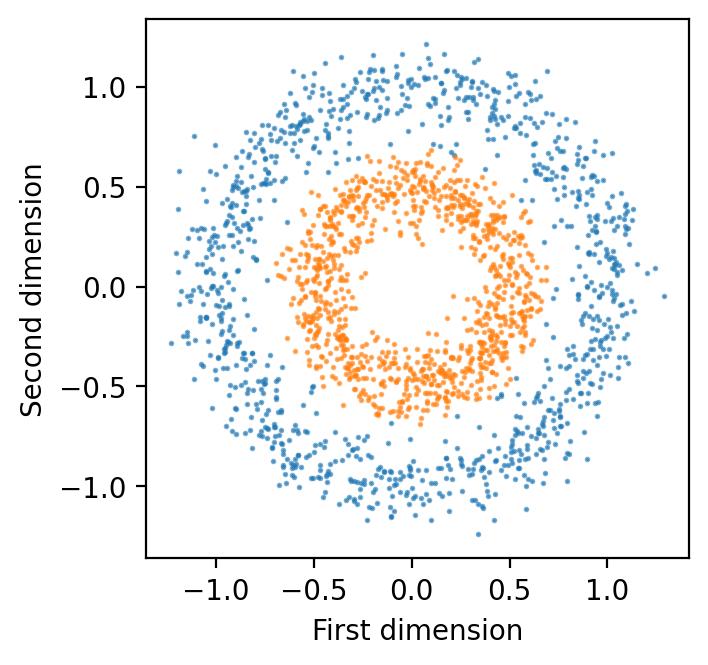}
    \caption{With QDA.}
  \end{subfigure}
  \begin{subfigure}[b]{0.244\linewidth}
    \includegraphics[width=\linewidth]{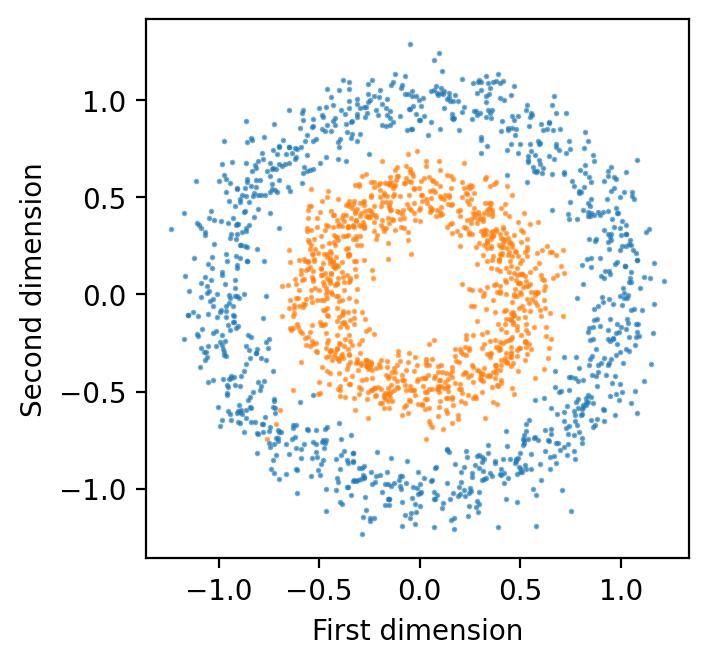}
    \caption{With CDA.}
    \label{fig:testcirclesCDA}
  \end{subfigure}
  \caption{Maximum likelihood classification on the Circles dataset. In (a), the colors indicate the true label. For the other figures, the colors indicate the predicted class.}
  \label{fig:testcircles}
\end{figure}

Figure \ref{fig:trainmoons} shows the training set for the Moons example. Figure \ref{fig:testmoons} shows the results of the maximum likelihood classification using LDA, QDA, and the CDA. $\Cllr$ measures are reported in Table \ref{tab:cllr}. Both LDA and QDA hardly separate the two classes while the CDA does better in terms of both discrimination and calibration with $\Cllr = 0.105$ bit and $\Cllrcal = \Cllr - \Cllrmin = 0.013$ bit.

\begin{figure}
  \centering
  \includegraphics[width=0.4\linewidth]{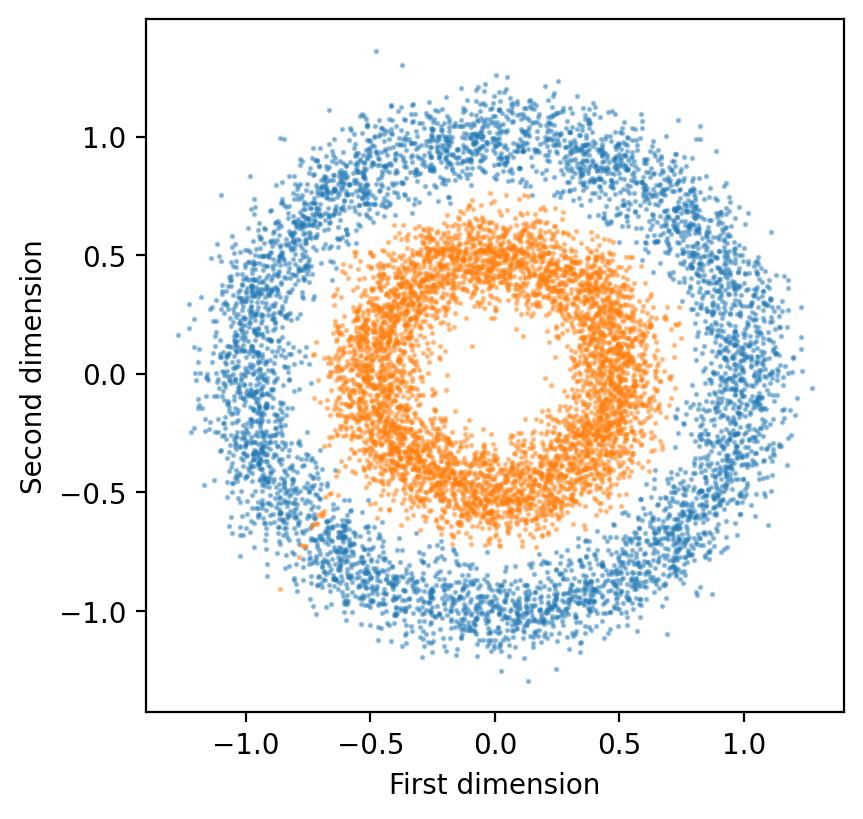}
  \caption{Compositional discriminant analysis on Circles training set (ground truth is given in Figure~\ref{fig:traincircles}). Since it is restricted to invertible and differentiable transformations, this discriminant analysis will never ``perfectly'' separate the two classes as the tiny slice of miss-classification illustrates.}
  \label{fig:traincirclesnfda}
\end{figure}

Figure \ref{fig:traincircles} shows the training set for the Circles example.  Figure \ref{fig:testcircles} shows the results of the maximum likelihood classification. $\Cllr$ measures are reported in Table \ref{tab:cllr}. Being linear, the LDA cannot separate the two classes. The QDA has the best discrimination with a $\Cllrmin$ of $0.023$, while the CDA has a slightly higher $\Cllrmin$ of $0.040$. We can indeed see in Figure \ref{fig:testcirclesCDA} a tiny slice of blue samples that are miss-classified as orange on the left-bottom part of the larger circle. This is more discernible on the training set since they are more points (see Figure \ref{fig:traincirclesnfda}). This is because the family of mappings is restricted to a family of diffeomorphisms where none allows a ``perfect'' transformation of these interleaving circles into two distinct Gaussians. However, even if QDA has the best discrimination ability---thanks to the quadratic nature of the circle-shape boundary---it is still based on Gaussian assumptions while the data are definitely not normally distributed. This results in a calibration that is not as good as the calibration of the CDA. The QDA has indeed a $\Cllrcal$ of $0.468$ while the CDA has a $\Cllrcal$ of $0.037$ on the testing set. In this example, the implicit assumption discussed in the remark of Section \ref{sec:diffeo} is not fulfilled, the invertibility and differentiability constraints of the CDA limit the discrimination ability. However, having a low $\Cllr$, the CDA still produces better LLRs than the QDA. This is a typical example where a good discrimination does not necessarily implies a reliable extraction of the information: even if the QDA separates well the classes, the modeling of the data is bad, resulting in bad calibration. On the contrary, having a lower discrimination ability does not implies to have a worst modeling quality.\\

\begin{figure}[h]
  \centering
  \begin{subfigure}[b]{0.32\linewidth}
    \centering
    \includegraphics[width=0.9\linewidth]{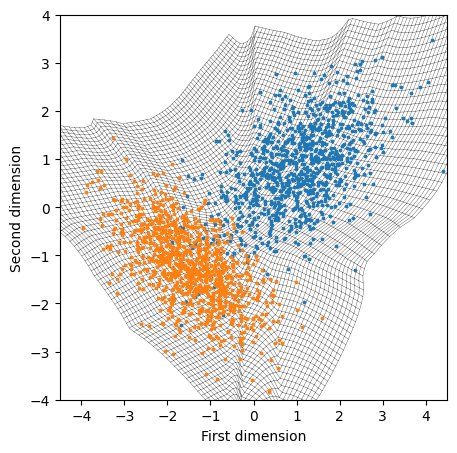}\\
    \includegraphics[width=0.9\linewidth]{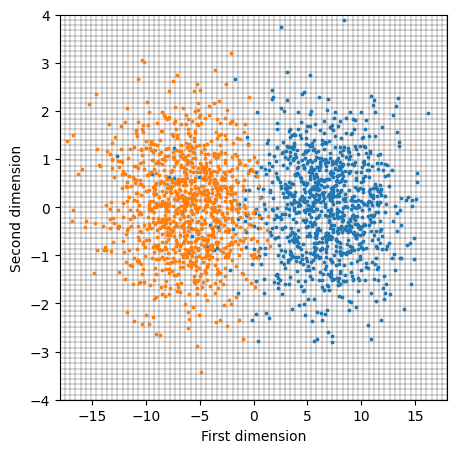}
    \caption{The Gaussians examples.}
  \end{subfigure}
  \begin{subfigure}[b]{0.32\linewidth}
    \centering
    \includegraphics[width=0.9\linewidth]{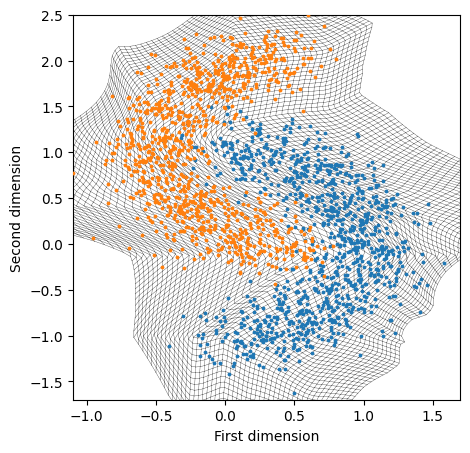}\\
    \includegraphics[width=0.9\linewidth]{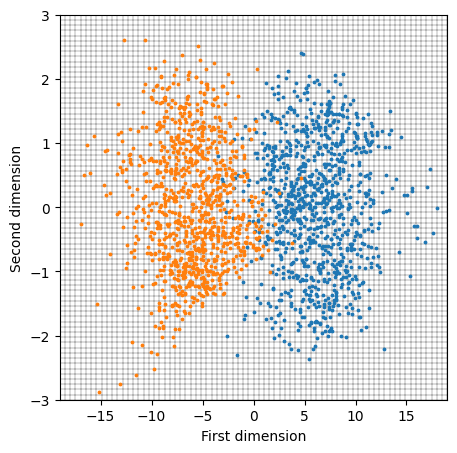}
    \caption{The Moons example.}
  \end{subfigure}
  \begin{subfigure}[b]{0.32\linewidth}
    \centering
    \includegraphics[width=0.93\linewidth]{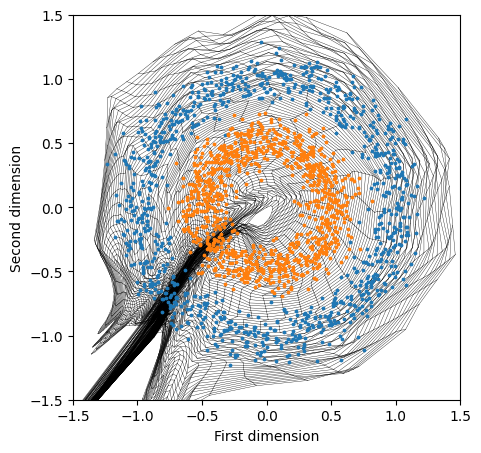}\\
    \includegraphics[width=0.9\linewidth]{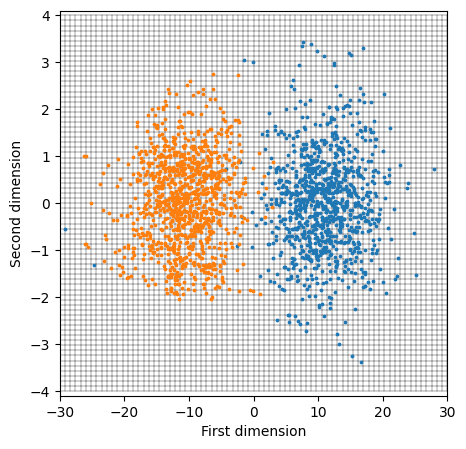}
    \caption{The Circles example.}
    \label{fig:slicecircle}
  \end{subfigure}
  \caption{Grid vizualisation of the learned diffeomorphisms on the two-classes and two-dimensional examples. For each example, the samples from the testing sets are shown over a grid. The regular grids in the base space are on the bottom, and the transformed grids in the feature space are on the top.}
  \label{fig:visdiffeo}
\end{figure}

\noindent
{\bf Grid visualisation of the learned transformation:} Since the above examples are two-dimensional, a learned diffeomorphism can be visualized as a transformation of a two-dimensional grid as shown in Figure \ref{fig:visdiffeo}. The bottom part of Figure \ref{fig:visdiffeo} shows the testing data in the learned base space for the three examples. As expected, for each class, the data looks normally distributed and symmetric around zero. For visualizing the learned diffeomorphism, a set of points is generated homogeneously and regularly to form a grid over the base space. The set of points are transformed through the learned diffeomorphism and the resulting deformed grid is visualized in the feature space (top of Figure \ref{fig:visdiffeo}). In the feature space, the transformed regular grid approximates the manifold on which the data lives. The colors represent the true label of the samples. One can see, for the \emph{Circles} example, the slice of miss-classification of the blue circle at the bottom left of the circle. The orange circle's samples are ``going through'' the blue circle at the expense of the few blue samples that will be miss-classified.

In the above examples, we discussed the two-classes case only\footnote{For a two-classes example on real speech data, see \cite{Noe22Icassp} and \cite{noe2022hiding}.}. In this case, the discriminant subspace is one-dimensional and is the log-likelihood-ratio line with the properties that have been presented in Section \ref{sec:llr}. In the following, we present examples with more classes to fully appreciate our results presented in the multiclass setting in Section \ref{sec:ILRL}.

\subsubsection{A Gaussian three-classes and four-dimensional example}
\label{sec:3gauss}

Here we consider three-classes in a four-dimensional feature space. For each class, the data is normally distributed with different mean vectors and different covariance matrices. The discriminant subspace, \ie~the space of the isometric-log-ratio transformed likelihood functions (ILRL), is two-dimensional and the residual subspace too. For conciseness, we report only the visualization of the test data in the discriminant and the residual space. More details for this example can be found in Appendix \ref{app:3gaussians}.

Figure \ref{fig:test_base_gaussians_nonshared} shows the base space on which the testing set is mapped using the learned transformation. Figure \ref{fig:test_ILRL_gaussians_nonshared} shows the discriminant subspace of the CDA, \ie~the ILRL space. The first dimension discriminates the two first classes (blue and orange) while the second dimension discriminates the third class (green) from the two others without discriminating the first two. This is in accordance with the used Aitchison basis corresponding to the bifurcation tree of Figure \ref{fig:bifurctree}. The distributions of the data in the discriminant subspace tend to respect the distributions of calibrated ILRL of Theorem \ref{prop:distribILRL} such that the discriminant component can be interpreted as a calibrated likelihood function. The residual components (in Figure \ref{fig:test_res_gaussians_nonshared}) are not discriminant and are normally distributed with zero mean and identity covariance matrix in accordance with the design of the CDA as presented in Section \ref{sec:classconddensitiescompo}. The learned parameter $\bm{\Sigma}$ 
% \begin{equation*}
%   \bm{\Sigma} = \begin{bmatrix} 35.50 & -0.87\\ -0.87 & 8.71\end{bmatrix},
% \end{equation*}
can be interpreted in terms of the following divergences, or separability between the classes, using Equation \ref{eq:fromcov2div}: $d_{1,2} = 35.5$, $d_{1,3} = 14.7$, and $d_{2,3} = 16.2$.

Figures \ref{fig:test_ldadisrcim_gaussians_nonshared} and \ref{fig:test_ldares_gaussians_nonshared} show the projection of the testing set using the LDA. The first two components shown in \ref{fig:test_ldadisrcim_gaussians_nonshared} are discriminant but are hardly interpretable by other means than the discriminative power given by the eigenvalues. The other dimensions seem less discriminant, but still contain class-related information, and are not identically distributed contrary to the CDA's residual space (Figure \ref{fig:test_res_gaussians_nonshared})\footnote{QDA is not designed to have an information-preserving mapping of the data into a same-dimensional space. This is why there are no results for the QDA in Figure \ref{fig:test_base_gaussians_nonshared}. More results on this example, including the QDA, can be found in Appendix \ref{app:3gaussians}.}.
\begin{figure}[h]
  \centering
  \begin{subfigure}[b]{0.49\linewidth}
    \centering
    \includegraphics[scale=0.15]{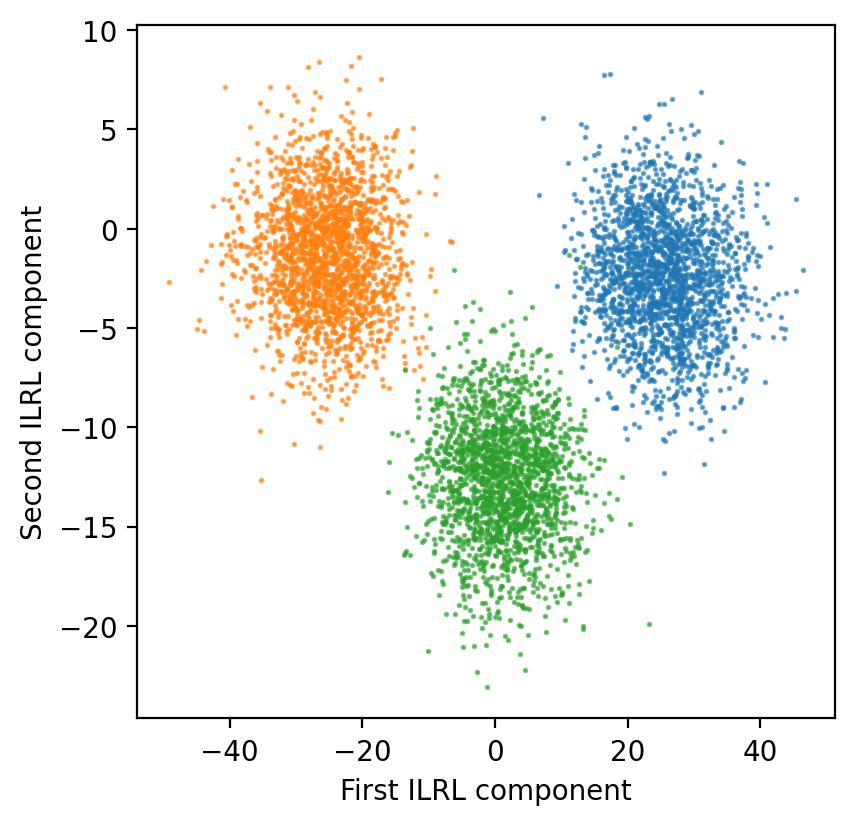}
    \caption{The CDA's ILRL components.}
    \label{fig:test_ILRL_gaussians_nonshared}
  \end{subfigure}
  \begin{subfigure}[b]{0.49\linewidth}
    \centering
    \includegraphics[scale=0.15]{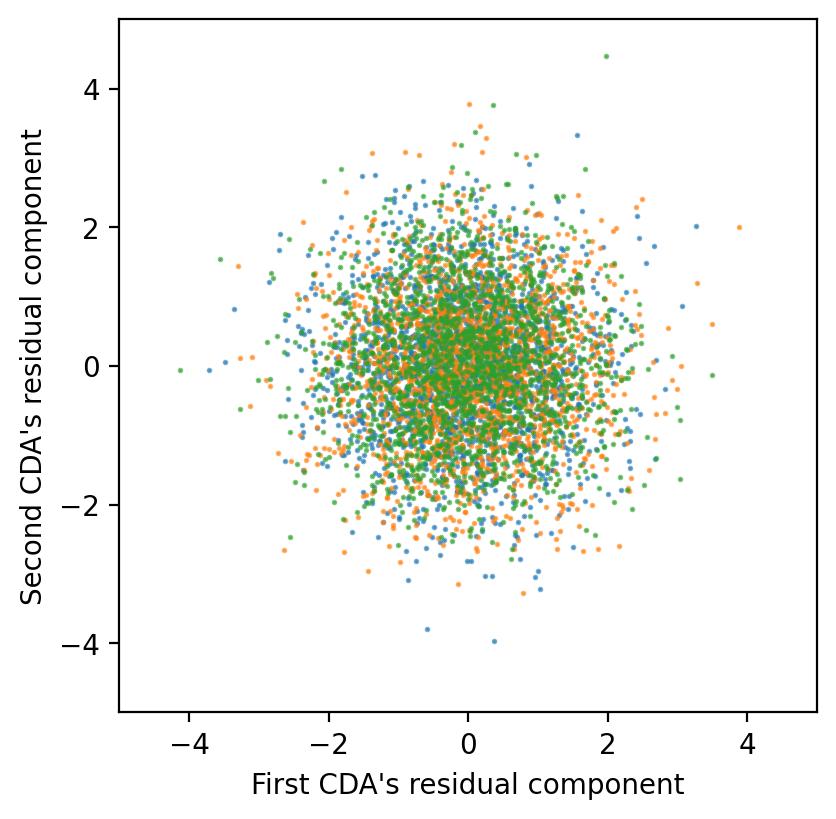}
    \caption{The CDA's residual components.}
    \label{fig:test_res_gaussians_nonshared}
  \end{subfigure}\\
  \begin{subfigure}[b]{0.49\linewidth}
    \centering
    \includegraphics[scale=0.15]{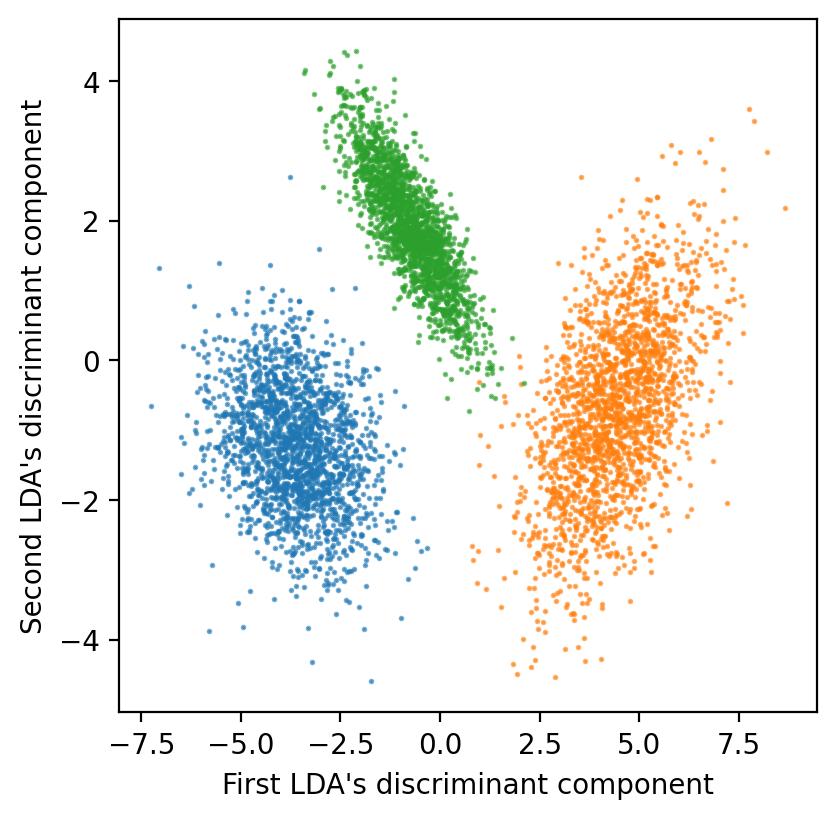}
    \caption{The LDA's discriminant components.}
    \label{fig:test_ldadisrcim_gaussians_nonshared}
  \end{subfigure}
  \begin{subfigure}[b]{0.49\linewidth}
    \centering
    \includegraphics[scale=0.15]{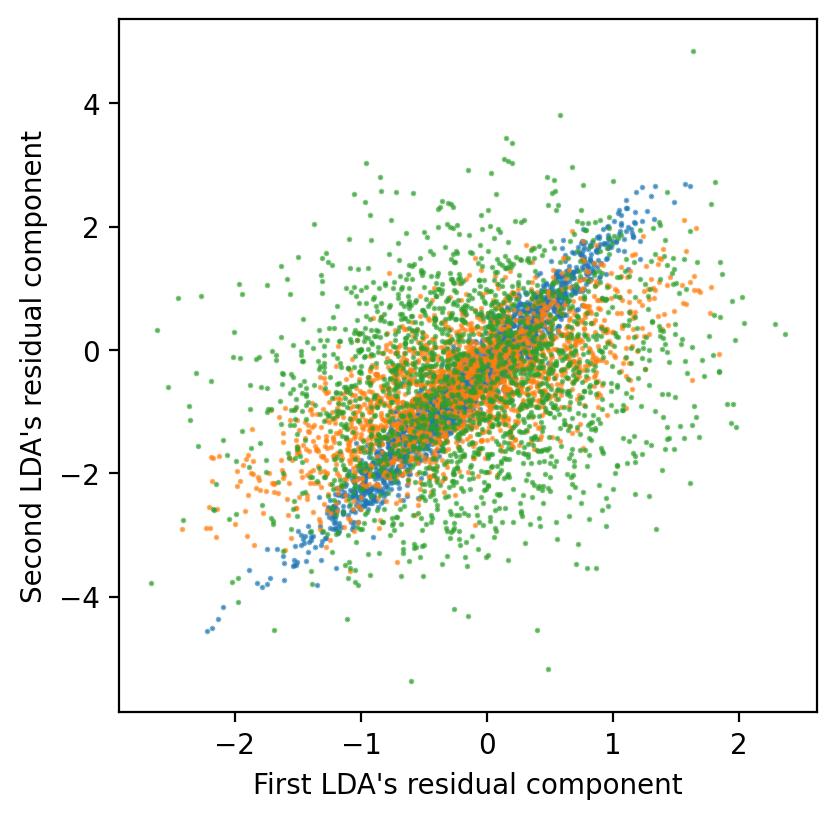}
    \caption{The LDA's residual components.}
    \label{fig:test_ldares_gaussians_nonshared}
  \end{subfigure}
  \caption{Testing set in the LDA and CDA base spaces for the three-classes non-shared covariance Gaussian example.}
  \label{fig:test_base_gaussians_nonshared}
\end{figure}

\subsubsection{Hand-written digits recognition and interpolation}
\label{sec:mnist}

\begin{figure}
  \centering
  \includegraphics[scale=0.12]{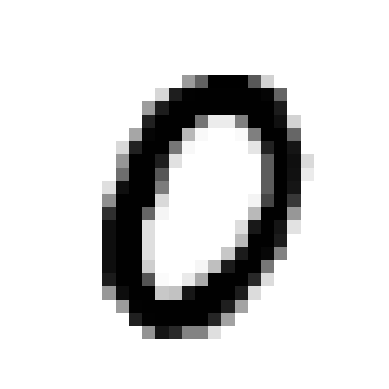}
  \includegraphics[scale=0.12]{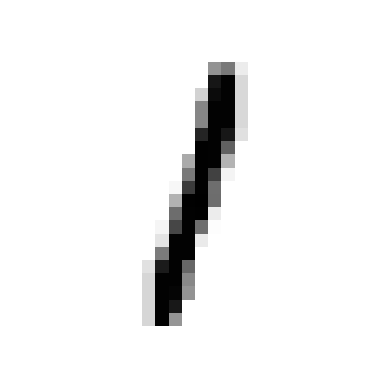}
  \includegraphics[scale=0.12]{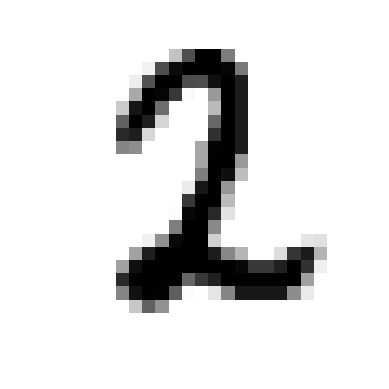}
  \includegraphics[scale=0.12]{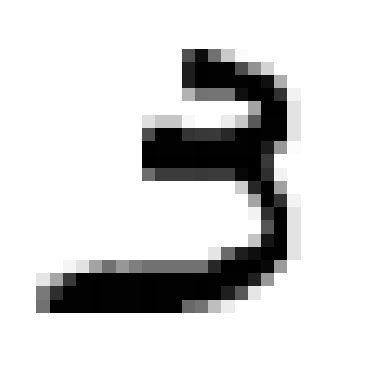}
  \includegraphics[scale=0.12]{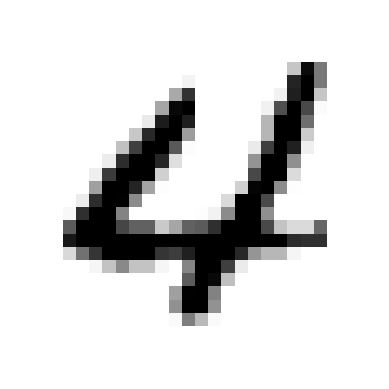}
  \includegraphics[scale=0.12]{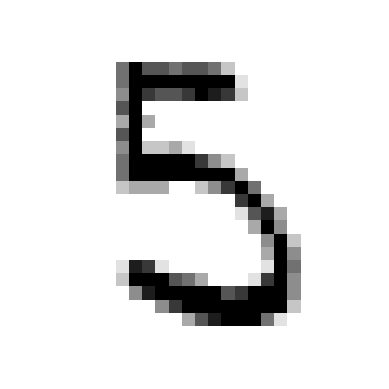}
  \includegraphics[scale=0.12]{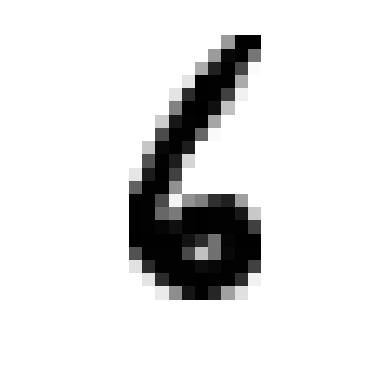}
  \includegraphics[scale=0.12]{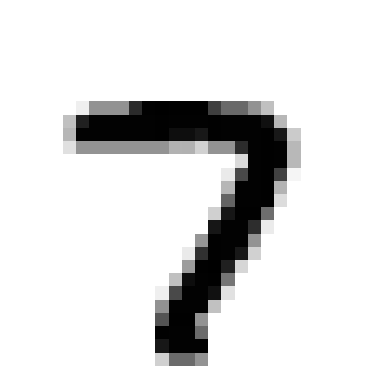}
  \includegraphics[scale=0.12]{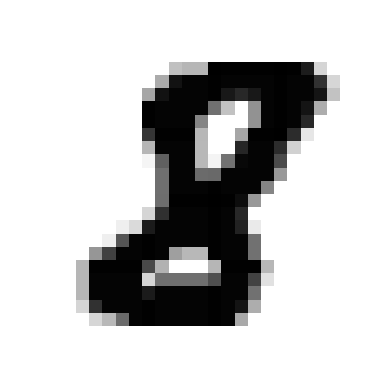}
  \includegraphics[scale=0.12]{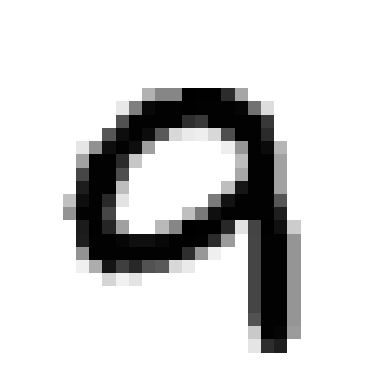}
  \caption{Samples from the MNIST database.}
  \label{fig:ex_digits}
\end{figure}

The MNIST database consists of grayscale images of size $28 \times 28$. Each image is a handwritten digit between $0$ and $9$ \citep{lecun1998mnist}.
The training set is made of $60000$ samples and the testing set is made of $10000$ samples. Figure \ref{fig:ex_digits} shows one randomly selected example for each class.

In the following experiment, each image is flattened and normalized into a $784$-dimensional feature vector\footnote{State-of-the-art discriminative approaches for classification on MNIST are based on CNN. Even if coupling layers \cite{dinh2017} can be made of convolutions, we do not consider this way of processing the images.}. One can see from Figure \ref{fig:ex_digits} that the pixels on the edges of the images tend to all have the same low intensity. This leads to collinearities between some of the features such that methods that require the inversion of covariance matrices in the feature space (like the LDA and the QDA) can not be directly used. The dimensionality of the feature vector is therefore reduced to 40 using a principal component analysis (PCA).

In the above Gaussian example, where only three classes were considered, the $\Cllr$ measures can still be reported for each of the three pairs of classes (Appendix \ref{app:3gaussians}). In the current example, there are 10 classes corresponding to 45 pairs of classes. Thus, we instead report the empirical expected log scoring rule (or cross-entropy loss) with a uniform prior. It is a multiclass extension of the $\Cllr$ and is defined as:
  \begin{equation}
    \label{eq:Cmc}
    C_{mc} = - \frac{1}{N} \sum_{i=1}^N \frac{1}{\lvert \mathcal{X}_i \rvert} \sum_{\bm{x} \in \mathcal{X}_{i}} s_i(\bm{x}),
  \end{equation}
  where $\mathcal{X}_i = \left\{ x \in \mathcal{X} \mid c = C_i \right\}$ and $s_i(\bm{x})$ is the log-likelihood prediction, for input $\bm{x}$ and class $C_i$, given by the classifier: $s_i(\bm{x}) = \left(\ilr^{-1} \left( g^{-1}(\bm{x})_{1:D-1} \right)\right)_i = \left(\ilr^{-1} \left( \bm{z}_{1:D-1} \right)\right)_i  $. Note that in the two classes case, the PAV algorithm calibrates a set of LLRs and minimizes the empirical expected scoring rule, without changing the discrimination quality. This allows the decomposition into a calibration term and a discrimination term. However, when more than two classes are involved, there is no available method to obtain perfectly calibrated probabilities or likelihoods as reference for the decomposition. Thus, $C_{mc}$ can not be decomposed into a calibration term and a discrimination term, but can still be used to summarise the amount of useful information given by the recognizer as done for instance in the context of language recognition \citep{rodriguez2013albayzin}.%Having a multiclass ECE lower than the entropy of the prior probability distribution informs that the system has extracted useful information.

  In our experiments, we report the empirical expected scoring rule $C_{mc}$. In addition, since we do not have a minimum expected scoring rule as a discrimination measure, we report the accuracy of the system for a maximum likelihood decision rule%\footnote{The accuracy is given by the ratio of well recognized examples over the total number of examples.}
  \footnote{Be aware that the cross-entropy-based measures and the accuracy differ by nature. The accuracy measures the goodness of hard decisions while the cross-entropy measures the goodness of probabilities or likelihoods regardless of the operating point or decision boundaries. In this way, the accuracy can not substitute a minimum expected scoring rule.}.

  Table \ref{tab:mnist_results} gives the  $C_{mc}$ in nat\footnote{A nat is a unit of information when the natural logarithm is used while a bit is a unit of information with the base two.} and the accuracy on the testing set for the LDA, the QDA, and CDA. All the systems result in a cross-entropy loss lower than the entropy of the uniform prior distribution: $C_{mc}< \log 10 \approx 2.30$, meaning that all the systems extract useful information from the images. Interestingly, QDA has the best accuracy but the worst cross-entropy loss. This confirms that having a good accuracy does not mean that a system models well the data and is good for making rational decisions in general \ie~expected cost minimizing decisions. The CDA results in the lowest cross-entropy which shows that it extracts the most useful information from the images.

\begin{figure}
  \centering
  \begin{subfigure}[b]{0.48\linewidth}
    \centering
    \includegraphics[scale=0.5]{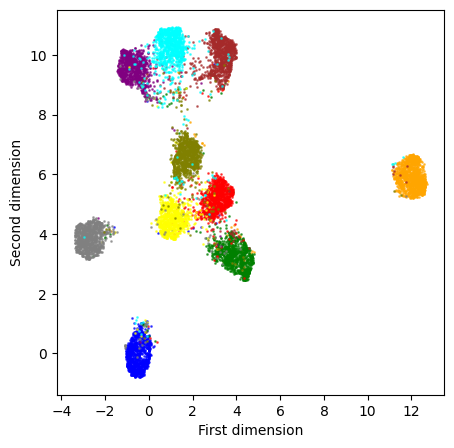}
    \caption{UMAP visualization of the estimated ILRLs.}
    \label{fig:ilrlmnist}
  \end{subfigure}
  \begin{subfigure}[b]{0.48\linewidth}
    \centering
    \includegraphics[scale=0.5]{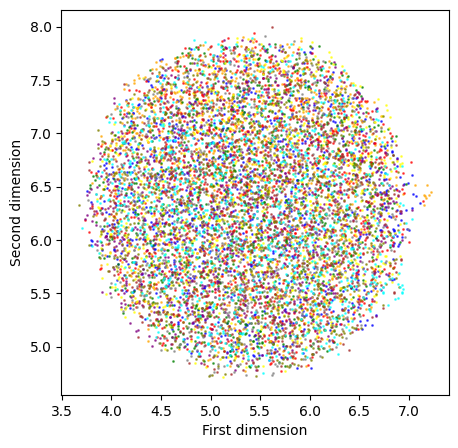}
    \caption{UMAP visualization of the residual.}
    \label{fig:resmnist}
  \end{subfigure}
  \caption{UMAP visualization of the MNIST testing data in the CDA's base space. The color indicates to which class a sample belongs: blue for 0, orange for 1, green for 2, red for 3, purple for 4, yellow for 5, gray for 6, brown for 7, olive for 8, and cyan for 9.}
  \label{fig:ilrlresmnist}
\end{figure}

  Being respectively 9-dimensional and 31-dimensional, the discriminant subspace and the residual subspace can not be visualized in 2-dimensional plots. We therefore report dimension-reduction based visualization using the uniform manifold approximation and projection method (UMAP) \citep{mcinnes2018umap}. Figure \ref{fig:ilrlmnist} shows an UMAP visualization of the nine first dimensions \ie~of the estimated ILRL. One can see the ten clusters. Note that the components given by the UMAP can not be interpreted, the figures are just for illustration and cluster visualization purpore. Figure \ref{fig:resmnist} shows an UMAP visualization of the residual where no clusters appear. This suggests, as expected, that the digit-related information is concentrated in the ILRL components. Table \ref{tab:divmnist} shows the estimated Kullback-Leibler divergences between the digit's class-conditional distributions in the base space. These divergences are computed from the estimated $\bm{\Sigma}$ using Equation \ref{eq:fromcov2div}. This informs us about the separability between the classes.% Even if the observed distance between the clusters in Figure \ref{fig:ilrlmnist} can not be read as they are, they share some overall tendencies with the estimated divergences.
\begin{table}
  \centering
  \caption{Cross-entropy ($C_{mc}$) and accuracy measures on the testing set for the MNIST's digit recognition task with LDA, QDA, and CDA.}
  \label{tab:mnist_results}
  \begin{tabular}{|c|c|c|}
    \hline
    system & $C_{mc}$ {\scriptsize[nat]} & accuracy {\scriptsize[\%]} \\
    \hline
    \hline
    LDA & 5.44{\tiny $10^{-1}$} & 87.67 \\
    \hline
    QDA & 8.56{\tiny $10^{-1}$} & 96.24 \\
    \hline    
    CDA & 2.23{\tiny $\bm{10^{-1}}$} & 94.43 \\
    \hline
  \end{tabular}
\end{table}
\begin{table}[t]
  \centering
  \caption{Estimated Kullback-Leibler divergences between the digit's conditional distributions in the base space.}
  \begin{tabular}{|c||c|c|c|c|c|c|c|c|c|c|}
    \hline
    digit & 0 & 1 & 2 & 3 & 4 & 5 & 6 & 7 & 8 & 9 \\
    \hline
    \hline
    0 & 0 & - & - & - & - & - & - & - & - & - \\
    \hline
    1 & 38.3 & 0 & - & - & - & - & - & - & - & - \\
    \hline
    2 & 18.5  & 15.6 & 0 & - & - & - & - & - & - & - \\
    \hline
    3 & 16.2 & 16.0 & 8.8 & 0 & - & - & - & - & - & - \\
    \hline
    4 & 24.8 & 25.0 & 17.8 & 17.4 & 0 & - & - & - & - & -\\
    \hline
    5 & 12.4 & 19.2 & 14.1 & {7.2} & 13.3 & 0 & - & - & - & -\\
    \hline
    6 & 18.4 & 25.9 & 13.3 & 19.6 & 13.0 & 13.5 & 0 & - & - & -\\
    \hline
    7 & 21.2 & 21.6 & 17.5 & 15.0 & 14.2 & 16.9 & 27.1 & 0 & - & -\\
    \hline
    8 & 18.0 & 16.6 & 8.3 & {7.2} & 13.8 & {6.7} & 14.5 & 17.5 & 0 & -\\
    \hline
    9 & 21.2 & 21.3 & 17.9 & 12.8 & {5.1} & 11.9 & 19.0 & {7.3} & 10.9 & 0\\
    \hline
  \end{tabular}
  \label{tab:divmnist}
\end{table}

The CDA can be used without dimensionality reduction beforehand. This allows the learning of an information-preserving transformation between the space of images and a base space. In this case, CDA can be directly used for generating images. The interpretability of the base space allows intuitive manipulation or generation of images.\\
% The parameter $\bm{\Sigma}$ of the class-conditional densities in the base space defines the centroid for each digit (see Section \ref{sec:classconddensitiescompo}) and the Euclidean structure allows linear interpolation for generating ``digits in between''\footnote{We trained the CDA without pre-dimensionality reduction (\ie~on the 784-dimensional features vectors) resulting in a $C_{mc}$ of $3.81${\tiny$10^{-1}$} nats and an accuracy of $89.88$\%.}.

\begin{figure}
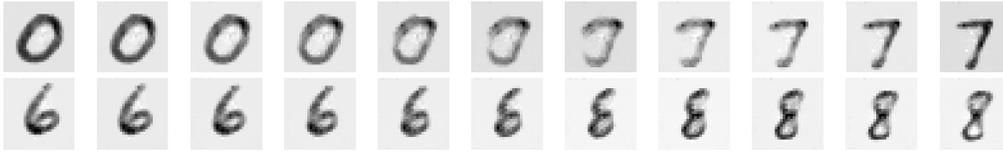

  \begin{center}
    \forloop{z}{0}{\value{z} < 11}{
      \includegraphics[scale=0.1]{discrim_ILRL/interpol_mnist/0-7/\thez.png}
    }\\
    
    \forloop{z}{0}{\value{z} < 11}{
      \includegraphics[scale=0.1]{discrim_ILRL/interpol_mnist/6-8/\thez.png}
    }
  \end{center}
  \caption{Examples of digit interpolation in the linear space of likelihood functions. From 0 to 7 (top), from 6 to 8 (bottom).}
  \label{fig:interpol}
\end{figure}

\noindent
{\bf Interpolation\footnote{We trained the CDA without pre-dimensionality reduction (\ie~on the 784-dimensional features vectors) resulting in a $C_{mc}$ of $3.81${\tiny$10^{-1}$} nat.}:} The Euclidean vector space structure of the base space allows easy interpolation between digits. This can be done with linear interpolation between the digits' centroid. The interpolation between the digit $i$ and the digit $j$ in the base space is given by:
\begin{equation}
  \bm{z}_{i,j}(\alpha) = \alpha \bm{m}_i + (1-\alpha) \bm{m}_j, 
\end{equation}
where $\alpha \in [0,1]$ and $\bm{m}_i$ and $\bm{m}_j$ are the learned digits' centroid as defined in Section \ref{sec:classconddensitiescompo}. The image can then be constructed by mapping $\bm{z}_{i,j}(\alpha)$ into the feature space using the learned diffeomorphism $g$: $\bm{x}_{i,j}(\alpha) = g(\bm{z}_{i,j}(\alpha))$. The feature vector is then unflattened to produce the image. Figure \ref{fig:interpol} shows two examples\footnote{More examples can be found in \cite{noe2023representing}.} of digit interpolation for $\alpha = \{0,0.1,0.2,0.3,0.4,0.5,0.6,0.7,0.8,0.9,1\}$. Because the discriminant space is a linear space of calibrated likelihood functions, the interpolated images have a neat interpretation: as an example, for the interpolation from 0 to 7, with $\alpha=0.5$, the interpolated image can be interpreted as equally likely to come from a 0 or a 7; alternativelely, with $\alpha = 0.2 = 1/5$, it is 5 times more likely to come from a 0 than a 7.

In this section, we saw how the discriminant space of a discriminant analysis can be constrained according to Theorem \ref{prop:distribILRL} to form the space of calibrated likelihood functions over the set of classes. We illustrated the relevance of this approach with simple experiments.

\section{Conclusion and perspectives}

This paper introduced the concept of calibrated likelihood functions. While this has been known for the binary case, \ie~when the likelihood function can be written in the form of a log-likelihood-ratio, we extended the definition of calibration to likelihood functions for any number of countable hypotheses. We also showed in Theorem \ref{prop:distribILRL}, that if, under one hypothesis, calibrated likelihood functions are normally distributed on the simplex, they are also normally distributed for the other hypotheses with some additional constraints on their parameter. This extends a result that has been known for decades in the binary case, \ie~for the weight-of-evidence and for calibrated log-likelihood-ratios. In order to do so, we have used the Aitchison geometry of the simplex, which has its origins in the field of compositional data analysis. It provides, to the probability simplex, an Euclidean vector space structure and recovers the additive form of the Bayes' rule. This allowed us to extend the concept of LLR, in a vector form, to any number of hypotheses; and therefore to extend the definition of the calibration, and the constraint on the distribution of calibrated likelihood functions.

The core of the paper is mainly conceptual and theoretical. However, an application of these results to machine learning has been presented. We introduced the Compositional Discriminant Analysis as a non-linear discriminant analysis where the discriminant subspace is designed to form a calibrated likelihood function over the classes. The distribution of the data in the discriminant space is constrained according to Theorem \ref{prop:distribILRL} and the discriminant mapping is learned through normalizing flow. This results in an easy-to-interpret and reliable discriminant analysis.

However, our contributions are more general and not limited to discriminant analysis. Theorem \ref{prop:distribILRL} gives a reference distribution for the likelihood functions to be calibrated. We therefore expect, in the future, several applications of our results to the calibration and the evaluation of probabilistic predictions in a multiple hypotheses and multiclass setting. Indeed, the results are not restricted to the space of likelihood functions since, given a prior, the posterior probability distribution and the likelihood function are isomorphic under a scale-invariant equivalence relation. Our results can therefore be pullbacked to the set of probabilistic prediction (by simply translating/perturbating the set of likelihood functions by the prior), which is more familiar to the calibration and statistical machine learning community.

Independently of the calibration, we expect that the use of the Aitchison geometry of the simplex will find many applications in machine learning, especially in multiclass settings, and as an alternative to the suboptimal one-vs-rest or one-vs-one approaches. As an example, the Aitchison geometry of the simplex has been used to extend the concept of Shapley value for explaining a probabilistic prediction in a multiclass context \citep{noe2024explaining}.

\appendix
\section{The distribution of calibrated LLRs}
\label{app:distributionLLR}

In this section we provide a detailed proof for Theorem \ref{prop:distrib_LLR}. The proof is similar as the one in \cite{leeuwen13_interspeech}. Alternative proofs can be found for the weight-of-evidence in \cite{good1985weight}, \cite{peterson1954}, and \cite{meester_slooten_2021}.\\

\noindent
{\bf Proof}.

Starting from the idempotence property, we have:
\begin{equation}
  \label{eq:idemppdf}
    l = \log \frac{f_{\mathcal{L}_1}\left(l \right)}{f_{\mathcal{L}_2}\left(l \right)} \iff f_{\mathcal{L}_2}\left(l\right) = e^{-l} f_{\mathcal{L}_1}\left(l \right).
\end{equation}
Let the density for the LLR under the first hypothesis be a Gaussian:
\begin{equation}
  \begin{aligned}
    l \mid H_1 &\sim \mathcal{N}(\mu, \sigma^2),~\text{where}~\mu \geq 0\\
    f_{\mathcal{L}_1}\left(l \right) &= \frac{1}{\sigma \sqrt{2 \pi}} \exp \left(-\frac{(l-\mu)^2}{2 \sigma^2} \right).
  \end{aligned}
\end{equation}
Thanks to Expression \ref{eq:idemppdf}, we have:
\begin{equation}
  \begin{aligned}
    f_{\mathcal{L}_2}\left(l \right) &= \frac{1}{\sigma \sqrt{2 \pi}} \exp \left(-\frac{(l-\mu)^2}{2 \sigma^2} \right) \exp \left( -l \right),\\
                  &= \frac{1}{\sigma \sqrt{2 \pi}} \exp \left( -\frac{\left( l- \left( \mu - \sigma^2 \right) \right)^2}{2 \sigma^2} \right) \exp \left( \frac{\sigma^2}{2}-\mu \right).
    \end{aligned}
\end{equation}
Since $f_{\mathcal{L}_2}\left( \cdot \right)$ is a probability density function, its integral is one:
\begin{equation}
  \begin{aligned}
    \int_{-\infty}^{+\infty} f_{\mathcal{L}_2}\left(l \right) dl = 1 &\iff \exp \left( \frac{\sigma^2}{2}-\mu \right) = 1,\\
    &\iff \sigma^2 = 2 \mu.
    \end{aligned}
  \end{equation}
  Therefore,
  \begin{equation}
    \begin{aligned}
      f_{\mathcal{L}_2}\left(l \right) &= \frac{1}{\sigma \sqrt{2 \pi}} \exp \left( -\frac{\left( l- \left(-\mu \right) \right)^2}{2 \sigma^2} \right),\\
      l \mid H_2 &\sim \mathcal{N}(-\mu, \sigma^2),
    \end{aligned}
  \end{equation}
  and $\sigma^2 = 2 \mu$. \hfill$\Box$

\section{The distribution of calibrated ILRLs}
\label{ap:distribILRL}

In this section, we provide a proof for Theorem \ref{prop:distribILRL}. Note that the use of the specific basis obtained with the Gram-Schmidt procedure in no way excludes the general aspect of the following results since Aitchison orthonormal bases are related through unitary transformations~\cite{egozcue2003isometric}.

Let's first recall Theorem \ref{prop:distribILRL}. Let $\bm{A} \in \mathcal{M}_{D-1,D-1}(\mathbb{R})$ be the following real square matrix:
\begin{equation}
  \label{eq:A}
\begin{aligned}
    \bm{A} &= \{\alpha_{ij}\}_{1\leq i,j \leq D-1}\\
    \alpha_{ij} &= \left\lbrace
  \begin{array}{@{}ll@{}}
    2\sqrt{\frac{i+1}{i}}, & \text{if}\ i=j \\
    \frac{2}{\sqrt{j(j+1)}}, & \text{if}\ j<i \\
    0, & \text{otherwise}
  \end{array}\right.
\end{aligned}
\end{equation}
and let $\bm{B} \in \mathcal{M}_{D-1,(D-1)^2}(\mathbb{R})$ be the block matrix:
\begin{equation}
  \label{eq:B1}
  \bm{B} = 
  \begin{bmatrix}
    \bm{B}^{(1)} & \bm{B}^{(2)} & \dots & \bm{B}^{(D-1)}
  \end{bmatrix}
\end{equation}
where $\bm{B}^{(b)} \in \mathcal{M}_{D-1, D-1}(\mathbb{R})$ is the $b$th block and is defined as:
\begin{equation}
  \label{eq:B2}
\begin{aligned}
    \bm{B}^{(b)} &= \{\beta^{(b)}_{ij}\}_{1\leq i,j \leq D-1}\\
    \beta^{(b)}_{ij} &= \left\lbrace
  \begin{array}{@{}ll@{}}
    \frac{b+1}{b}, & \text{if}\ i=j=b \\
    2\sqrt{\frac{i+1}{ib(b+1)}}, & \text{if}\ (i=j)\land(b<i) \\
    \frac{1}{jb\sqrt{(j+1)(b+1)}}, & \text{if}\ (b<i)\land(j<i) \\
    0, & \text{otherwise.}
  \end{array}\right.
\end{aligned}
\end{equation}
The idempotence property of the ILRL $\bm{l}$ leads to the following property on its distributions:\\

\textit{
  If $\bm{l} \mid H_1 \sim \mathcal{N}(\bm{\mu}_1,\bm{\Sigma})$, then $\forall i \in \left\{2,\dots D \right\}$, $~\bm{l} \mid H_i \sim \mathcal{N}(\bm{\mu}_i,\bm{\Sigma})$,
\textit{where}
\begin{equation*}
    \bm{\mu}_i = \bm{\mu}_1 - \bm{\Sigma} \bm{a}_{i-1} - \sum_{j=1}^{i-2} \frac{1}{j+1} \bm{\Sigma} \bm{a}_j,
\end{equation*}
and $\bm{\mu}_1 = \bm{A}^{-1}\bm{B} \vect(\bm{\Sigma})$, where the $(D-1)^2$-dimensional vector $\vect(\bm{\Sigma})$ is the vectorization of the covariance matrix~$\bm{\Sigma}$ and $\forall i \in \left\{1,\dots D-1 \right\}$,  $\bm{a}_i = \sqrt{\frac{i+1}{i}} \bm{e}_i$ where $\bm{e}_i$ is the $i$th vector of the standard canonical basis of $\mathbb{R}^{D-1}$.
}

\noindent
{\bf Proof}.
  Let's recall some notations. Let $\bm{w}_{\theta}(x) = \left[ f_{\theta_{\mathcal{X}_i}}(x) \right]_{1 \leq i \leq D}$ be the likelihood vector and $\tilde{\bm{w}}_{\theta}(x)= \ilr \left( \bm{w}_{\theta}(x) \right) = \bm{l}_{\theta}$ its ILR transformation. With Equation \ref{eq:generalilrcomponent}, the $i$th ILR component of a likelihood vector can be written as:
\begin{equation}
    l_{\theta i} = \tilde{w}_{\theta}(x)_i = \frac{1}{\sqrt{i\,(i+1)}} \log \left( \frac{\prod\limits_{j = 1}^{i} f_{\theta_{\mathcal{X}_j}}(x)}{( f_{\theta_{\mathcal{X}_{i+1}}}(x))^{i}} \right).
\end{equation}

Using the idempotence property we can replace every $f_{\theta_{\mathcal{X}_i}}(x)$ by $f_{\mathcal{L}_i}(\bm{l}_{\theta})$:
\begin{equation}
    l_{\theta i} = \frac{1}{\sqrt{i\,(i+1)}} \log \left( \frac{\prod\limits_{j = 1}^{i} f_{\mathcal{L}_j}(\bm{l}_{\theta}) }{\left( f_{\mathcal{L}_{i+1}}(\bm{l}_{\theta})  \right)^{i}} \right)\\
    \label{eq:generalli}
\end{equation}
After rewriting this expression and setting $\bm{a}_i = \sqrt{\frac{i+1}{i}} \bm{e}_i$ where $\bm{e}_i$ is the $i$th vector of the standard canonical basis for $\mathbb{R}^{D-1}$~\ie~with zero everywhere except with 1 at the $i$th position, we get:
\begin{equation}
    f_{\mathcal{L}_{i+1}}(\bm{l}_{\theta}) = \exp \left( -\bm{a}_i^T \bm{l} \right) \sqrt[i]{ \prod_{j=1}^{i} f_{\mathcal{L}_{j}}(\bm{l}_{\theta}) }.
    \label{eq:recursiveform}
\end{equation}
We thus have a recursive way to get any $ f_{\mathcal{L}_{i}}$ from $ f_{\mathcal{L}_{1}}$. With $\bm{l} \sim \mathcal{N}\left(\bm{\mu}_1 \mid \bm{\Sigma}\right)$ and:
  \begin{equation}
    \label{eq:ilrldensity1}
      f_{\mathcal{L}_1}(\bm{l}) = \frac{1}{(2\pi)^{\frac{D-1}{2}} |\bm{\Sigma}|^\frac{1}{2}} \exp \left( - \frac{1}{2}(\bm{l}-\bm{\mu}_1)^T \bm{\Sigma}^{-1}(\bm{l}-\bm{\mu}_1) \right),
\end{equation}
we can use Equation \ref{eq:recursiveform} to recursively compute the densities  $f_{\mathcal{L}_i}$ for $i\in \left\{2,\dots D \right\}$.\\

The main idea of the proof is to show---by induction and using the recursive relation of Expression \ref{eq:recursiveform}---that: 
\begin{equation}
  \begin{aligned}
    &\text{for all integer $D \geq 2$ we have:}\\
    &\forall i \in \left\{1,\dots D-1 \right\},\\
    &f_{\mathcal{L}_{i+1}}(\bm{l}) = \frac{1}{(2\pi)^{\frac{D-1}{2}} |\bm{\Sigma}|^\frac{1}{2}} \exp \left(
    -\frac{1}{2}(\bm{l} - \bm{\mu}_{i+1})^T \bm{\Sigma}^{-1}(\bm{l} - \bm{\mu}_{i+1}) \right),\\
    &\bm{\mu}_{i+1} = \frac{1}{i} \sum_{j=1}^{i}\bm{\mu}_j - \bm{\Sigma} \bm{a}_i,\\
    &\bm{\mu}_{i+1}^T \bm{\Sigma}^{-1} \bm{\mu}_{i+1} = \bm{\mu}_1^T \bm{\Sigma}^{-1} \bm{\mu}_1.
\end{aligned}
\label{eq:gen}
\end{equation}

\subsection*{The base case:}
From Equation \ref{eq:recursiveform} and Equation \ref{eq:ilrldensity1} we get:
\begin{equation}
    \begin{aligned}
    f_{\mathcal{L}_2}(\bm{l}) &=  \exp \left( -\bm{a}_1^T \bm{l} \right) f_{\mathcal{L}_1}(\bm{l}),\\
    &= \frac{1}{(2\pi)^{\frac{D-1}{2}} |\bm{\Sigma}|^\frac{1}{2}} \exp \left( - \frac{1}{2}(\bm{l}-\bm{\mu}_1)^T \bm{\Sigma}^{-1}(\bm{l}-\bm{\mu}_1) - \bm{a}_1^T\bm{l} \right),\\
                    &= \frac{1}{(2\pi)^{\frac{D-1}{2}} |\bm{\Sigma}|^\frac{1}{2}} \exp \left( -\frac{1}{2} \bm{l}^T \bm{\Sigma}^{-1} \bm{l} + \bm{l}^T \bm{\Sigma}^{-1}\left(\bm{\mu}_1 - \bm{\Sigma} \bm{a}_1 \right) -\frac{1}{2}\bm{\mu}_1^T \bm{\Sigma}^{-1}\bm{\mu}_1\right),\\
    &= \frac{1}{(2\pi)^{\frac{D-1}{2}} |\bm{\Sigma}|^\frac{1}{2}} \exp \left( -\frac{1}{2}(\bm{l}-\bm{\mu}_2)^T \bm{\Sigma}^{-1}(\bm{l}-\bm{\mu}_2)\right) \exp \left( \frac{1}{2}\bm{\mu}_2^T\bm{\Sigma}^{-1}\bm{\mu}_2 - \frac{1}{2}\bm{\mu}_1^T\bm{\Sigma}^{-1}\bm{\mu}_1    \right),
    \end{aligned}
  \end{equation}
  where $\bm{\mu}_2 = \bm{\mu}_1 - \bm{\Sigma} \bm{a}_1$. Since $f_{\mathcal{L}_2}$ is a probability density function, its integral is one:
\begin{equation}
\begin{aligned}
    &\int_{\bm{l}\in \mathbb{R}^{D-1}}  f_{\mathcal{L}_2}(\bm{l}) d\bm{l} = 1\\
    &\iff \exp \left( \frac{1}{2}\bm{\mu}_2^T\bm{\Sigma}^{-1}\bm{\mu}_2 - \frac{1}{2}\bm{\mu}_1^T\bm{\Sigma}^{-1}\bm{\mu}_1    \right) = 1,\\
    &\iff  \bm{\mu}_2^T\bm{\Sigma}^{-1}\bm{\mu}_2 = \bm{\mu}_1^T\bm{\Sigma}^{-1}\bm{\mu}_1.\\
    \label{eq:cond1}
\end{aligned}
\end{equation}
We just showed that $\bm{l} \mid H_2 \sim \mathcal{N} \left( \bm{\mu}_2, \bm{\Sigma} \right)$ where $\bm{\mu}_2 = \bm{\mu}_1 - \bm{\Sigma} \bm{a}_1$ and $\bm{\mu}_2^T\bm{\Sigma}^{-1}\bm{\mu}_2 = \bm{\mu}_1^T\bm{\Sigma}^{-1}\bm{\mu}_1$.
\subsection*{The induction step:}

Let's assume that for an integer $K$ we have:
\begin{equation}
\begin{aligned}
  &\forall i\in \left\{1,\dots K-1 \right\},\\
  & f_{\mathcal{L}_{i+1}}(\bm{l}) = \frac{1}{(2\pi)^{\frac{D-1}{2}} |\bm{\Sigma}|^\frac{1}{2}} \exp \left(
    -\frac{1}{2}(\bm{l} - \bm{\mu}_{i+1})^T \bm{\Sigma}^{-1}(\bm{l} - \bm{\mu}_{i+1}) \right)\\
    & \bm{\mu}_{i+1} = \frac{1}{i} \sum_{j=1}^{i}\bm{\mu}_j - \bm{\Sigma} \bm{a}_i,\\
    & \bm{\mu}_{i+1}^T \bm{\Sigma}^{-1} \bm{\mu}_{i+1} = \bm{\mu}_1^T \bm{\Sigma}^{-1} \bm{\mu}_1.
\end{aligned}
\label{eq:K}
\end{equation}
Let's show that this still holds for $K+1$. Using Equation \ref{eq:recursiveform} we can write:
\begin{equation}
\begin{aligned}
        &f_{\mathcal{L}_{K+1}}(\bm{l}) = \exp \left( -\bm{a}_{K}^T \bm{l} \right) \sqrt[K]{ \prod_{j=1}^{K}f_{\mathcal{L}_j}(\bm{l})},\\
        &~~~= \exp \left( -\bm{a}_{K}^T \bm{l} \right) \frac{1}{(2\pi)^{\frac{D-1}{2}} |\bm{\Sigma}|^\frac{1}{2}} \sqrt[K]{\prod_{j=1}^{K} \exp \left( -\frac{1}{2}(\bm{l}-\bm{\mu}_j)^T \bm{\Sigma}^{-1} (\bm{l}-\bm{\mu}_j)
        \right)},\\
        &~~~= \frac{1}{(2\pi)^{\frac{D-1}{2}} |\bm{\Sigma}|^\frac{1}{2}}  \exp \left(-\bm{a}_{K}^T \bm{l}  -\frac{1}{2K}\sum_{j=1}^{K}  (\bm{l}-\bm{\mu}_j)^T \bm{\Sigma}^{-1} (\bm{l}-\bm{\mu}_j)\right),\\
                            &~~~= \frac{1}{(2\pi)^{\frac{D-1}{2}} |\bm{\Sigma}|^\frac{1}{2}}  \exp \left(-\bm{a}_{K}^T \bm{l}  -\frac{1}{2K}\sum_{j=1}^{K}  \left( \bm{l}^T \bm{\Sigma}^{-1} \bm{l} + \bm{\mu}_j^T \bm{\Sigma}^{-1} \bm{\mu}_j - 2\bm{\mu}_j^T\bm{\Sigma}^{-1}\bm{l} \right)\right).
  \end{aligned}
\end{equation}
Since for all $j \in \left\{1,\dots K \right\}$, $\bm{\mu}_j^T \bm{\Sigma}^{-1} \bm{\mu}_j = \bm{\mu}_1^T \bm{\Sigma}^{-1} \bm{\mu}_1$, we have:
\begin{equation}
    \begin{aligned}
        &f_{\mathcal{L}_{K+1}}(\bm{l})= \frac{1}{(2\pi)^{\frac{D-1}{2}} |\bm{\Sigma}|^\frac{1}{2}}  \exp \left( -\bm{a}_{K}^T \bm{l} - \frac{1}{2}\bm{l}^T \bm{\Sigma}^{-1}\bm{l} - \frac{1}{2}\bm{\mu}_1^T \bm{\Sigma}^{-1}\bm{\mu}_1 + \frac{1}{K} \left( \sum_{j=1}^{K}\bm{\mu}_j \right)^T\bm{\Sigma}^{-1}\bm{l} \right),\\
        &~~~= \frac{1}{(2\pi)^{\frac{D-1}{2}} |\bm{\Sigma}|^\frac{1}{2}}  \exp \left(- \frac{1}{2}\bm{l}^T \bm{\Sigma}^{-1}\bm{l} - \frac{1}{2}\bm{\mu}_1^T \bm{\Sigma}^{-1}\bm{\mu}_1 + \bm{l}^T\bm{\Sigma}^{-1}\left(\frac{1}{K} \left( \sum_{j=1}^{K}\bm{\mu}_j \right) -\bm{\Sigma}\bm{a}_{K}\right) \right).
\end{aligned}
\end{equation}
Setting $\bm{\mu}_{K+1} = \frac{1}{K} \displaystyle  \sum_{j=1}^{K}\bm{\mu}_j -\bm{\Sigma}\bm{a}_{K}$ we get:
\begin{equation}
  \begin{aligned}
    f_{\mathcal{L}_{K+1}}(\bm{l}) &= \frac{1}{(2\pi)^{\frac{D-1}{2}} |\bm{\Sigma}|^\frac{1}{2}}  \exp \left(-\frac{1}{2}( \bm{l} - \bm{\mu}_{K+1} )^T\bm{\Sigma}^{-1} ( \bm{l} - \bm{\mu}_{K+1} ) \right)\\
    &~~~\times \exp \left( \frac{1}{2}\bm{\mu}_{K+1}^T \bm{\Sigma}^{-1}\bm{\mu}_{K+1} - \frac{1}{2}\bm{\mu}_{1}^T \bm{\Sigma}^{-1}\bm{\mu}_{1} \right),
 \end{aligned}
\end{equation}
Since $f_{\mathcal{L}_{K+1}}$ is a probability density function, its integral is one, which leads to:
\begin{equation}
    \bm{\mu}_{K+1}^T \bm{\Sigma}^{-1}\bm{\mu}_{K+1} = \bm{\mu}_{1}^T \bm{\Sigma}^{-1}\bm{\mu}_{1}.
\end{equation}
Therefore, \ref{eq:K} holds also for $K+1$. We hence have proved by induction the expressions \ref{eq:gen}.

\subsection*{A general formula for the means:}

We will here proof by induction the following expression for the derivation of the mean vectors: 
\begin{equation}
  \begin{aligned}
&\text{for all integer $D \geq 2$:}\\
&\forall i \in \left\{2,\dots D \right\},\\
    &\bm{\mu}_i = \bm{\mu}_1 - \bm{\Sigma} \bm{a}_{i-1} - \sum_{j=1}^{i-2} \frac{1}{j+1} \bm{\Sigma} \bm{a}_j,\text{ where $\bm{a}_0 = \bm{0}$ the zero vector.}
  \end{aligned}
    \label{eq:genmean}
\end{equation}
The base case is straightforward so we provide only the induction step. Let's assume that the expression is true for an integer $K$:
\begin{equation}
  \begin{aligned}
    &\forall i \in \left\{2,\dots K \right\},\\
    &\bm{\mu}_i = \bm{\mu}_1 - \bm{\Sigma} \bm{a}_{i-1} - \sum_{j=1}^{i-2} \frac{1}{j+1} \bm{\Sigma} \bm{a}_j.
  \end{aligned}
    \label{eq:assume}
\end{equation}
Let's show that this holds also for $K+1$. From Expression \ref{eq:gen} that we proofed above, we know that:
\begin{equation}
  \bm{\mu}_{K+1} = - \bm{\Sigma} \bm{a}_K +  \frac{1}{K} \sum_{j=1}^{K}\bm{\mu}_j,\\
\end{equation}
we replace $\bm{\mu}_j$ according to Expression \ref{eq:assume}:
\begin{equation}
\begin{aligned}
\bm{\mu}_{K+1} &= - \bm{\Sigma} \bm{a}_K + \frac{1}{K} \sum_{j=1}^{K} \left( \bm{\mu}_1 - \bm{\Sigma}\bm{a}_{j-1} - \sum_{k=1}^{j-2} \frac{1}{k+1} \bm{\Sigma} \bm{a}_k\right),\\
    &=\bm{\mu}_1 - \bm{\Sigma} \bm{a}_K
    - \frac{1}{K} \sum_{j=2}^{K} \bm{\Sigma} \bm{a}_{j-1}
    - \frac{1}{K} \sum_{j=3}^{K} \sum_{k=1}^{j-2} \frac{1}{k+1} \bm{\Sigma} \bm{a}_k,\\
    &=\bm{\mu}_1 - \bm{\Sigma} \bm{a}_K
    - \frac{1}{K} \sum_{j=1}^{K-1} \bm{\Sigma} \bm{a}_{j}
    - \frac{1}{K} \sum_{j=1}^{K-2} \frac{K-1-j}{j+1} \bm{\Sigma} \bm{a}_j,\\
    &=\bm{\mu}_1 - \bm{\Sigma} \bm{a}_K
    - \frac{1}{K} \bm{\Sigma} \bm{a}_{K-1}
      - \frac{1}{K} \sum_{j=1}^{K-2} \left( \bm{\Sigma} \bm{a}_j + \frac{K-1-j}{j+1}\bm{\Sigma} \bm{a}_j \right),
\end{aligned}
\end{equation}
\begin{equation}
  \begin{aligned}
    &=\bm{\mu}_1 - \bm{\Sigma} \bm{a}_K
    - \frac{1}{K} \bm{\Sigma} \bm{a}_{K-1}
    - \sum_{j=1}^{K-2} \frac{1}{j+1}\bm{\Sigma} \bm{a}_j,\\
    &=\bm{\mu}_1 - \bm{\Sigma} \bm{a}_K
    - \sum_{j=1}^{K-1} \frac{1}{j+1}\bm{\Sigma} \bm{a}_j,
\end{aligned}
\end{equation}
Hence,
\begin{equation}
  \begin{aligned}
  &\forall i \in \left\{2,\dots K+1 \right\},\\
  &\bm{\mu}_i = \bm{\mu}_1 - \bm{\Sigma} \bm{a}_{i-1} - \sum_{j=1}^{i-2} \frac{1}{j+1} \bm{\Sigma} \bm{a}_j,
  \end{aligned}
\end{equation}
The general expression \ref{eq:genmean} has therefore been proved by induction.

\subsection*{About matrices A and B}

In Proposition \ref{prop:distribILRL}, the mean vector $\bm{\mu}_1$ is expressed in terms of the covariance matrix $\bm{\Sigma}$ and two constant matrices $\bm{A}$ and $\bm{B}$ as follow:
\begin{equation}
  \bm{\mu}_1 = \bm{A}^{-1 }\bm{B} \vect(\bm{\Sigma}),
\end{equation}
where $\bm{A} \in \mathcal{M}_{D-1,D-1}(\mathbb{R})$ and is defined as:
\begin{equation}
\begin{aligned}
    \bm{A} &= \{\alpha_{ij}\}_{1\leq i,j \leq D-1}\\
    \alpha_{ij} &= \left\lbrace
  \begin{array}{@{}ll@{}}
    2\sqrt{\frac{i+1}{i}}, & \text{if}\ i=j \\
    \frac{2}{\sqrt{j(j+1)}}, & \text{if}\ j<i \\
    0, & \text{otherwise}
  \end{array}\right.
\end{aligned}
\end{equation}
and where $\bm{B} \in \mathcal{M}_{D-1,(D-1)^2}(\mathbb{R})$ is a block matrix:
\begin{equation}
  \bm{B} = 
  \begin{bmatrix}
    \bm{B}^{(1)} & \bm{B}^{(2)} & \dots & \bm{B}^{(D-1)}
  \end{bmatrix}
\end{equation}
where $\bm{B}^{(b)} \in \mathcal{M}_{D-1, D-1}(\mathbb{R})$ is the $b$th block and is defined as:
\begin{equation}
\begin{aligned}
    \bm{B}^{(b)} &= \{\beta^{(b)}_{ij}\}_{1\leq i,j \leq D-1}\\
    \beta^{(b)}_{ij} &= \left\lbrace
  \begin{array}{@{}ll@{}}
    \frac{b+1}{b}, & \text{if}\ i=j=b \\
    2\sqrt{\frac{i+1}{ib(b+1)}}, & \text{if}\ (i=j)\land(b<i) \\
    \frac{1}{jb\sqrt{(j+1)(b+1)}}, & \text{if}\ (b<i)\land(j<i) \\
    0, & \text{otherwise}
  \end{array}\right.
\end{aligned}
\end{equation}
In this paragraph, we show how these matrices are derived. The following system of equations, that comes from the expressions \ref{eq:gen}, can be written in a matrix form:
\begin{equation}
   \begin{aligned}
     &\forall~i \in \left\{1,\dots D-1 \right\},\\
     &\bm{\mu}_{i+1}^T \bm{\Sigma}^{-1} \bm{\mu}_{i+1} = \bm{\mu}_1^T \bm{\Sigma}^{-1} \bm{\mu}_1.
  \end{aligned}
\end{equation}
Using the general expression \ref{eq:genmean} for the means, the system of equations becomes:
\begin{equation}
    \begin{aligned}
    &\forall i \in \left\{1,\dots D-1 \right\},\\
    &2 \bm{a}_i^T\bm{\mu}_1 + 2\bm{\mu}_1^T\bm{\Sigma}^{-1} \sum_{j=1}^{i-1} \frac{1}{j+1}\bm{\Sigma}\bm{a}_j = \bm{a}_i^T \bm{\Sigma} \bm{a}_i + 2 \bm{a}_i^T \sum_{j=1}^{i-1}\frac{1}{j+1}\bm{\Sigma}\bm{a}_j + \sum_{j=1}^{i-1} \sum_{k=1}^{i-1} \frac{1}{(j+1)(k+1)} \bm{a}_j^T \bm{\Sigma} \bm{a}_k,
    \end{aligned}
\end{equation}
Since $\bm{x}^T \bm{\Sigma} \bm{y} = \vect (\bm{\Sigma})^T \vect (\bm{x} \bm{y}^T)$ and by setting $\bm{\theta}_{\Sigma} = \vect (\bm{\Sigma})$ we get:
\begin{equation}
    \begin{aligned}
    &\forall i \in \left\{1,\dots D-1 \right\},\\
    &\left(2\bm{a}_i + 2 \sum_{j=1}^{i-1} \frac{1}{j+1}\bm{a}_j \right)^T \bm{\mu}_1\\
      &~~~~~~~~~=  \left( \vect (\bm{a}_i \bm{a}_i^T ) + 2 \sum_{j=1}^{i-1}\frac{1}{j+1} \vect (\bm{a}_i  \bm{a}_j^T ) + \sum_{j=1}^{i-1} \sum_{k=1}^{i-1} \frac{1}{(j+1)(k+1)} \vect ( \bm{a}_j \bm{a}_k^T ) \right)^T \bm{\theta}_{\bm{\Sigma}}.
    \end{aligned}
    \label{eq:system}
\end{equation}

\noindent
$\bm{a}_i \bm{a}_j^T$ is a $(D-1)\times(D-1)$ matrix with zero everywhere except the element at the $i$th row and $j$th column which is $\sqrt{\frac{(i+1)(j+1)}{ij}}$. Its vectorization is therefore the $(D-1)^2$-dimensional vector with zero everywhere except the $\left( (j-1)(D-1)+i \right)$th element which is $\sqrt{\frac{(i+1)(j+1)}{ij}}$. Let's now rewrite this system in a matrix form:
\begin{equation}
    \bm{A}\bm{\mu}_1 = \bm{B} \bm{\theta}_\Sigma,
    \label{eq:system2}
\end{equation}

\noindent
where $\bm{A} \in M_{D-1,D-1}(\mathbb{R})$, $\bm{B} \in M_{D-1,(D-1)^2}(\mathbb{R})$. In \ref{eq:system}, the vector on the left side of $\bm{\mu}$ is the $i$th row of the matrix $\bm{A}$ and the vector on the left side of $\bm{\theta}_{\bm{\Sigma}}$ is the $i$th row of $\bm{B}$. This is straightforward that $\bm{A}$ is triangular with diagonal elements $2\sqrt{\frac{i+1}{i}}$ for $ i \in \left\{1,\dots D-1 \right\}$. Consequently, its determinant is non zero, and therefore $\bm{A}$ is invertible. The mean vector $\bm{\mu}_1$ can therefore be written in terms of the variances and covariances as follow:
\begin{equation}
    \bm{\mu}_1 = \bm{A}^{-1}\bm{B} \bm{\theta}_\Sigma = \bm{A}^{-1}\bm{B} \vect (\bm{\Sigma}).
    \label{eq:means}
\end{equation}\hfill$\Box$

\section{The covariance matrix of the ILRL distribution and the divergences}
\label{ap:covdiv}
In Section \ref{sec:distribILRL}, we have seen that the covariance matrix $\bm{\Sigma}$---which is the only parameter of the densities of normally distributed ILRLs---can be expressed in terms of the Kullback-Leibler divergences between each density. In this section, we provide details of the computation. The divergence between the density for the hypothesis $i$ and the density for the hypothesis $j$ can be written as:
\begin{equation}
\begin{aligned}
    d_{i,j} &= \frac{1}{2}(\bm{\mu}_i - \bm{\mu}_j)^T\bm{\Sigma}^{-1}(\bm{\mu}_i - \bm{\mu}_j)\\
            &=\frac{1}{2}\left( \bm{\mu}_j^T \bm{\Sigma}^{-1} \bm{\mu}_j -2 \bm{\mu}_i^T \bm{\Sigma}^{-1} \bm{\mu}_j + \bm{\mu}_i^T \bm{\Sigma}^{-1} \bm{\mu}_i \right),
\end{aligned}
\end{equation}
and since $\forall i \in \left\{1,\dots D-1 \right\}$, $\bm{\mu}_{i+1}^T \bm{\Sigma}^{-1} \bm{\mu}_{i+1} = \bm{\mu}_{1}^T \bm{\Sigma}^{-1} \bm{\mu}_{1}$ (see the Appendix \ref{ap:distribILRL}), 
\begin{equation}
    d_{i,j} = \bm{\mu}_1^T \bm{\Sigma}^{-1} \bm{\mu}_1 - \bm{\mu}_i^T \bm{\Sigma}^{-1} \bm{\mu}_j.
\end{equation}
Replacing $\bm{\mu}_i$ and $\bm{\mu}_j$ with the expression given in Theorem \ref{prop:distribILRL} we get:
\begin{equation}
\begin{aligned}
    d_{i,j}&= \bm{\zeta}_{i,j}^T \bm{\mu}_1 - \bm{\eta}_{i,j}^T \vect (\bm{\Sigma})
\end{aligned}
\end{equation}
where
\begin{equation}
\begin{aligned}
    \bm{\zeta}_{i,j} &= \left( \bm{a}_{i-1} + \bm{a}_{j-1} +\sum_{k=1}^{i-2} \frac{1}{k+1} \bm{a}_{k} + \sum_{k=1}^{j-2} \frac{1}{k+1} \bm{a}_{k} \right),\\
  \bm{\eta}_{i,j} &= \Bigg( \vect(\bm{a}_{i-1} \bm{a}_{j-1}^T) + \sum_{k=1}^{j-2}\frac{1}{k+1}\vect(\bm{a}_{i-1} \bm{a}_k^T) + \sum_{k=1}^{i-2}\frac{1}{k+1}\vect(\bm{a}_{j-1} \bm{a}_k^T)\\
  &~~~~~+ \sum_{k=1}^{i-2}\sum_{k=1}^{j-2} \frac{1}{(k+1)(l+1)}\vect(\bm{a}_k \bm{a}_l^T) \Bigg).
\end{aligned}
\end{equation}
When $2\leq i,j\leq D$ and $i=j$, these vectors are respectively the $(i-1)$th row of $\bm{A}$ and the $(i-1)$th row of $\bm{B}$. Since $\bm{\mu}_1 = \bm{A}^{-1} \bm{B} \vect (\bm{\Sigma})$, the divergences can be written as follow:
\begin{equation}
  \forall i \in \left\{1,\dots D-1 \right\},~
  \forall j \in \left\{i+1,\dots D \right\},~~~
  d_{i,j} = \left(\bm{\zeta}_{i,j}^T\bm{A}^{-1}\bm{B} - \bm{\eta}_{i,j}^T\right) \vect (\bm{\Sigma}).
\end{equation}
Let $\vech$ be the half-vectorization of a matrix and $\vech_{\neg\smallsetminus}$ be the half-vectorization without the diagonal elements. The above set of equations can therefore be written in the following matrix form:
\begin{equation}
  \label{eq:sigma_to_divs}
  \begin{aligned}
    \vech_{\neg\smallsetminus}(\bm{\Delta}) &= \underbrace{\begin{bmatrix} \bm{\zeta}_{1,2}^T\bm{A}^{-1}\bm{B} - \bm{\eta}_{1,2}^T \\
    \bm{\zeta}_{1,3}^T\bm{A}^{-1}\bm{B} - \bm{\eta}_{1,3}^T\\
    \vdots\\
    \bm{\zeta}_{N-1,N}^T\bm{A}^{-1}\bm{B} - \bm{\eta}_{D-1,D}^T
                                                        \end{bmatrix} \bm{D}_{D-1}}_{\bm{M}} \vech (\bm{\Sigma}),\\
    \vech_{\neg\smallsetminus}(\bm{\Delta}) &= \bm{M} \vech(\bm{\Sigma}),
  \end{aligned}
\end{equation}
where $\bm{D}_{N-1}$ is the duplication matrix \citep{magnus99} such that $\vect (\bm{\Sigma}) = \bm{D}_{N-1} \vech (\bm{\Sigma})$ and $\bm{M} \in \mathcal{M}_{\frac{D(D-1)}{2} \times \frac{D(D-1)}{2}}(\mathbb{R})$ is a real square matrix.

\section{Proof that the base space's first dimensions  form the ILRL}
\label{ap:ilrlzdims}

This section gives a proof for Lemma \ref{prop:ilrlzdims}. It shows that with the class-conditional distributions as defined in Equation \ref{eq:classcondbase}, the first $D-1$ dimensions of $\bm{z} \in \mathcal{Z}$ form the ILRL.

\noindent
{\bf Proof}.
The $i$th component of the ILRL vector of $\bm{z}$ is given by:
\begin{equation}
  \begin{aligned}
    \forall i \in \left\{1,\dots D-1 \right\}, ~~~l_i(\bm{z}) &= \frac{1}{\sqrt{i(i+1)}} \log \left( \frac{\prod\limits_{j = 1}^{i} f_{\mathcal{Z}_{j}}(\bm{z})}{f_{\mathcal{Z}_{i+1}}(\bm{z})^{i}} \right)\\
    &= \frac{1}{\sqrt{i(i+1)}} \log \left( \frac{ \prod\limits_{j = 1}^{i} \exp \left( -\frac{1}{2} \left( \bm{z} - \bm{m}_{j} \right)^T \bm{C}^{-1} \left(  \bm{z} - \bm{m}_{j}  \right) \right) }{ \exp \left( -\frac{i}{2}\left( \bm{z} - \bm{m}_{i+1} \right)^T \bm{C}^{-1} \left(  \bm{z} - \bm{m}_{i+1}  \right)  \right) } \right),
  \end{aligned}
\end{equation}
where $\bm{m}_i$ and $\bm{C}$ are respectively the mean vector and the covariance matrix as defined in the following of Equation \ref{eq:classcondbase},
\begin{equation}
  \begin{aligned}
    l_i(\bm{z}) &= \frac{1}{\sqrt{i(i+1)}} \left( \sum_{j=1}^{i} \left( -\frac{1}{2} \left( \bm{z} - \bm{m}_{j} \right)^T \bm{C}^{-1} \left(  \bm{z} - \bm{m}_{j}  \right) \right) + \frac{i}{2} \left( \bm{z} - \bm{m}_{i+1} \right)^T \bm{C}^{-1} \left(  \bm{z} - \bm{m}_{i+1}  \right) \right),\\
                &= \frac{1}{\sqrt{i(i+1)}} \left( \sum_{j=1}^{i} \left( \bm{m}_j^T\bm{C}^{-1}\bm{z} -\frac{1}{2} \bm{m}_j^T\bm{C}^{-1}\bm{m}_j \right) + \frac{i}{2} \bm{m}_{i+1}^T\bm{C}^{-1}\bm{m}_{i+1} - i\bm{m}_{i+1}^T \bm{C}^{-1} \bm{z} \right),\\
                &= \frac{1}{\sqrt{i(i+1)}} \left( \sum_{j=1}^{i} \left( \bm{\mu}_j^T\bm{\Sigma}^{-1}\bm{z}_{1:D-1} -\frac{1}{2} \bm{\mu}_j^T\bm{\Sigma}^{-1}\bm{\mu}_j \right) + \frac{i}{2} \bm{\mu}_{i+1}^T\bm{\Sigma}^{-1}\bm{\mu}_{i+1} - i\bm{\mu}_{i+1}^T \bm{\Sigma}^{-1} \bm{z}_{1:D-1} \right),
  \end{aligned}
\end{equation}
where $\bm{z}_{1:D-1} = \left[ z_1, z_2, \dots z_{D-1} \right]^T$ is the vector of the first $D-1$ components of $\bm{z}$. Since $\forall i \in \left\{1,\dots D \right\}$, $\bm{\mu}_i^T \bm{\Sigma} \bm{\mu}_i = \bm{\mu}_1^T \bm{\Sigma} \bm{\mu}_1$ (see Appendix \ref{ap:distribILRL}), we have:
\begin{equation}
  \begin{aligned}
    l_i(\bm{z}) = \frac{1}{\sqrt{i(i+1)}}\left( \sum_{j=1}^i \left( \bm{\mu}_j^T\bm{\Sigma}^{-1}\bm{z}_{1:D-1} \right) - i \bm{\mu}_{i+1}^T \bm{\Sigma}^{-1} \bm{z}_{1:D-1} \right),
  \end{aligned}
\end{equation}
using Equation \ref{eq:genmean2}, we get:
\begin{equation}
  \begin{aligned}
    l_i(\bm{z}) &= \frac{1}{\sqrt{i(i+1)}} \Bigg( \sum_{j=1}^i \left( \bm{A}^{-1}\bm{B} \vect(\bm{\Sigma}) - \bm{\Sigma} \bm{a}_{j-1} - \sum_{k=1}^{j-2} \frac{1}{k+1} \bm{\Sigma} \bm{a}_k \right)^T\bm{\Sigma}^{-1}\bm{z}_{1:D-1}\\
                &~~~~~~~~~~~~~~~~~~~~~~~-i \left( \bm{A}^{-1}\bm{B} \vect(\bm{\Sigma}) - \bm{\Sigma} \bm{a}_{i} - \sum_{k=1}^{i-1} \frac{1}{k+1} \bm{\Sigma} \bm{a}_k \right)^T \bm{\Sigma}^{-1} \bm{z}_{1:D-1} \Bigg),\\\
                &= \frac{1}{\sqrt{i(i+1)}} \Bigg( \sum_{j=1}^{i} \left( -\bm{a}_{j-1}^T\bm{z}_{1:D-1} - \sum_{k=1}^{j-2}\frac{1}{k+1} \bm{a}_k^T\bm{z}_{1:D-1} \right)\\
    &~~~~~~~~~~~~~~~~~~~~~~~+ i \sum_{k=1}^{i-1}\left( \frac{1}{k+1} \bm{a}_k^T\bm{z}_{1:D-1}\right) + i\bm{a}_i^T\bm{z}_{1:D-1} \Bigg).
  \end{aligned}
\end{equation}
In the next paragraph, we will see that:
\begin{equation}
  \label{eq:equal0}
  \sum_{j=1}^{i} \left( -\bm{a}_{j-1}^T\bm{z}_{1:D-1} - \sum_{k=1}^{j-2}\frac{1}{k+1} \bm{a}_k^T\bm{z}_{1:D-1} \right) + i \sum_{k=1}^{i-1}\left( \frac{1}{k+1} \bm{a}_k^T\bm{z}_{1:D-1}\right) = 0.
\end{equation}
We therefore have:
\begin{equation}
  \begin{aligned}
    l_i(\bm{z}) &= \frac{i}{\sqrt{i(i+1)}} \bm{a}_i^T\bm{z}_{1:D-1} = \frac{i}{\sqrt{i(i+1)}} \sqrt{\frac{i+1}{i}} \bm{e}_i^T \bm{z}_{1:D-1} = \bm{e}_i^T\bm{z}_{1:D-1},\\
    &= z_i,
  \end{aligned}
\end{equation}
the $i$th component of the ILRL is therefore the $i$th component of $\bm{z}$ for all $i\in \left\{1,\dots D-1 \right\}$.

\subsection*{Proof of Expression \ref{eq:equal0}:}
In the following, we will show by induction that Expression \ref{eq:equal0} is true for all $i \in \left\{1,\dots D-1 \right\}$ which is equivalent to show that:
\begin{equation}
  \forall i \in \left\{1,\dots D-1 \right\},~~~~~ \sum_{j=1}^{i} \left( -\bm{a}_{j-1} - \sum_{k=1}^{j-2}\frac{1}{k+1} \bm{a}_k \right) + i \sum_{k=1}^{i-1}\left( \frac{1}{k+1} \bm{a}_k\right) = \bm{0}.
\end{equation}
The base case of the proof by induction is straightforward, we therefore focus only on the induction step. We assume that the expression is true for a $i=n$ where $n \in \mathbb{N}$:
\begin{equation}
  \label{eq:inductionn}
  \sum_{j=1}^{n} \left( -\bm{a}_{j-1} - \sum_{k=1}^{j-2}\frac{1}{k+1} \bm{a}_k \right) + n \sum_{k=1}^{n-1}\left( \frac{1}{k+1} \bm{a}_k\right) = \bm{0}.
\end{equation}
Let's show this is still true for $i=n+1$:
\begin{equation}
  \begin{aligned}
    &\sum_{j=1}^{n+1} \left( -\bm{a}_{j-1} - \sum_{k=1}^{j-2}\frac{1}{k+1} \bm{a}_k \right) + (n+1) \sum_{k=1}^{n}\left( \frac{1}{k+1} \bm{a}_k\right)\\
    &~~~~~= \sum_{j=1}^{n} \left( -\bm{a}_{j-1}-\sum_{k=1}^{j-2}\frac{1}{k+1}\bm{a}_k \right) + n \sum_{k=1}^{n-1} \left( \frac{1}{k+1} \bm{a}_k \right)\\
    &~~~~~~~~~~- \bm{a}_n - \sum_{k=1}^{n-1} \left(\frac{1}{k+1}\bm{a}_k \right)+ \sum_{k=1}^{n-1} \left(\frac{1}{k+1}\bm{a}_k \right)+ \frac{n+1}{n+1} \bm{a}_n \\
    &~~~~~= \sum_{j=1}^{n} \left( -\bm{a}_{j-1}-\sum_{k=1}^{j-2}\frac{1}{k+1}\bm{a}_k \right) + n \sum_{k=1}^{n-1} \left( \frac{1}{k+1} \bm{a}_k \right)\\
    &~~~~~= \bm{0}\text{ according to Equation \ref{eq:inductionn}}.
  \end{aligned}
\end{equation}\hfill$\Box$

\section{Regarding the initialisation and estimation of the covariance matrix}
\label{ap:initSigma}

In our experiments, the training of the CDA turned out to be very sensitive to the initialization of $\bm{\Sigma}$. Here, we present an initialization strategy for starting the optimization with a $\bm{\Sigma}$ that we expect to be not too eccentric.

Section \ref{sec:distribILRL}, we saw how the covariance matrix $\bm{\Sigma}$ can be expressed by the Kullback-Leibler divergences ($\Dkl$) within each pair of classes\footnote{Keep in mind that since the densities in the base space are Gaussian with the same covariance matrix, the Kullback-Leibler divergences are symmetric.}. We propose here to initialize the mapping $g$ as the identity function and to initialize $\bm{\Sigma}$ with the $\Dkl$ measured in the feature space assuming that each class-conditional distributions are multivariate Gaussians with shared covariance (this is the standard LDA assumption). 
Even if there is no strong theoretical foundation for this choice of the initial $\bm{\Sigma}$ and $g$, these initializations appeared to be effective in our experiments. The intuition is that we initialize $\bm{\Sigma}$ with a kind of approximated divergence matrix not too eccentric and not too far from the ``true'' one\footnote{Note that the initialization is deterministic.}.

Note that this does not mean that an additional assumption on how the feature vectors are distributed is made. This is only for the initialization of the optimization. The parameters of $g$ and $\bm{\Sigma}$ are then free to take any value under the constraints of differentiability and invertibility for $g$ and symmetric positive definiteness for $\bm{\Sigma}$.

The symmetric positive definiteness of $\bm{\Sigma}$ is insured by optimizing instead the lower triangular matrix $\bm{L}$ from the Cholesky decomposition $\bm{\Sigma} = \bm{L} \bm{L}^T$, and since the diagonal elements of $\bm{L}$ must be positive, the $\log$-Cholesky parametrization is used \citep{pinheiro96unconstrainedparameterizations}. In our experiments, the estimation of $\bm{L}$ and the parameters of $g$ is done with automatic differentiation and gradient descent.

\section{A Gaussian three-classes and four-dimensional example}
\label{app:3gaussians}

In this Appendix, we provide complete the results of the Gaussian three-classes and four-dimensional example of Section \ref{sec:3gauss}. In this example, each class is generated by a multivariate normal distribution with its own mean and covariance matrix.
%\begin{tabular}{ll}
%    $\bm{\mu}_1 = \begin{bmatrix} 0\\ 5\\ 5\\ 5 \end{bmatrix}$ & $\bm{\Sigma}_1 = \begin{bmatrix} 1.32 & -0.43 & -0.61 & -0.10 \\ -0.43 & 2.24 & 2.55 & 0.66 \\ -0.61 & 2.55 & 2.97 & 0.72\\-0.1 & 0.66 & 0.72 & 0.22\end{bmatrix}$,\\
%    $\bm{\mu}_2 = \begin{bmatrix} 0.7\\ -0.4\\ 0\\ -1 \end{bmatrix}$ & $\bm{\Sigma}_2 = \begin{bmatrix} 2.40 & 0.57 & 0.07 & 1.19\\ 0.57 & 1.06 & 0.88 & 0.48\\ 0.07 & 0.88 & 1.16 & 0.63\\ 1.19 & 0.48 & 0.63 & 1.13 \end{bmatrix}$,\\
%    $\bm{\mu}_3 = \begin{bmatrix} -3\\ 2\\ 2\\ 2 \end{bmatrix}$ & $\bm{\Sigma}_3 = \begin{bmatrix} 1.14 & 1.00 &  0.19 &  0.30\\ 1.00 & 1.85 & 0.35 & 0.68\\ 0.19 & 0.35 & 1.17 & -0.80\\ 0.30 & 0.68 & -0.80 & 1.07\end{bmatrix}$.
%  \end{tabular}\\
\begin{figure}[h]
  \centering
  \begin{subfigure}[b]{0.325\linewidth}
    \centering
    \includegraphics[width=0.9\linewidth]{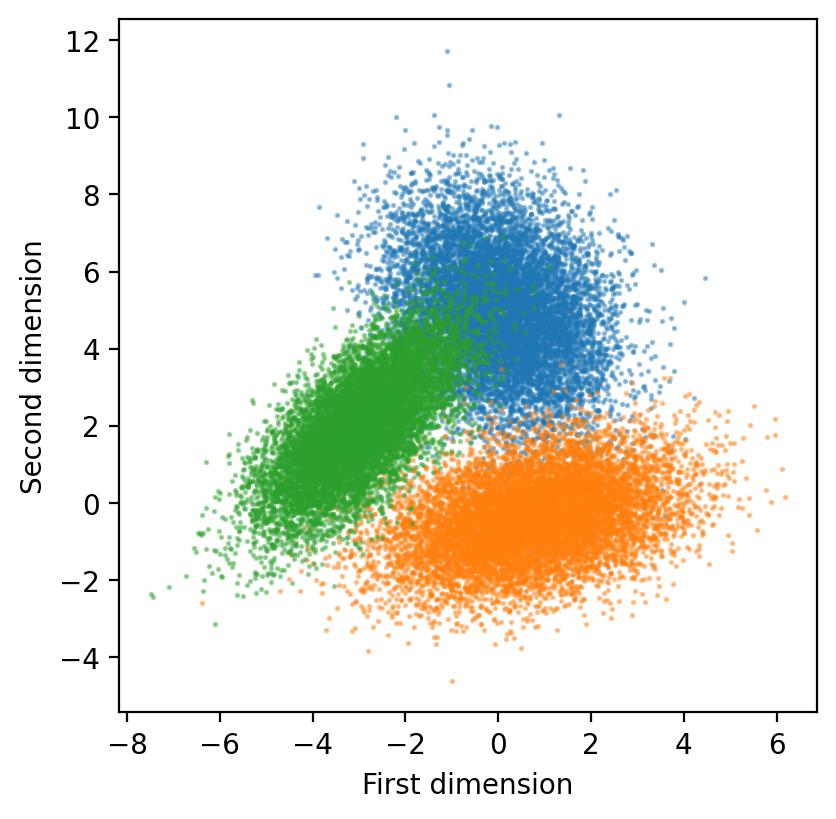}
    \caption{The 1st and 2nd dimensions}
  \end{subfigure}
  \begin{subfigure}[b]{0.325\linewidth}
    \centering
    \includegraphics[width=0.9\linewidth]{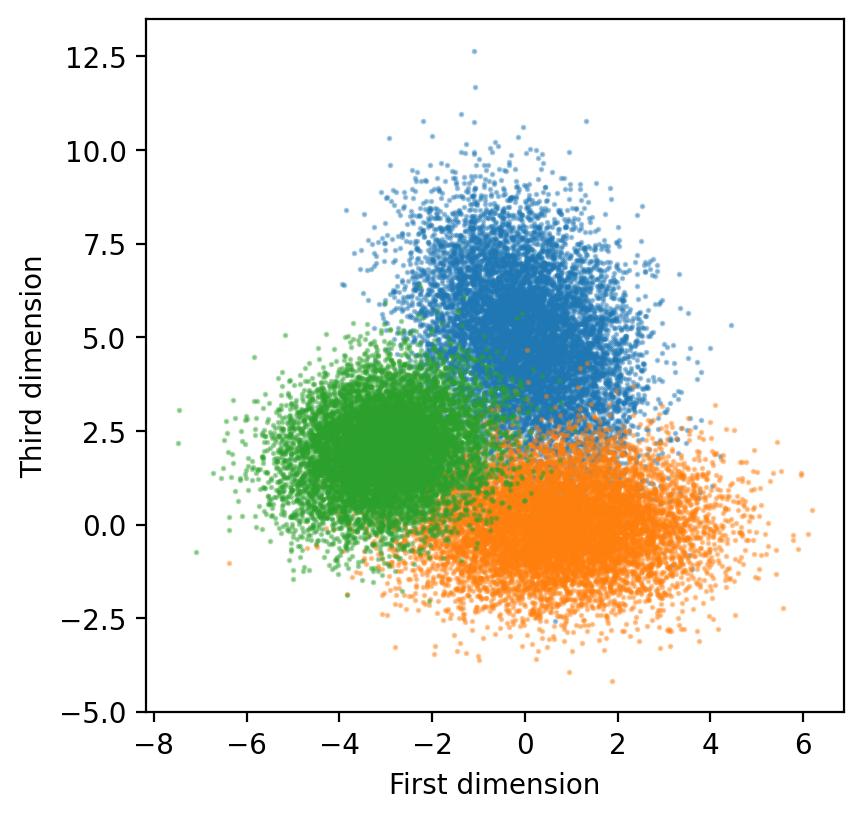}
    \caption{The 1st and 3rd dimensions.}
  \end{subfigure}
  \begin{subfigure}[b]{0.325\linewidth}
    \centering
    \includegraphics[width=0.9\linewidth]{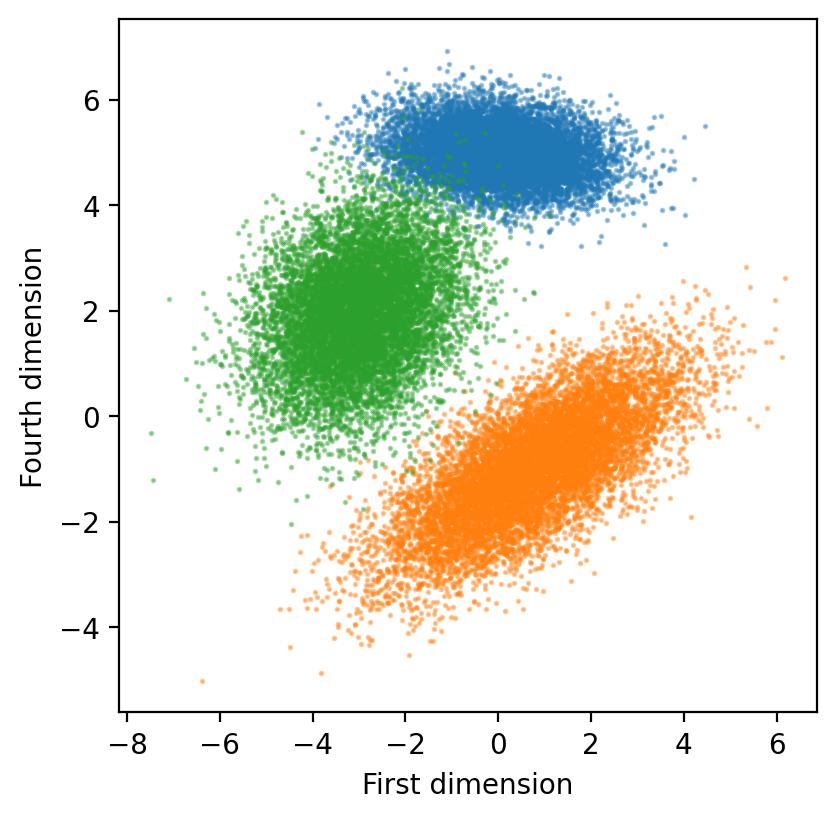}
    \caption{The 1st and 4th dimensions.}
  \end{subfigure}
  \begin{subfigure}[b]{0.325\linewidth}
    \centering
    \includegraphics[width=0.9\linewidth]{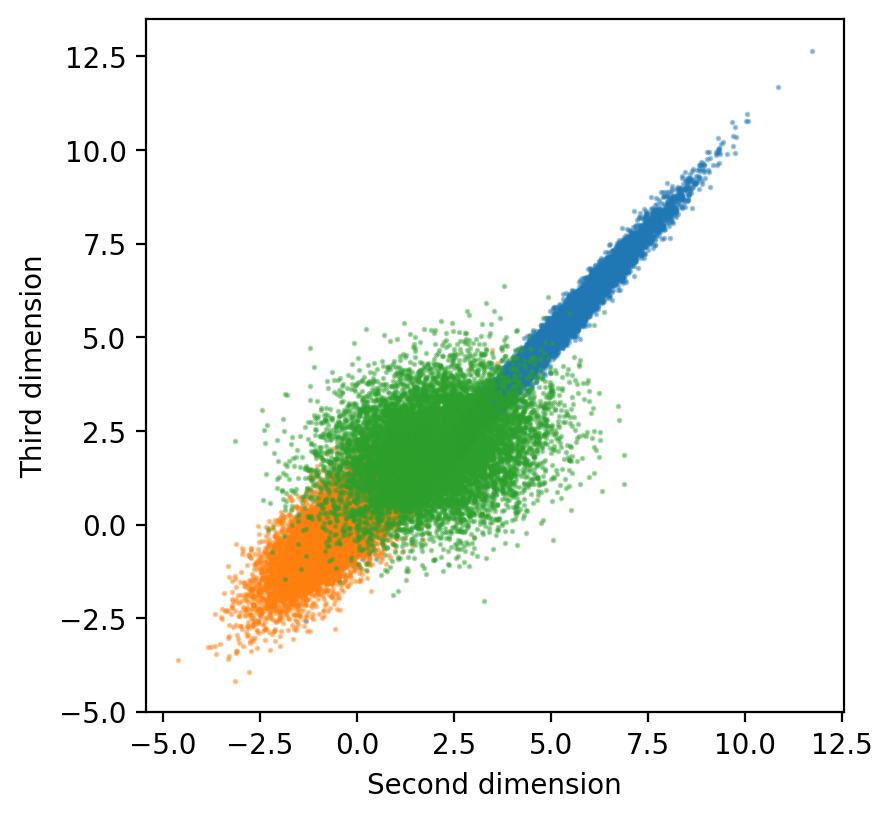}
    \caption{The 2nd and 3rd dimensions.}
  \end{subfigure}
  \begin{subfigure}[b]{0.325\linewidth}
    \centering
    \includegraphics[width=0.9\linewidth]{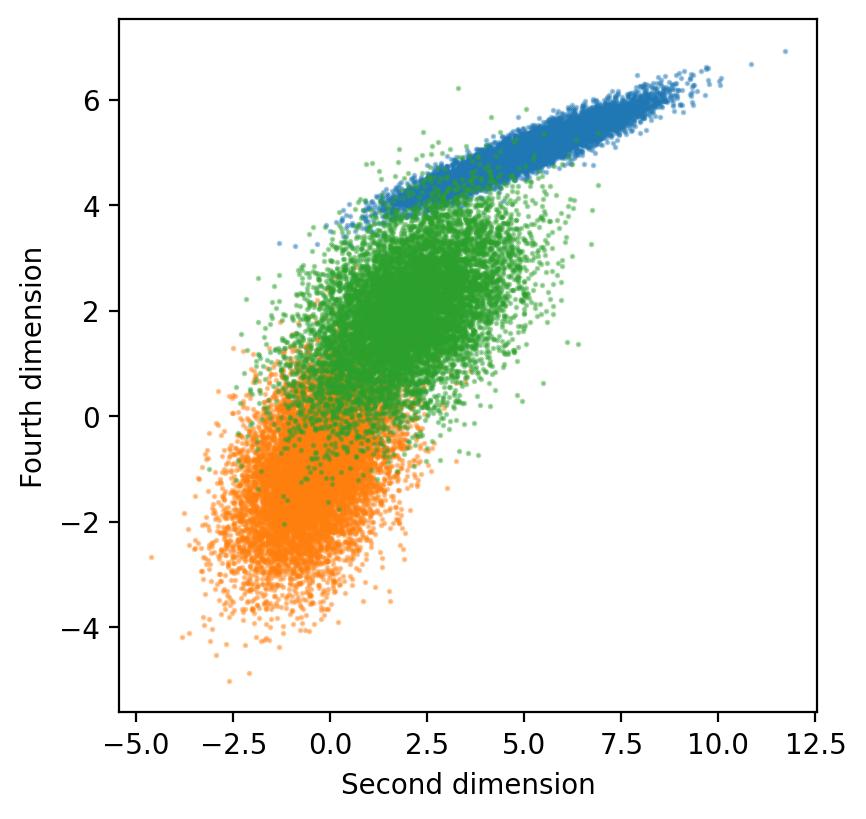}
    \caption{The 2nd and 4th dimensions}
  \end{subfigure}
  \begin{subfigure}[b]{0.325\linewidth}
    \centering
    \includegraphics[width=0.9\linewidth]{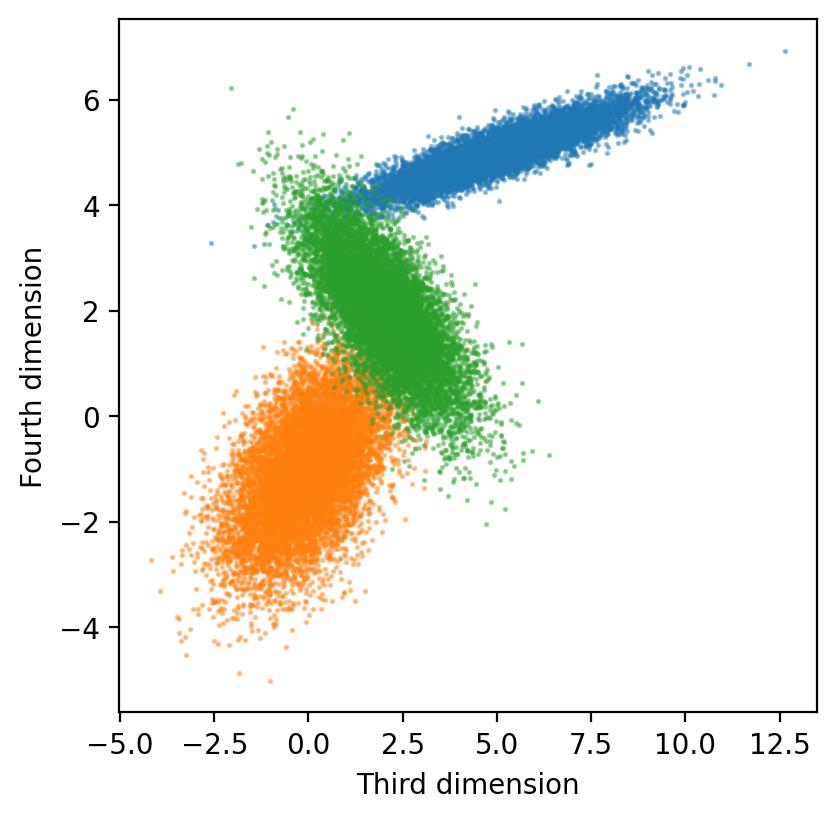}
    \caption{The 3rd and 4th dimensions.}
  \end{subfigure}
  \caption{Training set for the three classes CDA example with non-shared covariance Gaussian. The colors indicate to which of the three classes a sample belongs: blue for $C_1$, orange for $C_2$ and green for $C_3$.}
  \label{fig:train_ILRL_gaussians_nonshared}
\end{figure}

\begin{figure}[h]
  \centering
  \begin{subfigure}[b]{0.325\linewidth}\centering
    \includegraphics[scale=0.45]{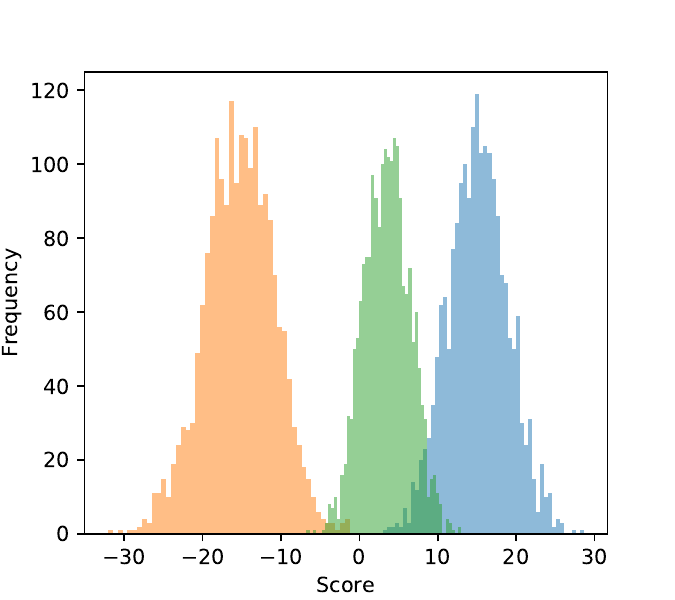}
    \caption{Class 1 against 2 with LDA.}
    \label{fig:score_12_lda}
  \end{subfigure}
  \begin{subfigure}[b]{0.325\linewidth}\centering
    \includegraphics[width=\linewidth]{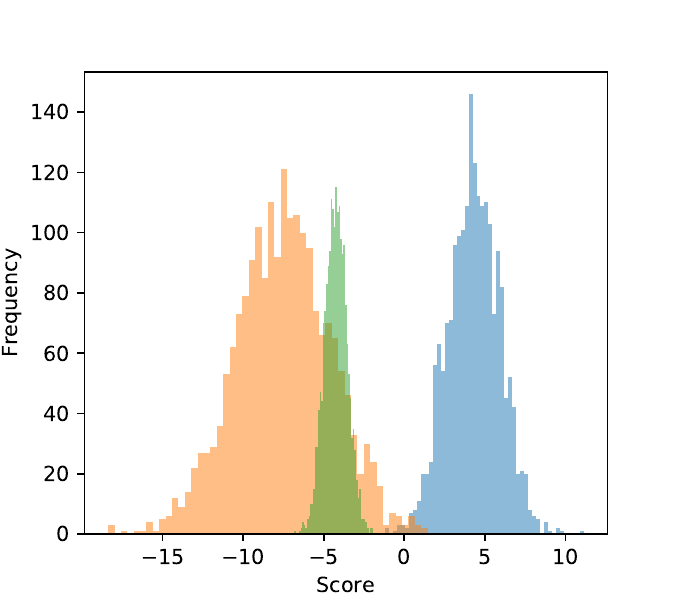}
    \caption{Class 1 against 3 with LDA.}
    \label{fig:score_13_lda}
  \end{subfigure}
  \begin{subfigure}[b]{0.325\linewidth}\centering
    \includegraphics[width=\linewidth]{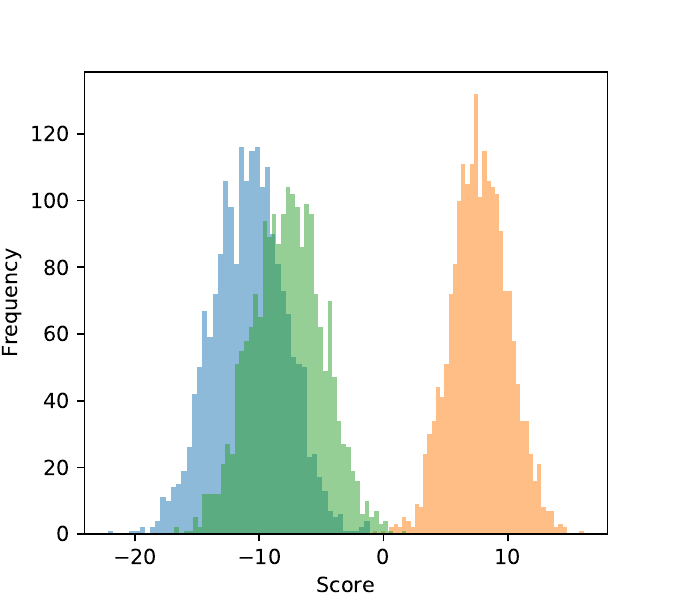}
    \caption{Class 2 against 3 with LDA.}
    \label{fig:score_23_lda}
  \end{subfigure}
  \begin{subfigure}[b]{0.325\linewidth}\centering
    \includegraphics[width=\linewidth]{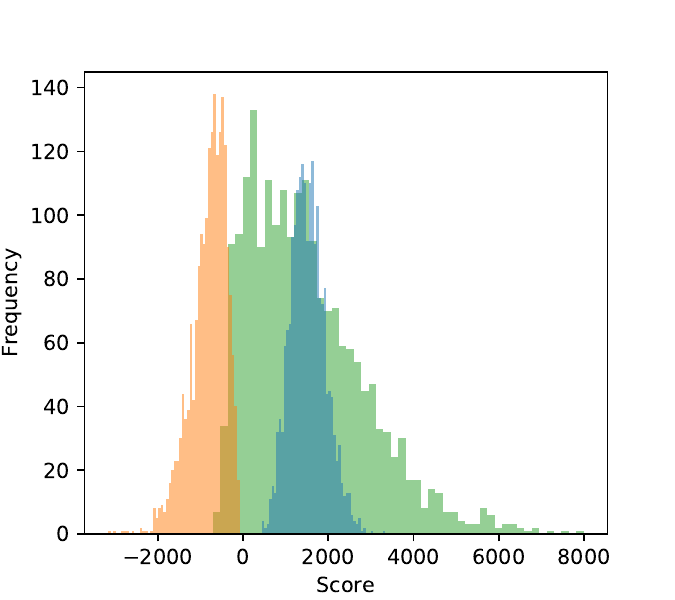}
    \caption{Class 1 against 2 with QDA.}
    \label{fig:score_12_qda}
  \end{subfigure}
  \begin{subfigure}[b]{0.325\linewidth}\centering
    \includegraphics[width=\linewidth]{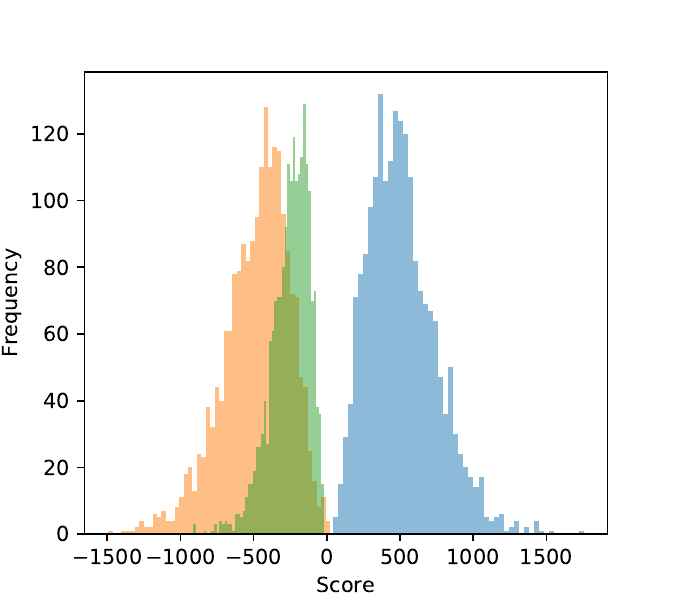}
    \caption{Class 1 against 3 with QDA.}
    \label{fig:score_13_qda}
  \end{subfigure}
  \begin{subfigure}[b]{0.325\linewidth}\centering
    \includegraphics[width=\linewidth]{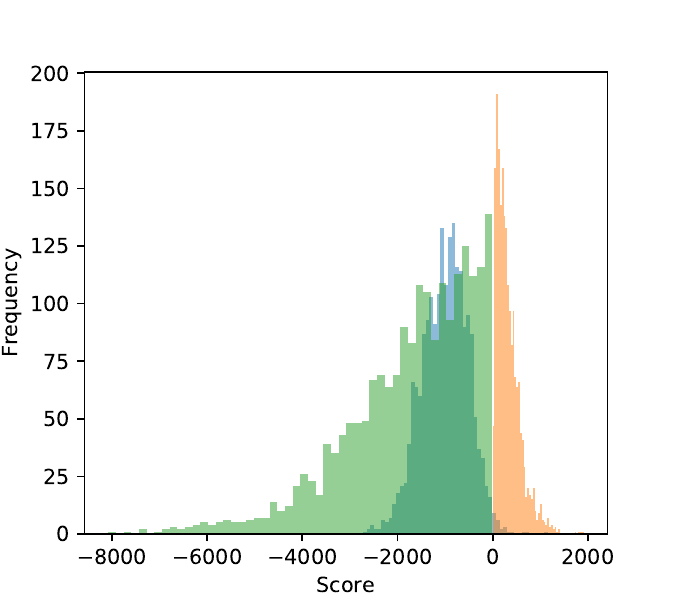}
    \caption{Class 2 against 3 with QDA.}
    \label{fig:score_23_qda}
  \end{subfigure}
  \begin{subfigure}[b]{0.325\linewidth}\centering
    \includegraphics[width=\linewidth]{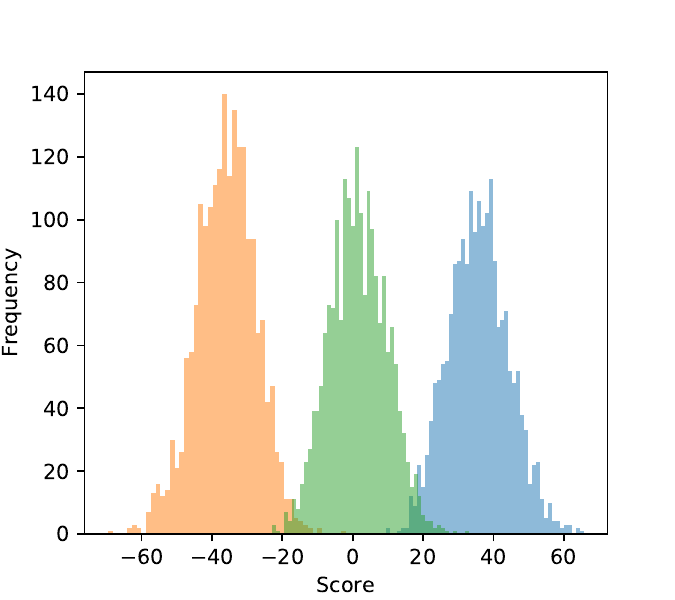}
    \caption{Class 1 against 2 with CDA.}
    \label{fig:score_12_cda}
  \end{subfigure}
  \begin{subfigure}[b]{0.325\linewidth}\centering
    \includegraphics[width=\linewidth]{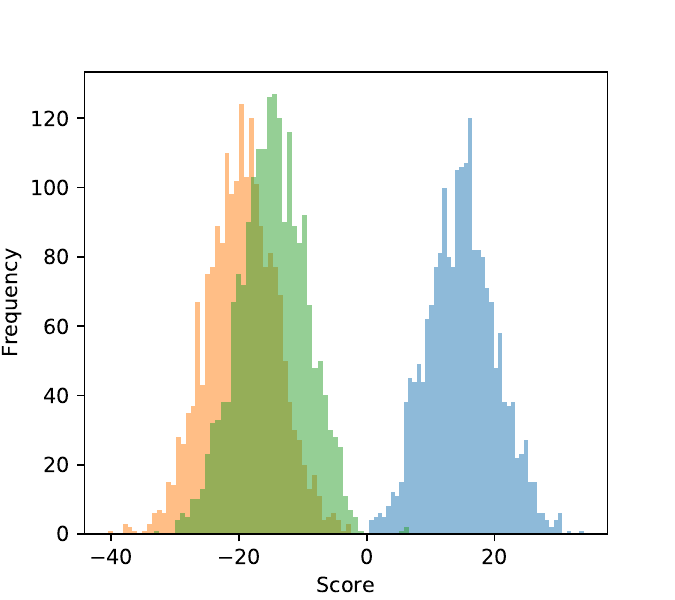}
    \caption{Class 1 against 3 with CDA.}
    \label{fig:score_13_cda}
  \end{subfigure}
  \begin{subfigure}[b]{0.325\linewidth}\centering
    \includegraphics[width=\linewidth]{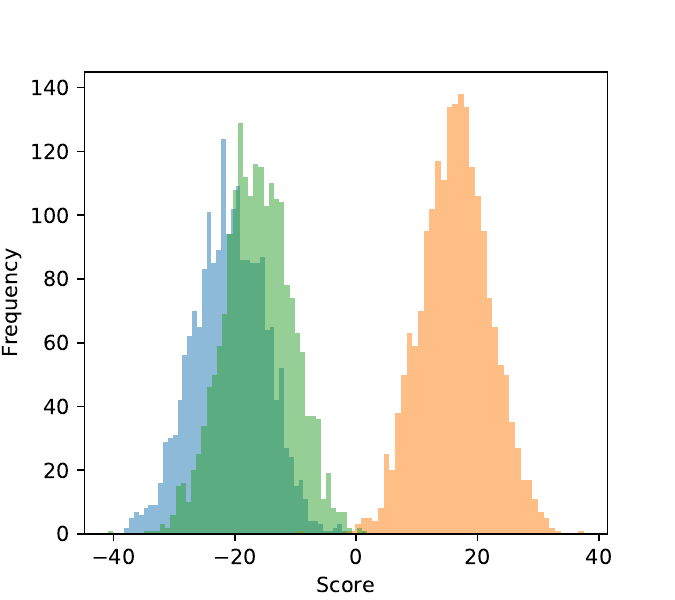}
    \caption{Class 2 against 3 with CDA.}
    \label{fig:score_23_cda}
  \end{subfigure}
\caption{Histograms of the LLRs of one class against another, for the non-shared covariance Gaussian example, given by LDA, QDA, and CDA. $C_1$, $C_2$, and $C_3$ are respectively blue, orange, and green.}
  \label{fig:hist_gaussians_nonshared}
\end{figure}

We already discussed Figure \ref{fig:test_base_gaussians_nonshared} in Section \ref{sec:3gauss}. QDA's results were not given because QDA does not have an information-preserving mapping of the data into a same-dimensional space. However, it can still be used to compute LLRs. The LLRs of class $i$ against class $j$ is given by:
\begin{equation}
  \label{eq:llrqda}
  \begin{aligned}
    \log \frac{f(\bm{x} \mid \bm{\mu}_i, \bm{\Sigma}_i)}{f(\bm{x} \mid \bm{\mu}_j, \bm{\Sigma}_j)} &= \frac{1}{2} \bm{x}^T \left( \bm{\Sigma}_j^{-1} - \bm{\Sigma}_i^{-1} \right) \bm{x} + \bm{x}^T \left( \bm{\Sigma}_i^{-1} \bm{\mu}_i - \bm{\Sigma}_j^{-1}\bm{\mu}_j \right)\\
    &~~~~+ \frac{1}{2} \left( \bm{\mu}_j^T \bm{\Sigma}_j^{-1} \bm{\mu}_j -  \bm{\mu}_i^T \bm{\Sigma}_i^{-1} \bm{\mu}_i \right) + \frac{1}{2} \log \frac{\lvert \bm{\Sigma}_j \rvert}{\lvert \bm{\Sigma}_i \rvert}
  \end{aligned}
\end{equation}
Figure \ref{fig:hist_gaussians_nonshared} shows the histograms of the LLRs obtained with LDA (Figures \ref{fig:score_12_lda},\ref{fig:score_13_lda} and \ref{fig:score_23_lda}), QDA (Figures \ref{fig:score_12_qda},\ref{fig:score_13_qda} and \ref{fig:score_23_qda}), and CDA (Figures \ref{fig:score_12_cda}, \ref{fig:score_13_cda} and \ref{fig:score_23_cda}). For the latter, the LLR in favor of a class against another is obtained by projecting the ILRL vector on the orthogonal direction of the maximum probability decision boundaries between the two classes\footnote{To be more precise, projecting the data into the unit vector orthogonal to the decision boundaries gives the LLR up to a scaling factor $\frac{1}{\sqrt{2}}$. See the definition of the ILR transformation in Equation \ref{eq:generalilrcomponent}: its first component is $\frac{1}{\sqrt{2}}$ times the log-ratio.}. For the LDA and the CDA, the class-conditional distributions of the LLRs look Gaussian as expected. However, for the LDA, they are not symmetric as required by the idempotence property. We therefore expect the LDA's LLRs to have a lower calibration quality. For the QDA the histograms are not symmetric but this does not suggest that the scores are not calibrated. Indeed, the idempotence constraint of Theorem \ref{prop:distrib_LLR} and Theorem \ref{prop:distribILRL} are for normally distributed LLRs, while they are here not Gaussian for the QDA\footnote{To be more precise, since the data is here normally distributed and the mapping is quadratic, the LLRs are distributed according to a generalised chi-squared distribution.}.

To better assess the discrimination and calibration quality of the LLRs, Table \ref{tab:cllr_gaussians_nonshared} provides $\Cllr$ measures. The LDA has the worst discrimination and calibration which is not surprising since it is based on the shared covariance assumption. QDA models the best the data and the resulting LLRs have the best discrimination and calibration which is again not surprising since the data is actually distributed as described by the model. However, as mentioned above, the QDA does not provide an information-preserving transformation necessary for data generation or conversion. On the contrary, the CDA does, and still has good discrimination and calibration performance.
\begin{table}
  \centering
  \caption{$\Cllr$ measures for the non-shared covariance example. Samples from the non-concerned class are discarded.}
  \label{tab:cllr_gaussians_nonshared}
  \begin{tabular}{|c|c|c|c|c|c|c|}
    \hline
   \multirow{2}{*}{compared classes} &  \multicolumn{2}{c|}{\makecell{LDA}} &  \multicolumn{2}{c|}{\makecell{QDA}} & \multicolumn{2}{c|}{\makecell{CDA}}\\
    \cline{2-7}
    & $\Cllr$ {\scriptsize[bit]} & $\Cllrmin$ {\scriptsize[bit]} & $\Cllr$ & $\Cllrmin$ & $\Cllr$ & $\Cllrmin$ \\
    \hline
    \hline
    1 vs 2 & 1.72{\tiny $10^{-3}$} & 0.0 & 0.0 & 0.0 & 4.85{\tiny $10^{-5}$} & 0.0 \\
    \hline
    1 vs 3 & 1.98 & 1.43{\tiny $10^{-1}$} & 1.72{\tiny $10^{-9}$} & 0.0 & 9.46{\tiny $10^{-3}$} & 5.04{\tiny $10^{-3}$}\\
    \hline
    2 vs 3 & 2.00{\tiny $10^{-1}$} & 1.76{\tiny $10^{-2}$} & 6.37{\tiny $10^{-4}$} & 0.0 & 8.46{\tiny $10^{-3}$} & 5.31{\tiny $10^{-3}$}\\
    \hline
  \end{tabular}
\end{table}
\newpage
\bibliography{biblio}

\end{document}